%% file: paper.tex
\documentclass{article}

\usepackage{arxiv}

\usepackage[utf8]{inputenc}
\usepackage[T1]{fontenc}
\usepackage{amsmath}
\usepackage{amssymb}
\usepackage{microtype}
\usepackage{booktabs}
\usepackage{graphicx}
\usepackage{algpseudocode}
\algrenewcommand\algorithmiccomment[1]{\hfill$\triangleright$\,{\scriptsize #1}}
\algtext*{EndWhile}\algtext*{EndIf}\algtext*{EndFor}
\usepackage{xcolor}
\definecolor{preprintlink}{HTML}{1A5276}

\usepackage[numbers,sort&compress]{natbib}

\usepackage{hyperref}
\hypersetup{
  colorlinks=true,
  linkcolor=black,
  urlcolor=preprintlink,
  citecolor=preprintlink,
  pdftitle={Where to cut, how deep: BPE and Unigram-LM on chemistry SMILES},
  pdfauthor={Hunter Heidenreich},
  pdfkeywords={subword tokenization, byte-pair encoding, Unigram-LM, SMILES,
    chemical language models, vocabulary, cheminformatics},
}


\renewcommand{\shorttitle}{Where to cut, how deep: BPE vs Unigram-LM}

\title{Where to cut, how deep: BPE and Unigram-LM on chemistry SMILES}

\author{Hunter Heidenreich \\
  Independent Researcher \\
  \texttt{hheiden0@gmail.com} \\
  \href{https://orcid.org/0009-0001-0335-4803}{orcid.org/0009-0001-0335-4803}}

\date{}

\begin{document}

\maketitle

\begin{abstract}
Every chemical language model reading SMILES begins with a
tokenizer, yet the field has inherited byte-pair encoding (BPE) from
natural language with little scrutiny. In natural language, BPE's
principal alternative, Unigram-LM, is known to build structurally
different vocabularies. Whether that
contrast survives in chemistry was open: the complete glyph base already
covers every conformant molecule, so the learned pieces add compression rather than
coverage, and a tiny alphabet under hard valence
constraints could drive two frequency-based algorithms to converge. We
report a controlled comparison of BPE and Unigram-LM over a
fixed 165-token chemistry base, at the small vocabulary sizes where token
embeddings are learnable, across three corpus typologies
(diverse, drug-like, natural-products) and both
pre-tokenization boundary policies. The two do not converge.
In all 22 matched conditions they build \emph{near-disjoint}
subword vocabularies: cross-algorithm Jaccard overlap on the
learned pieces above the shared base never exceeds $0.161$, and at most
$0.05$ once weighted toward the high-frequency pieces a model updates most. Unigram-LM also segments
held-out molecules into $29$--$41\%$ more tokens; the arms largely agree
on \emph{where} to cut but not \emph{how deeply}, so
BPE's segmentation is a strict coarsening of
Unigram-LM's on $80$--$99\%$ of molecules. The separation holds across
corpus, boundary, and vocabulary size, persisting even at eight times
that scale, past where embeddings remain
learnable; only token-frequency imbalance attenuates in magnitude, shrinking
with vocabulary size and most on the natural-products corpus, without closing.
The subword algorithm is therefore a modeling decision,
not a free default. We release all
trained tokenizers and per-condition measurements.
\end{abstract}

\begin{figure}[ht]
  \centering
  \includegraphics[width=0.65\linewidth]{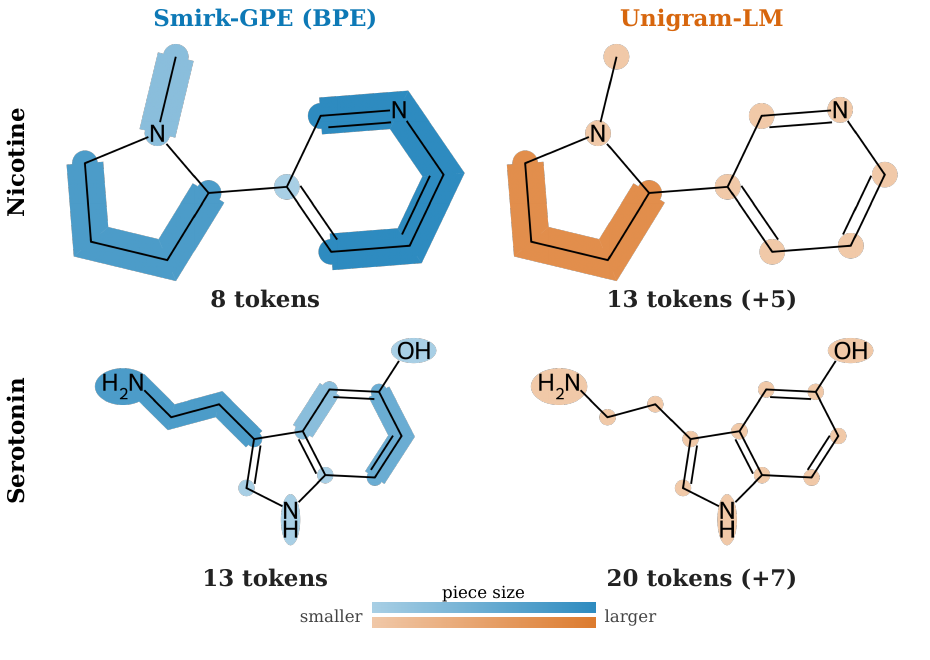}
  \caption{Nicotine and serotonin under the two algorithms' $V{=}1024$
  vocabularies (molecules in rows, algorithms in columns). Hue marks the
  algorithm (blue Smirk-GPE/BPE, orange Unigram-LM); darker shading marks a
  larger piece, on the same scale for both. BPE builds a few large pieces
  spanning whole rings and chains; Unigram-LM stays near-atomic, a scatter of
  small beads, emitting more tokens per molecule ($+5$ and $+7$ here, a wider
  gap than the $\sim$one-third average across held-out molecules).}
  \label{fig:graphical-abstract}
\end{figure}

\section{Introduction}\label{sec:intro}

Every chemical language model that operates on a SMILES string
\citep{weininger1988smiles} begins with a tokenizer that maps it to integer
IDs, fixing the vocabulary the model embeds, the granularity
at which it operates, and hence the effective sequence
length. It is a foundational design choice
\citep{alqahtani2026stop,kalamkar2025tokenization}, almost always made by
default: SMILES tokenizers inherit byte-pair encoding (BPE)
\citep{sennrich2016bpe-nmt} from natural-language practice, and its chemistry
descendants (SPE, APE, Smirk-GPE) are all BPE variants. The principal
alternative, Unigram-LM \citep{kudo2018subword-reg}, sees occasional chemistry
use but has never been compared to BPE at matched conditions on a fixed
chemistry-grammatical base.

That choice was long overshadowed by a more basic axis, vocabulary
\emph{coverage}: whether every string gets a token or part falls through to
\texttt{[UNK]}.
\citet{wadell2026smirk} found that coverage, not the subword scheme, separates
chemistry tokenizers downstream, but only across heterogeneous bases at large
native vocabularies, never isolating the subword-\emph{algorithm} axis. Smirk
\citep{wadell2026smirk} closes coverage with a complete $165$-token OpenSMILES
base \citep{james2016opensmiles},\footnote{Following \citet{wadell2026smirk}: $158$
OpenSMILES glyphs plus seven special tokens (including \texttt{[UNK]}).
We write ``glyph'' for the $158$ chemistry-grammatical pieces.}
emitting no \texttt{[UNK]} on conformant input. Our study begins there: pushing
the vocabulary into the small regime, we isolate the subword algorithm, still
inherited from natural language, as the design choice under study.

The inheritance is not obviously safe. Chemistry SMILES is statistically unlike
language, with a tiny base alphabet and strings under hard structural
constraints, so natural-language priors may not transfer, and the ones that do
cut both ways. Priors point toward divergence: \citet{bostrom2020bpe-suboptimal}
on natural language and the one prior chemistry head-to-head both report it,
though under confounds we revisit in \S\ref{ssec:algo}. Yet a convergence null
stays live: in natural language a key segmentation statistic (whole-pretoken
absorption) converges across the two algorithms at large $V$
\citep{reddy2025diminishing-tokenization}, and in chemistry both select pieces by
frequency over a small, $158$-glyph alphabet where valence and ring-closure
constraints sharply concentrate which glyph sequences occur; that concentration
could drive them to absorb the same high-frequency substructures and converge.

We therefore ask: \emph{on chemistry SMILES, holding everything but the
algorithm fixed, does the choice between BPE and Unigram-LM still shape the
structure of the resulting tokenizer, or does chemistry's small,
valence-constrained alphabet drive the two to converge on the same
vocabulary?} We find that the algorithm shapes the tokenizer. At a matched
condition (same corpus, vocabulary size, and boundary policy), BPE and Unigram-LM
build measurably different vocabularies (Figure~\ref{fig:graphical-abstract}) in
three senses: \emph{which} pieces a
vocabulary contains (membership), \emph{how finely} it segments held-out
molecules (granularity), and \emph{how unevenly} token usage is distributed
(distribution); two further results fix its scope. All five
are stable properties of the algorithm pair, holding across corpus typology,
boundary policy, and vocabulary size, with only the distribution gap's magnitude
attenuating on one corpus (Section~\ref{sec:results}).

This is a tokenizer-level study: we train no language models and make no claim
about which algorithm is better downstream. It is the precondition for that
comparison, establishing that the two do not yield interchangeable vocabularies,
dislodging the field's inherited default. Our
contributions are fourfold:
(i) the first controlled,
chemistry-grammatical, small-vocabulary comparison of BPE and Unigram-LM
across a corpus-typology range and both boundary policies; (ii)
chemistry-side measurements, previously unreported for SMILES, of three of the
four mechanism diagnostics carried
from the natural-language literature (dead-zone surplus, scaffold fraction, and
whole-pretoken absorption; the remaining one, segmentation entropy, is intrinsic to
Unigram-LM),
alongside a structural-character and non-canonicity battery that localizes the
divergence; (iii) two results fixing its scope, that the near-disjoint
vocabularies nonetheless parse compatibly and that the divergence persists at
$8\times$ the headline vocabulary, past the learnable regime; and
(iv) all trained tokenizers and per-condition measurements, released for
re-analysis.

\section{Background and related work}\label{sec:background}

\subsection{Chemistry SMILES tokenizers}\label{ssec:landscape}

Chemistry-side tokenizers are best organized by how each handles
\emph{coverage} (\S\ref{sec:intro}). \emph{Atom-wise}
regex tokenizers \citep{schwaller2018found} split on atom symbols and
treat each bracketed atom as one token; DeepChem's implementation
\citep{ramsundar2019deepchem}, popularized by ChemBERTa
\citep{chithrananda2020chemberta}, is widely used. Because a
bracketed atom jointly encodes element, isotope, chirality, charge, and
hydrogen count, the OpenSMILES bracketed-atom space exceeds $28$ trillion
permutations \citep{wadell2026smirk}, far beyond any fixed vocabulary; the
unrepresented remainder becomes \texttt{[UNK]}. \emph{SMILES Pair Encoding} (SPE)
\citep{li2021smiles-pe} and \emph{Atom Pair Encoding} (APE)
\citep{leon2024smiles-selfies-tokenization-chemical-lm} are BPE adapted
to SMILES with atom-level initialization, learning $\sim$3{,}000--5{,}300
tokens; both inherit this bracketed-atom-as-token
convention and still emit \texttt{[UNK]} \citep{wadell2026smirk}; for SPE this
reaches $\sim$19\% of tokens on MoleculeNet \citep{moleculenet} and $\sim$50\% on
tmQM \citep{balcells2020tmqm}. \emph{Atom-in-SMILES}
\citep{ucak2023atom-in-smiles} replaces atomic tokens with
environment-aware tokens; it is bijective but has an implicit,
open-ended vocabulary. \emph{Smirk} \citep{wadell2026smirk} is a regex
glyph decomposition of OpenSMILES over a $165$-token base; any
OpenSMILES-conformant string decomposes into these glyphs with no
\texttt{[UNK]}. \emph{Smirk-GPE} is BPE on top of that glyph base. A separate
line of work sets coverage aside: rather than constructing a vocabulary, it
post-processes SMILES token sequences for compression (e.g.\ trie-based
refinement and transition-graph filtering \citep{radhakrishnan2026optimizing}),
building on the same frequency-driven subword primitives whose structure we
characterize here.

We study SMILES throughout; alternative notations such as SELFIES
\citep{krenn2020selfies}, whose strings always decode to valid molecules, and
DeepSMILES \citep{oboyle2018deepsmiles} vary the
representation rather than the subword algorithm and are out of scope. That notation
choice is consequential (SMILES-based generators outperform SELFIES on
distribution-learning metrics, and the invalid strings SMILES admits act as
implicit quality control \citep{skinnider2024invalid-smiles}), yet orthogonal
to our algorithm contrast.

\subsection{BPE versus Unigram-LM}\label{ssec:algo}

BPE and Unigram-LM construct their vocabularies in opposite directions
(Appendix Fig.~\ref{fig:algos}). BPE
\citep{sennrich2016bpe-nmt} builds bottom-up, greedily merging the
most-frequent adjacent symbol pair until the target size is reached; it
is deterministic given the corpus and target size. Unigram-LM
\citep{kudo2018subword-reg} builds top-down, fitting a unigram
probability distribution over a large seed vocabulary by
expectation-maximization and pruning the pieces whose removal least hurts
corpus likelihood; it provides per-segmentation probabilities and
supports subword regularization at training time. In chemistry it appears in
models such as BARTSmiles \citep{chilingaryan2024bartsmiles} and
ReactionT5\,/\,CompoundT5 \citep{sagawa2023reactiont5}. (WordPiece
\citep{schuster2012jp-kr-voice-search} builds bottom-up like BPE, differing
only in its merge criterion, and is out of scope.)

Both algorithms run downstream of a pre-tokenizer, the step that first splits the
raw string into \emph{pretokens} (the chunks within which subword merges stay
confined). In natural language that boundary is the highest-impact design
choice across studies, above vocabulary
size and corpus \citep{wegmann2025tokenization, reddy2025diminishing-tokenization}.
Recent methods relax it for compression \citep{liu2025superbpe,
schmidt2025boundless-bpe}. We treat its chemistry analog, whether merges may
cross the bracketed-atom boundary, as a first-class axis (\S\ref{ssec:training}).

Prior work has measured \emph{where} BPE and Unigram-LM diverge, each study
supplying a diagnostic we carry to chemistry. \citet{bostrom2020bpe-suboptimal}
found Unigram-LM matches or beats BPE on natural language for structural reasons,
though the advantage varied by an order of magnitude across their two corpora;
that structural account has three
legs: better morphological alignment, less over-merging of frequent affixes,
and a smaller dead-zone surplus (BPE carries $\sim$1{,}500 more
intermediate-merge tokens that fire too rarely to train an embedding).
Morphological alignment has no gold-standard SMILES segmentation to score
against, so Bostrom's better-alignment metric does not transfer; the
over-merging leg becomes our granularity contrast (fertility), and the dead-zone
surplus is a symmetric cross-arm count we measure directly
(\S\ref{ssec:r-mechanism}).
\citet{lian2024scaffold-bpe} quantify the BPE-side fraction as
$6.07\%$ ``scaffold tokens'' for the LLaMA-32K tokenizer \citep{touvron2023llama} and show
that removing them yields downstream gains at 6.7B parameters.
\citet{reddy2025diminishing-tokenization} report a related diagnostic, the
fraction of pretokens absorbed as single tokens, and find it ``remarkably
high across all algorithms,'' increasing with vocabulary size.

The single prior chemistry-side head-to-head is
\citet{temizer2024chemical-tokenization}, who learned BPE, WordPiece, and
Unigram vocabularies on $\sim$2.3M ChEMBL \citep{zdrazil2023chembl}
SMILES at $V \in \{8\text{K}, 16\text{K}, 32\text{K}\}$. They report that
BPE and WordPiece vocabularies overlap by $\sim$80\% while Unigram
overlaps the others by at most $\sim$47\%, that the overlap shrinks as
$V$ grows, and that Unigram ``chemical words'' run longer and are less
often RDKit-valid. Their study establishes that a contrast is
\emph{visible}. But it runs on raw SMILES (no fixed chemistry base, confounding
the algorithm with the absence of an atomic level) and one drug-like corpus
under a single pre-tokenization policy, leaving open whether the contrast is
algorithmic, corpus-stable, or boundary-robust (the three questions we test). A fourth
confound, large $V$, places it outside the learnable small-vocabulary regime
(\S\ref{ssec:typology}).

\subsection{Corpus typology and the small-vocabulary
regime}\label{ssec:typology}

The chemistry analog of the English-versus-Japanese axis of
\citet{bostrom2020bpe-suboptimal} is not morphology but alphabet diversity
$\times$ topology $\times$ functional-group distribution. We span it with
three headline corpora: \textbf{PubChem} \citep{pubchem2023} (diverse),
\textbf{ZINC-22} \citep{zinc22} (drug-like), and \textbf{COCONUT}
\citep{sorokina2021coconut} (natural-products), plus the 1\% Sample of
Enamine's make-on-demand \textbf{REAL-Space} \citep{grygorenko2020real,
wadell2026smirk}, which serves only as an anchor (per-corpus
characteristics in Table~\ref{tab:corpora}). Combinatorially enumerated from a
fixed building-block set, REAL-Space has a narrow alphabet, and even a 1\%
sample is large enough to exhaust the available merges, making it a
\emph{super-saturated} drug-like probe.

Vocabulary size is constrained by embedding learnability: above roughly
$1{,}000$ tokens learned pieces can fire too rarely to train embeddings, a
token-usage dead-zone (the regime boundary, not the cross-arm surplus of
\S\ref{ssec:algo}) that \citet{wadell2026smirk}
report on the super-saturated REAL-Space (their SI Fig.~S1). The small-vocabulary regime is therefore
where the choice between the two algorithms is most consequential: with slots
scarce, each must commit to \emph{which} high-frequency patterns to absorb
(\S\ref{sec:intro}). We track each condition against the learnability bar of
\citet{gowda2020optimal-vocab-nmt} (\S\ref{ssec:vregime}).

\section{Methods}\label{sec:methods}

\subsection{Experimental design: the grid}\label{ssec:grid}

Each trained tokenizer (one \emph{arm}) is an (algorithm, corpus, $V$,
boundary) tuple. A \emph{condition}, equivalently a grid cell, is a
(corpus, $V$, boundary) point; in a \emph{matched} condition both arms are
trained, so the only cross-arm difference is the selection algorithm. One control and two
robustness axes keep that contrast clean: (i) fixing the base alphabet
(\S\ref{ssec:training}) removes coverage as a confound; (ii) spanning three
corpus typologies (\S\ref{ssec:corpora}) separates an algorithmic effect from a
single-corpus artifact; and (iii) varying the boundary policy
(\S\ref{ssec:training}) separates that effect from a bracket-boundary artifact.
Stability of the membership and granularity contrasts across these last two axes
is what we report as robustness; $V$ checks scale.

The headline grid is $2\,\text{algorithm} \times 3\,\text{corpus} \times
3\,V \times 2\,\text{boundary} = 36$ tokenizers ($18$ matched
conditions), with $V \in \{256, 512, 1024\}$. To it we add a four-tokenizer
PubChem $V{=}2048$ sensitivity block and a four-tokenizer REAL-Space $V{=}1024$
anchor block (two matched conditions each), for $44$ trained tokenizers, the
two arms of each of $22$ matched conditions. At a single
coordinate no pair can be formed: on ZINC-22 at $V{=}2048$ the narrow alphabet
pushes both arms past the learnable regime, the Unigram-LM arm so far that we
leave it untrained (\S\ref{sec:results}). We train only the BPE arm there, to
demonstrate that ceiling, and report it single-arm; its two boundary variants
are additional to the $44$.

\subsection{Corpora and preprocessing}\label{ssec:corpora}

All four corpora pass through the same policy: RDKit \citep{rdkit2026} isomeric
canonicalization, exact-string deduplication, and a base-conformance filter
(procedure, drop rates, and pinned versions in Appendix~\ref{app:preproc}). We
pin the RDKit build, since canonical SMILES is toolkit-dependent
\citep{oboyle2012universal}, and deduplicate because both algorithms select by
frequency. The policy is otherwise narrower than a standard pipeline: we do
\emph{not} salt-strip (real input carries multi-component salts and the
\texttt{.} disconnection symbol), neutralize charges
(collapsing the charged bracketed atoms \texttt{[O-]}, \texttt{[NH3+]} central to
the boundary axis), or cap heavy-atom count (clipping the large COCONUT molecules
that define its typology).

\input{tables/corpora}

A final filter enforces the study's premise: the Smirk base covers any
OpenSMILES-conformant string with no \texttt{[UNK]}, but RDKit's non-Kekul\'e
canonicalization can emit off-base atoms (aromatic Si, Te). We drop any molecule
the bare base cannot cover without \texttt{[UNK]}, closing each corpus under the
base; left in, these atoms would contaminate both vocabularies and perturb the
overlap. Only PubChem loses molecules (under $0.004\%$); the rest are fully conformant.

The corpora span $\sim$740K (COCONUT, total) to $\sim$136M molecules
(Table~\ref{tab:corpora});
each carries a deterministic held-out test split, used
for every held-out-evaluated measurement, leaving COCONUT $\sim$702K molecules
to train on.

\subsection{Tokenizer training}\label{ssec:training}

Both algorithms train from Smirk's $165$-token base behind a shared
pre-tokenization module, so each consumes an identical stream of glyph-id words.
The BPE arm is Smirk's \texttt{GpeTrainer}; the Unigram-LM arm is a sibling
trainer we add that delegates to HuggingFace \texttt{tokenizers}'
\texttt{UnigramTrainer} \citep{kudo2018subword-reg, moi2023tokenizers}. Both run at their reference
defaults (with three by-design exceptions, none of them tuning,
Appendix~\ref{app:training}), so neither is hand-tuned to favor the contrast.

The base is installed as length-1 pieces, so both arms target the same $V$
(Table~\ref{tab:hyperparams}). A
\emph{matched} condition matches this \emph{target} $V$; the arms need not
realize it identically. BPE keeps the full base and fills the remaining budget
with merges; Unigram-LM seeds a large pool and prunes back, halting at or below
target. The two realize near-equal sizes on diverse corpora, but on a narrow alphabet at
larger $V$ the Unigram-LM arm falls well below target. Realized per-arm sizes are
reported per condition (Table~\ref{tab:results-realized-vocab}; effect on overlap
in \S\ref{ssec:r-jaccard}). Because
several measurements carry no confidence interval, every tokenizer is also
trained twice and asserted byte-identical.

The \textbf{boundary policy} is the one knob we vary on that shared
pre-tokenization module: under
\emph{no-merge-brackets} (NMB) a bracketed atom is opaque and merges cannot cross
it, while under \emph{merge-brackets} (MB) the boundary is permeable. MB is Smirk-GPE's default; \citet{wadell2026smirk} also
trained an NMB variant, reporting ``similar results'' at their single
merge-exhausted vocabulary, a finding we test systematically across the
small-$V$ regime.

\subsection{What we measure}\label{ssec:measurements}

We compute seven structural quantities per condition: some from the realized
vocabularies, the dead-zone audit from the training corpus, and the rest on the
single per-corpus held-out test split. The first three are the
\emph{direct cross-arm contrasts}, one per sense of a ``different vocabulary''
(the remaining four are mechanism diagnostics, gated by the learnability bar and
defined in \S\ref{ssec:vregime}):

\begin{itemize}
  \item \textbf{Membership: vocabulary overlap (Jaccard).} The cross-arm
    overlap on the multi-glyph subword set is
    \begin{equation}\label{eq:jaccard}
      J = \frac{|\mathcal{V}^{\mathrm{multi}}_{\text{BPE}} \cap \mathcal{V}^{\mathrm{multi}}_{\text{UL}}|}{|\mathcal{V}^{\mathrm{multi}}_{\text{BPE}} \cup \mathcal{V}^{\mathrm{multi}}_{\text{UL}}|},
    \end{equation}
    writing $\mathcal{V}^{\mathrm{multi}}_a$ for the multi-glyph pieces
    (length $\ge 2$ glyphs) arm $a$ learns above the shared base, as distinct
    from the full vocabulary $\mathcal{V}_a$ and the target size $V$ (a scalar).
    We exclude the shared base, which both arms retain in full and which would
    otherwise dominate the overlap at small $V$ ($\approx 0.5$ at $V{=}256$)
    while saying little about what each algorithm selects.
  \item \textbf{Granularity: fertility.} The standard subwords-per-word
    tokenizer metric \citep{rust2021how}, here mean held-out tokens per
    molecule, with glyphs per token as the compression ratio (both tabulated per
    condition with CIs in Appendix~\ref{app:fertility}); we report the absolute gap
    $|\Delta f|$ and, as the headline contrast, the relative gap
    $\mathrm{rel}|\Delta f|$,
    \begin{equation}\label{eq:fertility}
      |\Delta f| = |f_{\text{BPE}}-f_{\text{UL}}|, \qquad
      \mathrm{rel}|\Delta f| = \frac{|\Delta f|}{\bar{f}}, \qquad
      \bar{f} = \tfrac{1}{2}(f_{\text{BPE}}+f_{\text{UL}}).
    \end{equation}
  \item \textbf{Distribution: token-frequency imbalance.} The divergence from
    uniform \citep{gowda2020optimal-vocab-nmt} is
    \begin{equation}\label{eq:imbalance}
      D = \tfrac{1}{2}\sum_i \big|p_i - 1/V\big|,
    \end{equation}
    with $p_i$ the held-out relative frequency of token $i$; the cross-arm
    contrast is the gap $|\Delta D| = |D_{\text{BPE}}-D_{\text{UL}}|$, and we
    report $D$ with
    normalized Shannon entropy $\eta = H/\log V$ and R\'enyi efficiency at
    $\alpha{=}2.5$ \citep{zouhar2023noiseless-channel} as within-family
    diagnostics \citep{schmidt2024tokenization-more-than-compression,
    cognetta2024two-counterexamples} (per-condition values in
    Appendix~\ref{app:distribution}).
\end{itemize}

\paragraph{Membership robustness.}
Three companions test whether this membership overlap is an artifact of
counting, crossing two axes into a $2\times2$: \emph{weighting} (unweighted
vs.\ frequency-weighted) by \emph{masking} (all multi-glyph pieces vs.\
structural pieces only). The frequency-weighted variant weights each subword
$x$ by its held-out emission count (normalized per arm, with a bootstrap CI),
reflecting the pieces a model actually sees:
\begin{equation}\label{eq:jw}
  J_{\mathrm{w}} = \frac{\sum_x \min\!\big(w_{\text{BPE}}(x), w_{\text{UL}}(x)\big)}
                        {\sum_x \max\!\big(w_{\text{BPE}}(x), w_{\text{UL}}(x)\big)}.
\end{equation}
The structural variants $J_{\mathrm{struct}}$ and $J_{\mathrm{w,struct}}$
recompute $J$ and $J_{\mathrm{w}}$ over \emph{structural} pieces only (dropping
bracket-internal pieces, those that only ever occur inside a bracketed atom),
isolating cross-pretoken merges; together they separate a low $J_{\mathrm{w}}$
into genuine high-frequency structural disagreement versus an artifact of
Unigram-LM's bracket-internal pieces
(Appendix~\ref{app:tables} gives the bracket-internal definition and the
structural renormalization).

\paragraph{Granularity, read positionally.}
Because both arms segment the \emph{same} glyph stream, both draw their cuts
from the same inter-glyph positions, so beyond \emph{how many} tokens
each emits we can ask \emph{where} its cuts fall relative to the other arm's. At
each position the two arms either agree to cut, agree to merge, \emph{nest}
(Unigram-LM cuts where BPE merges), or \emph{conflict} (BPE cuts where
Unigram-LM merges). We summarize this as the boundary Jaccard $J_{\partial}$
(agreement over cut positions, so agree-cut over agree-cut $+$ nest $+$
conflict), the nest and conflict rates (over all positions), the fraction of
molecules with zero conflict (those whose BPE parse is a strict coarsening of
Unigram-LM's), and, for the conflicts that remain, their localization by
substructure class (heteroatom, unsaturated carbon, saturated carbon). This adds
no new quantity: the nest rate is the fertility gap counted cut by cut.
Per-condition values are in
Appendix~\ref{app:nestedness}.

\subsection{Learnability gate and mechanism diagnostics}\label{ssec:vregime}

The comparison is posed where token embeddings are learnable, so we certify
that regime per condition. For each arm we compute the clearance $c_n$, the
fraction of its \emph{learned} vocabulary pieces (those above the shared base) firing
at least $n$ times in the training corpus, and apply the learnability bar
$F_{p,n}\!: c_n \ge p$, with $F_{95\%,100}$ ($c_{100} \ge 0.95$) the headline. We
apply \citet{gowda2020optimal-vocab-nmt}'s bar to the learned vocabulary because
the fixed $165$-token base is identical across both arms and chosen by neither
algorithm. This
is an undertrained-token audit, the training-corpus analog of the
weight-based audits on deployed LLMs \citep{land2024fishing} and one of the
low-cost tokenizer diagnostics \citet{alqahtani2026stop} advocate, applied here
to chemistry: it flags pieces that fire too rarely in training to learn a
reliable embedding (Appendix~\ref{app:learnability}; the per-arm $c_n$ sweep is
in Appendix~\ref{app:mechanism}, from which the bar at any $p$ follows). A
condition whose arm fails the bar is flagged and never pooled into a
cross-corpus reading. Because the bar is an absolute count, a failure indexes
corpus size, not typology. The same size/typology confound colors COCONUT's one
corpus-dependent contrast, the distribution gap, which the size-matched probe
(\S\ref{ssec:r-extras}) resolves as typological. The bar gates only the rare-token-tail diagnostics, not
the overlap and fertility contrasts; its clearance $c_n$ also supplies the
dead-zone surplus that opens the diagnostics below.

\noindent The remaining four are \emph{mechanism diagnostics}:

\begin{itemize}
  \item \textbf{Dead-zone surplus.} The cross-arm difference in clearance at the
    bar's count $n$, $\Delta c_n = c^{\text{BPE}}_n - c^{\text{UL}}_n$, the
    chemistry analog of the dead-zone surplus of
    \citet{bostrom2020bpe-suboptimal}.
  \item \textbf{Whole-pretoken absorption.} The fraction of pretokens
    absorbed as a single token \citep{reddy2025diminishing-tokenization}.
  \item \textbf{Scaffold fraction.} The BPE-side fraction of
    intermediate-merge ``scaffold'' tokens \citep{lian2024scaffold-bpe}, which we
    identify on the trained vocabulary as pieces whose end-of-training
    standalone frequency falls below the last committed merge's candidate
    frequency (a static analog of their per-iteration removal rule, hence a
    conservative lower bound);
    zero for Unigram-LM by construction.
  \item \textbf{Segmentation entropy.} Mean Shannon entropy of the fitted
    Unigram-LM's distribution over a held-out molecule's segmentations
    \citep{kudo2018subword-reg}, normalized per glyph (total entropy over total
    glyphs, the reported $\Delta H_g$ gap); zero for BPE by construction.
\end{itemize}

\paragraph{Quantifying and reading the measurements.}
Bootstrap CIs (95\%, $1000$ resamples, molecule as the resample unit)
accompany the held-out-evaluated quantities; the exact-set quantities
($J$, $J_{\mathrm{struct}}$, scaffold fraction) carry no CI. We read each
measurement as an effect size against the scale of measurement noise, not
against a pass/fail threshold: for membership, the cross-arm overlap is scaled
against the internal self-overlap ceiling (\S\ref{ssec:r-jaccard}) and the
prior-work overlap values we cite (\S\ref{ssec:algo}), never gated on a fixed
value. The one measured threshold we apply is on the distribution sense, whose gap
is small enough to check against the corpus-draw noise floor estimated from the
subsample redraws (\S\ref{ssec:r-extras}).

\subsection{Sensitivity analysis and structural probes}\label{ssec:sensitivity}

Beyond the headline grid we run a \textbf{sensitivity analysis}: each training
hyperparameter is swept one-factor-at-a-time across a ladder bracketing its
reference default (Table~\ref{tab:hyperparams}), with four
pairwise interactions and four structural probes
(Appendix~\ref{app:sensitivity}). Because BPE imposes no piece-length cap, its
only free training knob is the merge frequency, so the sweep is necessarily
Unigram-LM-weighted. To stay tractable the battery runs on a fixed representative
subsample while the headline grid stays full-corpus.

\section{Results}\label{sec:results}

We establish the divergence across the three direct contrasts
(\S\ref{ssec:r-jaccard}), show it stable across the grid
(\S\ref{ssec:r-stability}) and robust under hyperparameter and structural
probes (\S\ref{ssec:r-extras}), trace its mechanism (\S\ref{ssec:r-mechanism}),
and find the granularity gap generalizes off-domain
(\S\ref{ssec:r-generalization}). All numbers below are generated from the
deposited per-condition measurements by the accompanying scripts. Our analysis
covers the $22$ matched conditions of \S\ref{ssec:grid} (both arms trained); the
two single-arm ZINC-22 $V{=}2048$ coordinates are excluded, the BPE arm there
clearing only $c_{100} \approx 0.52$ (below the $F_{95\%,100}$ bar) and the Unigram-LM arm, unsafe by a wider
margin, left untrained.

Table~\ref{tab:results-seven} collects all seven measurements across the $22$
conditions, grouped into the three direct contrasts and the four mechanism
diagnostics; the subsections below read its columns in turn, the contrasts in
\S\ref{ssec:r-jaccard} and the diagnostics in \S\ref{ssec:r-mechanism}, with the
full per-condition detail and bootstrap CIs in Appendix~\ref{app:tables}.

\input{tables/seven_measurements}

\subsection{The two algorithms build measurably different
vocabularies}\label{ssec:r-jaccard}

We report in turn the three senses in which the vocabularies differ (membership,
granularity, distribution). They carry different weight: membership is the more
\emph{surprising} contrast, the one a convergence would show up in first;
granularity the more \emph{robust and actionable}, carrying a bootstrap CI and
untouched by the realized-size asymmetry; and distribution the weakest. Read
positionally, granularity yields a fourth result that is not a fourth sense of
difference but an answer to a question the first two raise: the near-disjoint
vocabularies nonetheless parse \emph{compatibly} (\emph{Compatibility}, below).

\paragraph{Membership.}

\begin{figure}[htbp]
  \centering
  \includegraphics[width=\linewidth]{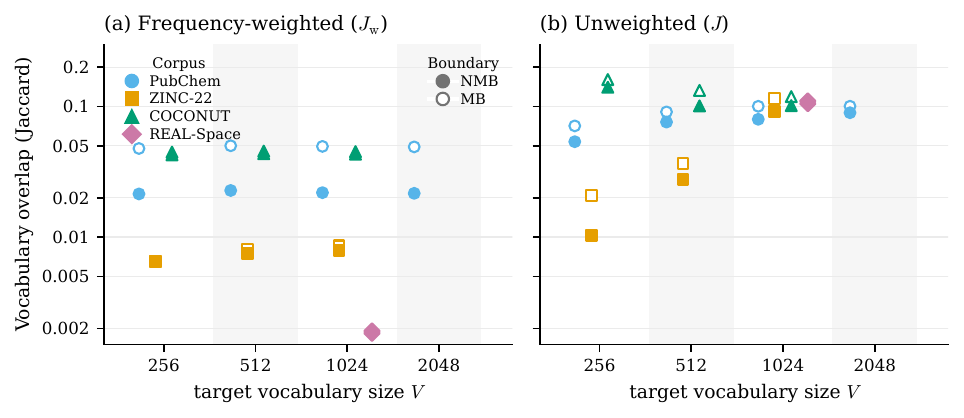}
  \caption{Cross-algorithm vocabulary overlap for $22$ matched
  conditions, frequency-weighted ($J_{\mathrm{w}}$, left) and unweighted ($J$,
  right) on a shared log axis; color and marker by corpus, filled NMB / open
  MB, with target $V$ on the $x$-axis. Every
  condition stays near-disjoint, and $J_{\mathrm{w}} < J$ throughout. Structural
  variants in Fig.~\ref{fig:overlap-struct}; bootstrap CIs on $J_{\mathrm{w}}$ are
  smaller than the markers.}
  \label{fig:overlap}
\end{figure}

\begin{figure}[htbp]
  \centering
  \includegraphics[width=\linewidth]{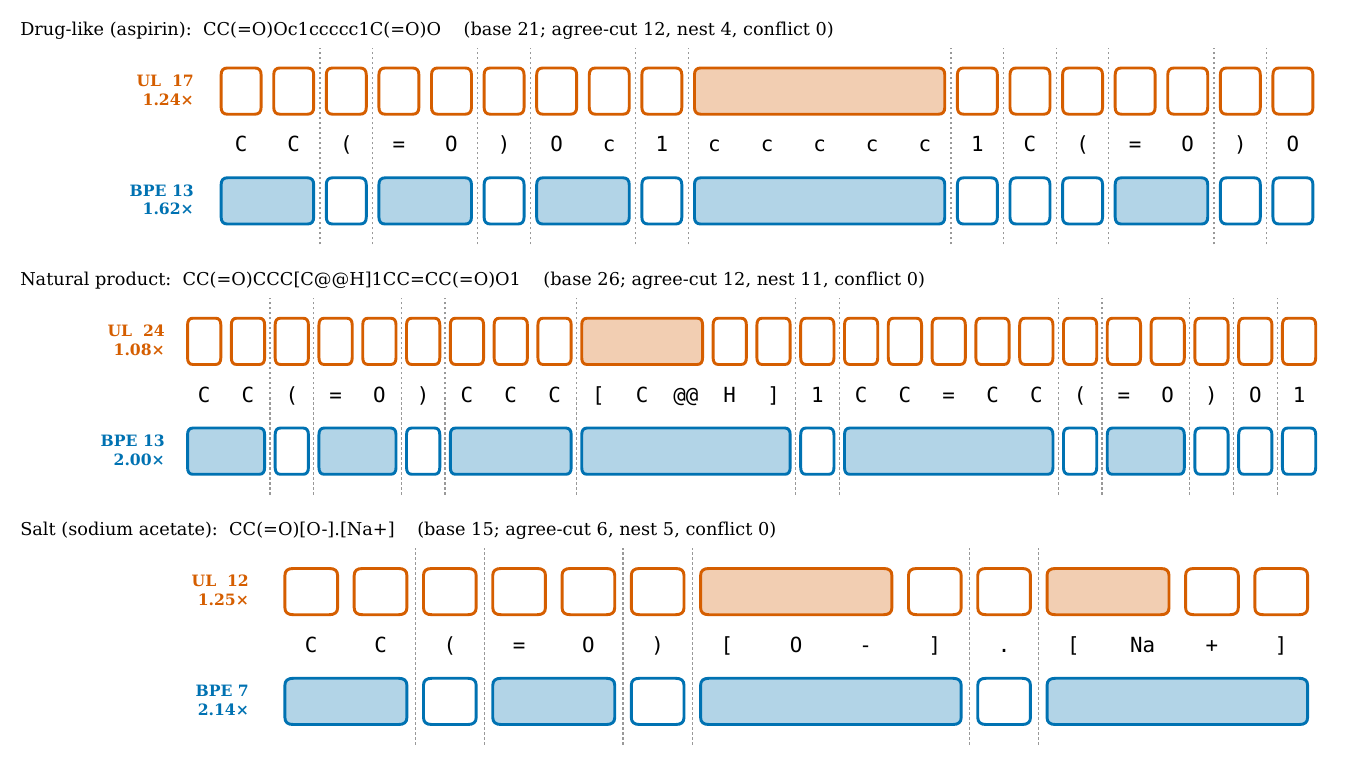}
  \caption{Segmentation and nesting on three held-out molecules under the
  matched PubChem $V{=}1024$ (MB) pair, the per-molecule face of all three
  contrasts. Each panel draws the shared glyph stream with Unigram-LM tokens as
  boxes above (orange) and BPE below (blue); a box is \emph{filled} when it is a
  learned multi-glyph piece and \emph{hollow} when it is a single base glyph, and
  dashed guides mark BPE cuts. \textbf{Membership:} the two arms make different
  substructures atomic: Unigram-LM's pieces here are \texttt{ccccc},
  \texttt{[C@@}, \texttt{[O-}, \texttt{[Na}, BPE's are
  \texttt{CC}, \texttt{=O}, \texttt{[C@@H]}, \texttt{CC=CC}. \textbf{Granularity:}
  per-row token counts and compression (glyphs/token) give the fertility gap per
  molecule. \textbf{Compatibility:} every BPE box spans a run of Unigram-LM boxes
  (\emph{nest}) with no \emph{conflict} (neither arm cutting where the other
  merges across it), so BPE's parse is a strict coarsening of Unigram-LM's, the
  $96$--$99\%$ case across the $V{=}1024$ cells (Table~\ref{tab:results-nestedness}).}
  \label{fig:segmentation}
\end{figure}

Given the same glyph base and corpus, BPE and Unigram-LM build
vocabularies that barely overlap, and they overlap least on the
highest-frequency pieces. The frequency-weighted overlap $J_{\mathrm{w}}$ (Eq.~\ref{eq:jw}) is at
most $0.05$ and falls to $0.002$; the unweighted overlap $J$ never exceeds $0.161$ and falls
as low as $0.010$ (Table~\ref{tab:results-jaccards}, Fig.~\ref{fig:overlap};
per-condition split in Fig.~\ref{fig:membership}).
Figure~\ref{fig:segmentation} makes this concrete on three molecules
segmented by the same PubChem $V{=}1024$ (MB) tokenizers: BPE forms multi-glyph
chunks (\texttt{CC}, \texttt{CC=CC}) and keeps the stereocenter
\texttt{[C@@H]} whole, whereas Unigram-LM stays near-atomic, even fragmenting
\texttt{[C@@H]} into \texttt{[C@@}, \texttt{H}, \texttt{]}. The two partition the
identical glyph stream into almost entirely different units.

The right scale for this overlap is internal:
the \emph{same} algorithm retrained on the same corpus reproduces its vocabulary
exactly (self-overlap $J{=}1.0$; the determinism check makes every cell
byte-identical on retrain, Appendix~\ref{app:training}), and $J$ moves at most
$0.005$ across independent subsample redraws (\S\ref{ssec:r-extras}). Against that
near-unity ceiling the cross-arm $0.010$--$0.161$ is a qualitatively different
regime, and the disagreement is sharpest exactly where the convergence null
predicted agreement. $J_{\mathrm{w}}$ falls \emph{below} the unweighted $J$ in all $22$
conditions ($0.021$ vs.\ $0.054$ on PubChem $V{=}256$ NMB; $0.002$ vs.\
$0.105$ on REAL-Space): the pieces the two arms \emph{share} are
disproportionately rare, so they disagree most on the high-frequency core. This
refutes the convergence null of \S\ref{sec:intro}: weighting by
the mass a model sees does not soften the divergence but deepens it.

The two arms disagree not just on \emph{how many} pieces they share but on
\emph{what kind} of substructure each makes atomic
(Table~\ref{tab:composition}). Aromatic-ring
pieces (\texttt{cccc}, \texttt{ccccc}) make up $27$--$37\%$ of BPE's exclusive
pieces across the three corpora but only $0$--$7\%$ of Unigram-LM's, which
instead skews to aliphatic heteroatom chains (up to $92\%$ of its exclusive set
on ZINC-22): BPE greedily merges ring fragments into single tokens while
Unigram-LM keeps them near-atomic, the population-level form of the segmentation
example. This BPE-side ring-merging is also visible in large-$V$ chemistry
vocabulary extension \citep{kalamkar2025tokenization}, where a BPE-style
tokenizer likewise captures aromatic rings as single tokens. The same split is visible
at the level of glyph adjacencies within pieces
(Fig.~\ref{fig:glyph-cooccurrence}): Unigram-LM's pieces run longer on average and
skew to sulfur and oxygen chains, while BPE reaches further into the
aromatic--aliphatic junction.

That disagreement on the high-frequency core reflects genuine chemical
substructure, not an artifact of how bracketed atoms are split: restricting the overlap to structural pieces
barely moves it in either weighting ($J_{\mathrm{w,struct}}$ exceeds
$J_{\mathrm{w}}$ by at most $0.026$, $J_{\mathrm{struct}}$ never exceeds $0.178$,
both still near-disjoint; Table~\ref{tab:results-jaccards}). The
bracket-internal component is largest on the scaffold-rich COCONUT and negligible
elsewhere, yet even there the structural disagreement dominates.

On narrow alphabets the two arms diverge for a second, structural reason: they do
not even build the same-size vocabulary. At $V{=}1024$ the Unigram-LM arm saturates
well below target (ZINC-22 $310$--$357$ vs.\ BPE's $867$ multi-glyph pieces,
REAL-Space $267$--$293$ vs.\ $867$), its narrow alphabet offering too few
high-likelihood pieces for the pruning to reach the budget. This mechanically
caps the \emph{unweighted} overlap, but the weighted headline is immune (a piece
an arm never realizes carries no emission mass), so $J_{\mathrm{w}}$ stays at
$0.002$--$0.009$ on these cells (ceilings in Appendix~\ref{app:narrow}).

\paragraph{Granularity.}

\begin{figure}[htbp]
  \centering
  \includegraphics[width=\linewidth]{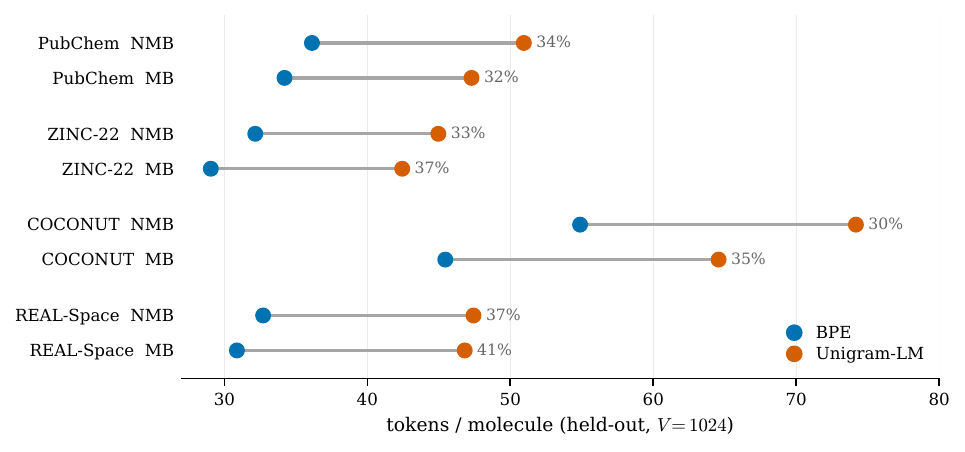}
  \caption{Held-out fertility at $V{=}1024$ as a dumbbell chart: each row joins
  BPE (blue) to Unigram-LM (orange) on a shared tokens-per-molecule axis, so
  connector length is the absolute fertility gap, comparable across corpora.
  Unigram-LM lies to the right of BPE (more tokens, less compression) in every
  corpus and under both boundary policies (NMB, permeable MB), with the relative
  gap rel$|\Delta f|$ (Eq.~\ref{eq:fertility}) annotated per row ($30$--$41\%$).
  The absolute gap is largest on the natural-products corpus (COCONUT), whose
  molecules are longest. The relative gap's rise with
  $V$ is in Appendix Fig.~\ref{fig:cross-v}; per-condition values across all $V$, with
  bootstrap CIs, are in Table~\ref{tab:results-fertility}.}
  \label{fig:fertility-curves}
\end{figure}

The vocabularies do not just differ in membership; they segment molecules
differently, and consistently so. Unigram-LM emits \emph{more} tokens per
held-out molecule than BPE at matched $V$ in every condition
(Figure~\ref{fig:fertility-curves}), with a
relative fertility gap of $29.2$--$41.0\%$ (the $41.0\%$ maximum on the
REAL-Space anchor; Table~\ref{tab:results-seven}, column rel$|\Delta f|$). In absolute terms the gap is large: on
PubChem $V{=}1024$ (NMB) the average held-out molecule is $36.1$ tokens under BPE
against $50.9$ under Unigram-LM (per-arm absolute fertilities, glyphs-per-token
compression ratios, and bootstrap CIs for every condition in
Appendix~\ref{app:fertility}, Table~\ref{tab:results-fertility}). Seen as
compression the same gap is corpus-stable where absolute length is not: BPE packs
$\sim$$1.4$--$1.8$ glyphs per token across all three corpora while Unigram-LM
stays near-atomic at $\sim$$1.0$--$1.2$, barely above the one-glyph-per-token
floor.

Unlike membership, granularity is unaffected by the realized-size asymmetry: it
is read from held-out segmentation, so the unequal
vocabulary sizes that cap the unweighted overlap on narrow alphabets leave it
untouched even at those cells. It is thus the contrast that most cleanly
separates the two algorithms everywhere. Beyond the training grid, this gap also
\emph{transfers}: it survives reuse
off-domain and on chemistry far outside the training distribution
(\S\ref{ssec:r-generalization}).

\paragraph{Compatibility: the near-disjoint vocabularies nonetheless parse
compatibly.}
The near-disjoint membership and the fertility gap together raise a question they
do not answer: do the two arms cut molecules in fundamentally incompatible places,
or in the same places to different \emph{depths}? Because both segment the same
glyph stream, we can read the answer position by position
(\S\ref{ssec:measurements}, Figure~\ref{fig:segmentation}). The gap is not merely a
difference in count; it is \emph{ordered}. The arms agree on most cuts (the
boundary Jaccard $J_{\partial}$ is $0.65$--$0.74$ across the grid) and nearly all
of their disagreement is \emph{nesting} rather than crossing: Unigram-LM cuts where
BPE merges (the nest rate, $0.24$--$0.34$, is the fertility gap read positionally)
while genuine crossing conflict stays near zero, below $0.7\%$ of positions in
every condition and at most $0.1\%$ once $V{\ge}1024$. The molecule-level
consequence is sharp: on PubChem $V{=}1024$ (NMB) the BPE segmentation is a strict
coarsening of Unigram-LM's on $97.0\%$ of held-out molecules, rising above $99\%$
on the $V{=}1024$ ZINC-22 and REAL-Space cells, and never dropping below $80\%$
even at $V{=}256$ (Table~\ref{tab:results-nestedness}). This resolves the apparent
paradox: the two algorithms build near-disjoint \emph{vocabularies} yet impose
near-compatible \emph{parses}. BPE lays down essentially the cut skeleton Unigram-LM does
and then merges further along it. What little conflict remains is chemically
structured, falling on heteroatom and unsaturated pieces rather than the
saturated-alkyl backbone, on the diverse and natural-products corpora (the
$\mathrm{cut}^{c}$ columns of Table~\ref{tab:results-nestedness}).

\paragraph{Distribution.}
The third sense, token-frequency distribution, separates the two algorithms in
the same direction but by a smaller margin. The token-imbalance gap
$|\Delta D|$ exceeds the corpus-draw noise floor ($\sim$0.002, the one measured
threshold we apply; \S\ref{ssec:r-extras}) in every condition, the smallest
gap ($0.038$, COCONUT $V{=}1024$) still sitting an order of magnitude above it
(Table~\ref{tab:results-seven}, column $|\Delta D|$).
Distribution is corpus-sensitive only in the \emph{magnitude} of the
gap, not whether it separates, a typological attenuation we examine in
\S\ref{ssec:r-stability}.

\subsection{The divergence is stable across corpus,
boundary, and scale}\label{ssec:r-stability}

\begin{figure}[htbp]
  \centering
  \includegraphics[width=\linewidth]{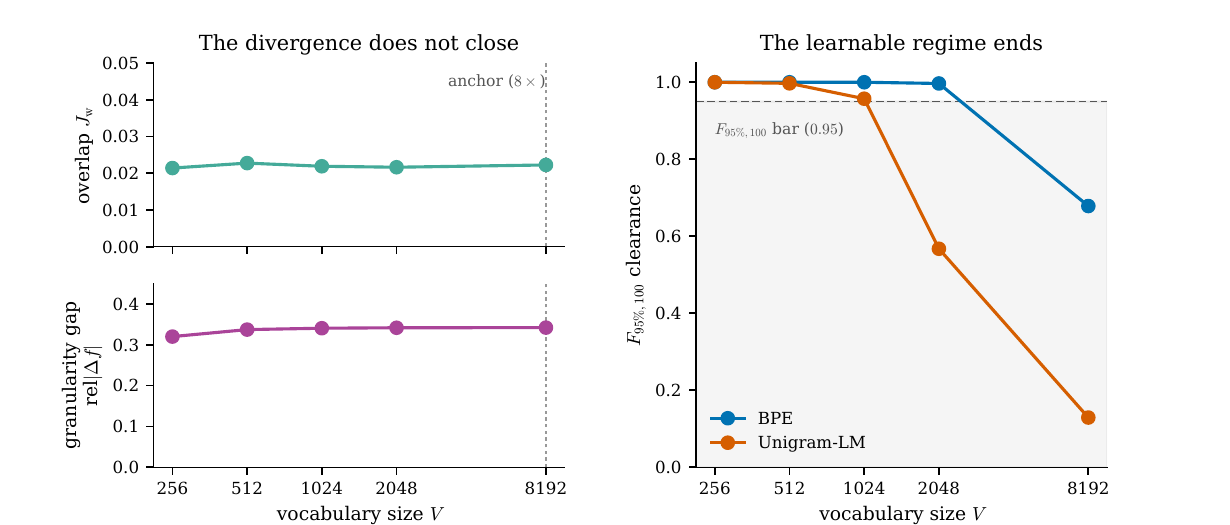}
  \caption{Scale does not dissolve the divergence, and could not be pushed much
  further if it did. Both panels track PubChem (NMB) across the headline grid and
  the $V{=}8192$ anchor on a log-$V$ axis. \textbf{Left:} the frequency-weighted
  overlap $J_{\mathrm{w}}$ and the relative granularity gap
  $\mathrm{rel}|\Delta f|$ run flat through the anchor, $8\times$ the headline
  vocabulary, so neither contrast converges. \textbf{Right:} the $F_{95\%,100}$
  learnability clearance for each arm, with the $0.95$ bar (shaded region below is
  the undertrained regime). The knee is sharp: the Unigram-LM arm crosses the bar
  by $V{=}2048$ and BPE by $V{=}8192$, so the anchor already sits past where the
  embeddings are learnable. Per-arm clearances across the grid are in
  Table~\ref{tab:results-nsweep}, with the $V{=}8192$ anchor in
  Appendix~\ref{app:learnability}.}
  \label{fig:scale}
\end{figure}

The divergence is stable along all three axes. Across corpus typology
(Table~\ref{tab:results-seven}), membership and granularity separate the two
algorithms in all six headline $(V, \text{boundary})$ conditions; across
boundary policy no contrast's outcome flips between the opaque-bracket (NMB)
and permeable-bracket (MB) regimes (Appendix Fig.~\ref{fig:interaction}); and
across scale the fertility gap rises only slightly with $V$ while overlap stays
near-disjoint throughout, rising on PubChem and ZINC-22 and
falling on COCONUT (Appendix Fig.~\ref{fig:cross-v}). That overlap rise is
one-directional, BPE's later merges absorbing substructures Unigram-LM had
already isolated at lower $V$ (marginal cross-arm Jaccard $\le 0.016$); COCONUT
instead starts high, its dominant natural-product motifs shared early and then
diluted as the near-disjoint vocabulary grows.

To test whether the divergence is a small-$V$ artifact, we train one matched
pair far above the grid, at $V{=}8192$ on PubChem (NMB), where the wide alphabet
lets \emph{both} arms fill to target (no Unigram saturation), so the comparison
is clean. It does not close: the overlaps $J{=}0.083$ and
$J_{\mathrm{w}}{=}0.022$ barely move from their $V{=}1024$ values of $0.080$ and
$0.022$, with $J$ shifting $0.003$ (inside the $\lesssim 0.005$ subsample-redraw
band, \S\ref{ssec:r-extras}) and $J_{\mathrm{w}}$ by only $0.0004$, and Unigram-LM still segments $34.2\%$ more
finely, on a flat $32$--$34\%$ trend from $V{=}256$ through $V{=}8192$. Whatever
large-$V$ convergence the natural-language literature reports
(\S\ref{sec:discussion}) therefore lies past the entire learnable range, not
just the small-$V$ regime we headline (Figure~\ref{fig:scale}).

The one corpus-sensitive feature is the \emph{magnitude} of the distribution
gap: it stays above the noise floor in every condition but on the
natural-products corpus runs about half that on the diverse and drug-like
corpora at $V{=}1024$ ($|\Delta D| = 0.038$ vs.\ $\sim$0.07), a size-matched
probe resolving that attenuation as typological, not a small-sample artifact
(\S\ref{ssec:r-extras}). Beyond the gap, the absolute intrinsics show a coherent
within-family signature: both arms are highly imbalanced ($D \approx
0.76$--$0.97$), but BPE is the more uniform of the two in all $22$ conditions
(lower $D$, higher normalized entropy and R\'enyi efficiency;
Appendix~\ref{app:distribution}, Fig.~\ref{fig:distribution-intrinsics}).

\subsection{Robustness of the contrasts}\label{ssec:r-extras}

Robustness is assessed by a full sensitivity battery, not isolated
spot-checks (\S\ref{ssec:sensitivity}, Appendix~\ref{app:sensitivity},
Figs.~\ref{fig:sensitivity-curves}--\ref{fig:interaction}): the separation
survives every rung of every ladder and every interaction surface by a wide
margin.

\paragraph{Hyperparameter sweeps.}
Read as response curves, the two headline contrasts hold on every rung:
overlap stays near-disjoint ($J \le 0.123$, within the $\le 0.161$ headline
ceiling) and the fertility gap stays large ($\mathrm{rel}|\Delta f| \ge 23\%$),
both on the fixed PubChem $V{=}512$ NMB subsample, with the seed pool and BPE
merge frequency inert. The most overlap-favorable hyperparameter corner we could construct is the
shortest-piece ($L{=}4$) setting on the narrow-alphabet corpora, where minimal
Unigram-LM pieces most resemble BPE's merges. Even there $J$ peaks at only $0.29$
(Appendix Fig.~\ref{fig:interactions}), above the
headline $\le 0.161$ range but still leaving the two vocabularies more than
$70\%$ disjoint. One rung is a
caveat rather than a contrast: the shrinking factor never moves the cross-arm
separation, but read against the default schedule it is the one knob that
perturbs the Unigram vocabulary itself, which we revisit in
\S\ref{sec:discussion}.

\paragraph{Structural probes.}
These hold the algorithm knobs fixed and complete the picture.
Across three independent subsample redraws the cross-arm contrasts barely move
(overlap spread $\lesssim 0.005$, imbalance-gap spread $\lesssim 0.0017$); the
latter sets the corpus-draw noise floor for the distribution sense, which every
condition's $|\Delta D| \gtrsim 0.038$ clears by more than an order of
magnitude. A within-PubChem size sweep locates the marginal PubChem $V{=}512$
NMB case as size-driven rather than typological, and the REAL-Space
merge-exhaustion anchor confirms the continuity premise (Smirk's
\texttt{GpeTrainer} terminating naturally at $|\mathcal{V}| = 4{,}331$ on
Enamine's published REAL 1\% Sample, placing the small-$V$ regime on the same
merge trajectory as prior large-$V$ work). Finally, a \textbf{size-matched
typology} probe resolves the size$\times$typology confound on COCONUT
(\S\ref{ssec:vregime}) and the distribution-magnitude attenuation of
\S\ref{ssec:r-stability}: subsampling PubChem and ZINC-22 to COCONUT's
$\sim$702K training size, the token-imbalance gap stays near $0.06$ on both
non-natural-products corpora while COCONUT at the same size reads roughly
two-thirds that ($\sim$$0.038$), so the smaller natural-products gap is a typology effect,
not a small-sample artifact. At this reduced size the overlap rises on the
narrow-alphabet ZINC-22 to $J \approx 0.34$, above the full-corpus $\le 0.161$ headline but still leaving
the two vocabularies about two-thirds disjoint.

\subsection{How the two algorithms diverge}\label{ssec:r-mechanism}

\paragraph{Mechanism diagnostics.}
Table~\ref{tab:results-seven} collects all seven scalars per condition. Of the
four, three line up with the segmentation contrast and the
fourth, the dead-zone surplus, does not transfer. The substantive one is whole-pretoken absorption,
higher for BPE in every condition (the per-pretoken counterpart of the fertility
gap): BPE absorbs almost every pretoken as a single token ($0.89$--$1.00$, e.g.\
$0.97$ on PubChem $V{=}1024$ NMB) against Unigram-LM's $\sim$$0.72$
(Appendix~\ref{app:mechanism}). The other two are one-sided by construction and
so weaker: segmentation entropy is carried entirely by Unigram-LM, and the BPE
scaffold fraction is non-zero (Unigram-LM's is exactly zero) but faint at
these sizes (at most $0.6\%$). That is an order of magnitude below the $\sim$6\%
\citet{lian2024scaffold-bpe} report at $V{=}32$K, a gap that reflects both the
smaller $V$ and our conservative static criterion (\S\ref{ssec:vregime}); the
direction, a large-$V$ phenomenon, is unchanged.

\paragraph{The dead-zone surplus.}
If anything, it cuts against the natural-language account: where
\citet{bostrom2020bpe-suboptimal} found BPE carrying the larger dead zone, in
chemistry both arms clear the $F_{95\%,100}$ bar wherever the tail can be
certified, with negligible surplus ($|\Delta c_{100}| \le 0.043$, mixed sign;
Table~\ref{tab:results-delta-f}). The large surpluses occur only in conditions
too small to certify, and there the sign reverses, Unigram-LM stranding the
rarely-firing near-atomic pieces its top-down pruning keeps, so the asymmetry is
corpus-size-driven, not algorithmic. Bostrom's BPE-heavy surplus joins
morphological alignment as a natural-language leg that does not carry to
chemistry SMILES.

\paragraph{One mechanism, several views.}
The near-disjoint vocabularies, the consistently signed fertility gap, and the
absorption gap are not independent confirmations: absorption and fertility are
the per-pretoken and per-molecule faces of the same coarser BPE segmentation. The realized-size asymmetry is one
more view: on narrow alphabets at large $V$, Unigram-LM's pruning stops short
while BPE fills to target (\S\ref{ssec:r-jaccard}), the same preference for finer
pieces that drives the fertility gap and depresses the overlap.

\paragraph{A structural reading.}
The same picture emerges from structure alone: BPE's bottom-up merging makes its vocabulary
\emph{compositionally closed} (every piece concatenates two it already holds),
whereas Unigram-LM's likelihood pruning leaves roughly half its pieces unable to
decompose into in-vocab parts, the same long heteroatom chains
it keeps exclusively. Read on real molecules, BPE keeps the multiply-bonded
heteroatoms that define functional groups (carbonyl $=$O, nitrile $\#$N) inside a
single token almost always ($0.95$--$1.00$), where Unigram-LM almost never does
($\le 0.03$). Five appendix analyses, piece composition, length, glyph
co-occurrence, compositional closure, and functional-bond locality
(Appendix~\ref{app:structural}), all resolve this same
greedy-versus-probabilistic signature.

\subsection{Generalization: the fertility gap is a property of the
algorithm pair, not the corpus}\label{ssec:r-generalization}

\input{tables/transfer_matrix}

Sections~\ref{ssec:r-jaccard}--\ref{ssec:r-stability} established the granularity
gap and its stability across the training grid. We now show it is a property of
the \emph{algorithm pair} itself rather than of any particular training corpus:
it persists when a tokenizer is read on a corpus it was never fit to
(cross-domain transfer) and even on chemistry far outside its training
distribution (out-of-distribution), while coverage stays a non-issue throughout.
This is a granularity-specific check by construction: a vocabulary's
membership overlap is fixed at training and cannot be re-probed on new corpora,
whereas fertility is re-measured wherever the tokenizer is read. Reading the
trained pairs on non-canonical rewrites of their own held-out split closes the
section, yielding both a final robustness check on the gap and a new contrast in
its own right, \emph{write-stability}.

\paragraph{Cross-domain transfer.}
The near-disjoint vocabularies raise a practical question: is a tokenizer fit
to one corpus usable on another, or must each domain retrain? We read every
$V{=}1024$ (NMB) tokenizer on the held-out split of each \emph{other} corpus
(Table~\ref{tab:transfer}, off-diagonal; the diagonal native baseline is each
corpus's own-tokenizer fertility, set to $1.00$). Two findings. First, coverage
is a non-issue: the shared $165$-token base keeps atom-level out-of-vocabulary
below $0.01\%$ in every cell. Unlike natural-language tokenizers, chemical ones
do not shatter off-domain input into unknowns, because the atomic alphabet is
small and nearly universal. Second, the
off-domain \emph{fertility} penalty is small and arm-dependent: BPE transfers
within $\pm 1\%$ of native everywhere, so its sequence length is effectively
corpus-agnostic, whereas Unigram-LM is modestly domain-sensitive
($-6\%$ to $+8\%$ of native), with the largest inflation exactly where the
combinatorial REAL-Space specialist meets the natural-products corpus
(REAL-Space$\to$COCONUT, $+8\%$). The algorithms diverge in \emph{which}
substructures they make atomic, not in how efficiently they tokenize off-domain
chemistry.

\input{tables/ood_eval}

\paragraph{Out-of-distribution chemistry.}
The transfer matrix stays within ordinary organics, so it never asks whether the
contrast holds on genuinely exotic chemistry. We test that directly by reading
the PubChem diverse-corpus generalist ($V{=}1024$, NMB, both arms) on two
adversarial in-spec corpora it was not fit to (Table~\ref{tab:ood-eval}):
tmQM \citep{balcells2020tmqm}, $\sim$97K transition-metal complexes whose rare
bracketed metals (\texttt{[Sc+3]}, \texttt{[Y+3]}, \texttt{[MoH2+2]}) probe the
coverage axis, and CycPeptMPDB \citep{li2023cycpeptmpdb}, $\sim$8K cyclic
peptides ($\sim$110--150 tokens each) whose length probes the fertility axis.
Coverage holds even here: atom-level OOV is negligible for both arms (zero on
CycPeptMPDB; $0.002\%$ of tmQM tokens, from a residual off-base aromatic silicon
that survives the dative-bond derivation documented in
Appendix~\ref{app:preproc}), so even transition-metal chemistry stays covered
under the shared $165$-token base. The granularity contrast \emph{survives
off-distribution}: Unigram-LM still segments more finely than BPE on chemistry
neither arm was trained on, by $29.5\%$ on the cyclic peptides, just inside
the $29.2$--$41.0\%$ in-distribution band, and by $23.5\%$ on the metal
complexes, more attenuated still, below that band. That attenuation is the expected one: on the rare
tokens neither arm learned a merge for, BPE falls back toward the shared
glyph base and Unigram-LM's already-near-atomic segmentation, so the two
partially converge, exactly the regime the metal corpus occupies. The fertility gap therefore belongs to the algorithm pair, not the training corpora.

\paragraph{Robustness to non-canonical input.}
The first two checks vary the chemistry; this one varies the \emph{notation}
instead. Reading the trained pairs on identity-preserving SMILES rewrites of their
held-out split (five rewrite families, Appendix~\ref{app:noncanon}) gives two
results. The gap \emph{survives}: on the randomized orbit Unigram-LM stays
$23$--$35\%$ finer than BPE in all $22$ conditions, the relative gap attenuating
only slightly off-canonical while both arms' absolute fertilities rise, so the
canonical fertilities lower-bound a deployed model's sequence length. The orbit also
separates the arms in a \emph{new} sense, \emph{write-stability}: Unigram-LM's
piece selection moves less under rewriting, its bag-instability (the fraction of
the token multiset that changes) running $0.11$--$0.22$ against BPE's
$0.24$--$0.38$ and holding the same ranking under an independent canonicalizer
(OpenBabel: $0.09$--$0.15$ vs.\ $0.19$--$0.24$), in all $22$ conditions.

\section{Discussion}\label{sec:discussion}

The structural difference that \citet{bostrom2020bpe-suboptimal} found
between BPE and Unigram-LM on natural language is present in chemistry
SMILES, and not marginally. At the vocabulary sizes where embeddings are
learnable, the two algorithms build near-disjoint vocabularies and segment
molecules to systematically different lengths, robustly across the diverse
$\to$ drug-like $\to$ natural-products range and under both boundary policies,
yet their cuts nest far more than they cross (Appendix~\ref{app:nestedness}). A small,
valence-constrained alphabet could plausibly have forced the two to converge.
That it does not, and that the divergence is algorithmic, corpus-stable, and
boundary-robust rather than an artifact of the confounds prior work left open,
is what the controlled design establishes. The one prior chemistry head-to-head,
\citet{temizer2024chemical-tokenization}'s Unigram-versus-others comparison on raw
ChEMBL, glimpsed the same contrast but at large $V$ without a fixed base; our
controlled design recovers it in a far more extreme form.

\textbf{Why the algorithm separates here when the field-wide picture
suggests it should not.} On the same three intrinsic metrics we use,
\citet{wadell2026smirk} found the whole field of chemistry tokenizers
scoring within a single $90\%$ quantile band, and argued that coverage,
not subword scheme, separates them downstream. Our result is the
controlled complement of theirs, not a contradiction: their survey spans
heterogeneous bases at each tokenizer's native, mostly large vocabulary,
where coverage dominates; once the base is fixed and the vocabulary pushed
into the small regime, the algorithm axis their survey could not isolate
emerges sharply on overlap and fertility. Consistent with their reading, our
\emph{weakest} contrast is precisely token imbalance (the metric on which they
report all schemes clustering near $D \approx 0.5$), and it is the one whose
margin shrinks most across typology, a typological attenuation that never
overturns the separation (\S\ref{ssec:r-extras}). Its monotonic contraction with
$V$ echoes the large-$V$ washout \citet{reddy2025diminishing-tokenization} report
for natural language, but only that weakest contrast does so: the $V{=}8192$
anchor leaves membership and granularity essentially unmoved from their
$V{=}1024$ values ($J_{\mathrm{w}}{=}0.022$, gap $34.0\%$).

\textbf{Why not a language-model-scale vocabulary?} The divergence is not an
artifact of staying small: chemistry's \emph{learnable} ceiling is low and the
$V{=}8192$ anchor already sits above it. The $F_{95\%,100}$ audit that gates the
grid shows the Unigram-LM arm crossing into the undertrained tail by $V{=}2048$
even on PubChem, and both arms unsafe by the anchor, a ceiling set by the
$165$-token base and consistent with deployed chemistry tokenizers staying small
(\S\ref{ssec:vregime}, Appendix~\ref{app:learnability}). Pushing a chemistry
vocabulary toward language-model scale would therefore not extend the present
comparison but dissolve it: both arms would fill with pieces too rare to train,
so any rise in overlap would reflect a shared exhaustion of the substructure
space rather than agreement on what to make atomic.

\textbf{What this means for chemical-LM builders.} The practical takeaway
is that the subword algorithm is a genuine modeling decision, not a neutral
default to inherit from natural-language tooling. Even with the atomic level
pinned by a fixed $165$-token base, swapping BPE for Unigram-LM changes which
substructures become atomic embedding units (near-disjoint vocabularies,
overlapping least on the high-frequency pieces a model updates most) and the
effective sequence length
by roughly a third (the fertility gap). BPE buys shorter sequences (cheaper
attention, longer context, and the NMT compression that tracks downstream
quality at a fixed vocabulary \citep{galle2019investigating}) at the cost of
coarser, less granular pieces; Unigram-LM retains finer pieces and supports
principled subword regularization \citep{kudo2018subword-reg}, where BPE has only
merge dropout \citep{provilkov2020bpe-dropout}. The training \emph{corpus}, by
contrast, matters little for efficiency: reused off-domain a tokenizer segments
at near-native fertility (\S\ref{ssec:r-generalization}), so the consequential
choice is the algorithm, not the corpus. Which arm is preferable downstream is
the comparison this study sets up rather than settles, the more so since
intrinsic separation need not translate into a downstream gap
\citep{ali2024tokenizer-choice}; the choice is nonetheless worth justifying.

\textbf{Limits of the mechanism we can claim.} We claim the
base-independent part of the mechanism: near-disjoint vocabularies, a
consistently signed fertility gap, higher whole-pretoken absorption for BPE, and
the one-sided segmentation-entropy signal, which together show the two algorithms
partition the same constrained glyph stream differently. We do \emph{not} claim
the morphological-alignment leg of the natural-language account (SMILES has no
gold-standard segmentation), nor that the gap's \emph{direction} is portable: on
protein sequences the same comparison finds Unigram-LM the \emph{more} compressive
arm \citep{suyunu2024linguistic}, so which algorithm segments more finely is
domain-specific. One caveat survives on the Unigram reading itself, the
prune-schedule sensitivity flagged in \S\ref{ssec:r-extras}, the one knob that
visibly moves its piece set.

\section{Limitations}\label{sec:limitations}

As a tokenizer-level study, this work does not measure the downstream
language-model quality a follow-on takes up, nor does it
exercise Unigram-LM's subword-regularization capability (\S\ref{sec:discussion}).

The headline contrast rests in part on two exact-set Jaccard quantities ($J$ and
$J_{\mathrm{struct}}$) that carry no per-condition error bar. Two variance
sources bound them. The determinism check measures the \emph{retrain} jitter
(the symmetric-difference piece count, same data retrained) at zero for every
cell, bounding it below $\sim$0.001 on the $J$ scale; the three subsample redraws
(\S\ref{ssec:r-extras}) bound the \emph{corpus-draw} variance directly, with
$J$ moving $\lesssim 0.005$ across redraws; because these run at a smaller
subsample than the full-corpus headline cells, that bound is conservative for
them. Both are far smaller
than the cross-arm overlaps they qualify ($0.010$--$0.161$) and than the gap to
self-identity ($J{=}1.0$), so the near-disjointness is safe; a conclusion resting
on $J$ resolved to finer than $\sim$0.005 would not be, but none here is.

The corpus axis
carries a size confound (the three headline corpora differ in size as well
as typology); the REAL-Space anchor, the within-PubChem size sweep, and
the size-matched typology probe (\S\ref{ssec:r-extras}) address it, the
last by subsampling PubChem and ZINC-22 to COCONUT's training size and
showing the one corpus-dependent contrast is typological rather than
size-driven. The learnability bar $F_{95\%,100}$ is
NMT-derived \citep{gowda2020optimal-vocab-nmt} and its transfer to
chemistry is itself an assumption. We fix the base alphabet
(Smirk), the canonicalization (RDKit isomeric, non-Kekul\'e), and the
algorithm set (BPE, Unigram-LM); a different base, a Kekul\'e
representation, and WordPiece are well-posed separate experiments. More
broadly, the study presupposes a fixed-vocabulary subword tokenizer;
learned-boundary and byte-level architectures, such as dynamic chunking
\citep{hwang2025dynamic-chunking-hnet}, hierarchical byte-word models
\citep{neitemeier2025hierarchical-autoregressive}, and byte-latent patching
\citep{pagnoni2024byte-latent-transformer}, dissolve the discrete vocabulary
altogether and so lie outside the BPE-versus-Unigram contrast drawn here.

\section{Conclusion}\label{sec:conclusion}

On chemistry SMILES, BPE and Unigram-LM do not build the same vocabulary.
Across all $22$ matched conditions (three corpus typologies,
two boundary policies, and the small vocabulary sizes where embeddings are
learnable), the two algorithms produce near-disjoint subword vocabularies
(the learned pieces above the shared $165$-token base; overlap $\le 0.161$)
and Unigram-LM segments molecules into $29$--$41\%$
more tokens, with the contrast stable across corpus and boundary and only one
secondary measure attenuating in magnitude on a single corpus, which we flag
explicitly.
The convergence that a small, valence-constrained alphabet could have
forced does not happen, and it does not arrive with scale either: a $V{=}8192$
anchor, eight times the headline vocabulary, leaves both contrasts essentially
unmoved. This removes the premise that the two algorithms
are interchangeable at the tokenizer level, the default the field has
inherited without testing. The lever is the algorithm, not the training
\emph{corpus}: the vocabularies transfer across domains at near-native cost
(within $\pm 1\%$ for BPE). Whether either arm serves a downstream model
better is the experiment this result now motivates.

\section*{Data and code availability}

\begin{sloppypar}
All trained tokenizers (the $44$ two-arm artifacts of the $22$ matched
conditions, plus the single-arm coordinates, the sensitivity sweep, and
structural probes), the per-condition measurement
data, and the table- and
figure-generating scripts are released as a versioned archive
[DOI: \texttt{10.5281/zenodo.21228245}]. This article and its figures are
licensed under CC-BY~4.0.
\end{sloppypar}

\section*{Acknowledgments}

This work builds directly on the Smirk tokenizer and its fixed
OpenSMILES glyph base released by \citet{wadell2026smirk} (the BattModels
group); both arms of the comparison are trained over that base using a pinned
fork of their implementation. I thank them for making the tokenizer and its
configuration openly available.

\section*{Funding}

This work received no external funding.

\section*{Competing interests}

The author declares no competing interests.

\bibliographystyle{plainnat}
\bibliography{refs}

\clearpage
\appendix
\renewcommand{\thetable}{A\arabic{table}}
\renewcommand{\theHtable}{app.\arabic{table}}
\setcounter{table}{0}

\section{Methods and reproducibility detail}\label{app:methods}

This appendix collects the study's methodological detail: corpus
construction and conformance filtering (Appendix~\ref{app:preproc}), tokenizer
training and byte-level determinism (Appendix~\ref{app:training}), the
one-factor-at-a-time sensitivity battery and structural probes
(Appendix~\ref{app:sensitivity}), and the learnable-regime confirmation
(Appendix~\ref{app:learnability}).

\subsection{Preprocessing and corpus construction}\label{app:preproc}

Each corpus passes through canonicalization, deduplication, and
conformance filtering.
RDKit isomeric canonicalization maps every input \texttt{MolFromSmiles}
$\to$ \texttt{MolToSmiles} (non-Kekul\'e, so aromatic systems are
emitted in RDKit's aromatic form); inputs RDKit cannot parse or round-trip
are dropped, the per-corpus drop rate is recorded, and the RDKit version is
pinned (\texttt{rdkit==2026.03.1}), since canonical output is version-dependent. Deduplication is
exact-string only, so tautomers and protonation states survive as distinct
entries.

A base-conformance filter then closes each corpus under the Smirk base. The
bare base tokenizer (the $165$-token regex with no merges) is the oracle: a
molecule is dropped iff its decomposition emits \texttt{[UNK]} (id $0$), a
frequency-independent per-molecule test that needs no trained tokenizer, so it
runs ahead of subsampling, the split, and training. We certified
all four corpora ($\sim$344M molecules scanned): COCONUT, ZINC-22, and
REAL-Space are $100\%$ conformant ($0$ dropped across $738{,}801$, $83.6$M, and
$136.4$M molecules), while PubChem drops $4{,}351$ of $123.4$M ($0.0035\%$).
All $4{,}351$ PubChem offenders carry an aromatic-silicon or aromatic-tellurium
atom that RDKit emits but OpenSMILES does not define, or one of a handful of
Si-family strings RDKit itself cannot re-parse, which the base-\texttt{[UNK]}
test catches where an RDKit check would error. Dropped
molecules are deposited per corpus for inspection.

Corpora exceeding the target size are reduced to the per-corpus totals
of Table~\ref{tab:corpora} by deterministic hash-partition subsampling,
so the conformance-scan counts above are pre-subsample. Each corpus's held-out
test split is $5\%$, capped at $10^6$ molecules, and is disjoint from training.

The two out-of-distribution evaluation corpora
(\S\ref{ssec:r-generalization}) are read with the trained PubChem tokenizers and
need only light handling; CycPeptMPDB needs none. tmQM ships a dative-bond
serialization (\texttt{->}\,/\,\texttt{<-}) that is not OpenSMILES and would
route to \texttt{[UNK]} on the \texttt{>} glyph; we derive a conformant encoding
by converting the dative bonds to single bonds and recomputing valence, a
representation choice that shifts implicit-hydrogen bookkeeping on metal-bound
atoms, so the tmQM reading rests on our derivation rather than an independent
reference encoding. The derivation deliberately bypasses the
canonicalize-and-dedup stage (and hence the base-conformance filter above) to
avoid re-perceiving the metal coordination, so a residual $0.115\%$ of complexes
retain an off-base aromatic-silicon atom; these account for the only
\texttt{[UNK]} either arm emits on tmQM ($0.002\%$ of tokens).

\subsection{Tokenizer training and determinism}\label{app:training}

Both arms call the same alphabet-construction and pre-tokenization routines
(\texttt{compute\_alphabet}, \texttt{tokenize\_words}), so the boundary policy
is applied inside those shared routines rather than per arm. The Unigram-LM
arm's \texttt{UnigramTrainer} \citep{kudo2018subword-reg} performs
frequent-substring seeding scored by frequency $\times$ length, an EM/pruning
loop, and finalization, following the SentencePiece reference
\citep{kudo2018sentencepiece}. Reference defaults are SentencePiece's for
Unigram-LM (maximum piece length $16$, seed pool $10^6$, two EM
sub-iterations, shrinking factor $0.75$) and a minimum merge frequency of $2$
for BPE (Sennrich's default and, since BPE imposes no piece-length cap, its
only free training knob). Table~\ref{tab:hyperparams} collects every knob, its
default, and its sensitivity ladder.

\begin{figure}[htbp]
  \centering
  \begin{minipage}[t]{0.485\textwidth}
    \centering
    \textbf{BPE} {\footnotesize(greedy, bottom-up)}\\[2pt]
    \rule{\linewidth}{0.8pt}
    \raggedright\footnotesize
    \begin{algorithmic}[1]
      \Require boundary-filtered glyph-id words; target $V$
      \State $\mathcal{V} \gets$ $165$-token base
      \While{$|\mathcal{V}| <$ target $V$}
        \State count adjacent token pairs
        \State $(p^{*}, c^{*}) \gets$ most-frequent pair
        \If{$c^{*} < f$} \textbf{stop} \Comment{\texttt{min\_frequency}}
        \EndIf
        \State merge $p^{*}$; add to $\mathcal{V}$
      \EndWhile
      \State \Return $\mathcal{V}$, ordered merge list
    \end{algorithmic}
  \end{minipage}\hfill
  \begin{minipage}[t]{0.485\textwidth}
    \centering
    \textbf{Unigram-LM} {\footnotesize(probabilistic, top-down)}\\[2pt]
    \rule{\linewidth}{0.8pt}
    \raggedright\footnotesize
    \begin{algorithmic}[1]
      \Require boundary-filtered glyph-id words; target $V$
      \State seed $\gets$ substrings, $\text{len}\le L$
        \Comment{\texttt{max\_piece\_length}}
      \State \quad capped at top $S$
        \Comment{\texttt{seed\_size}}
      \State assign each piece a probability
      \While{$|\mathcal{V}| >$ target $V$}
        \For{$m$ sub-iterations} \Comment{\texttt{n\_sub\_iterations}}
          \State E-step: expected piece usage
          \State M-step: re-estimate; drop $\mathrm{E}[\text{use}]<0.5$
        \EndFor
        \State prune: $\max(V, s|\mathcal{V}|)$
          \Comment{\texttt{shrinking\_factor}}
      \EndWhile
      \State finalize to target $V$
    \end{algorithmic}
  \end{minipage}
  \caption{BPE grows a vocabulary bottom-up; Unigram-LM prunes a large seed
  top-down. The $\triangleright$ comments mark where each training hyperparameter
  enters: one for BPE, four for Unigram-LM (defaults in
  Table~\ref{tab:hyperparams}).}
  \label{fig:algos}
\end{figure}

\input{tables/hyperparameters}

Three settings depart from the SentencePiece reference by design, not tuning.
Character coverage is $1.0$ rather than $0.9995$, because Smirk's base spans
the OpenSMILES alphabet by construction and leaves no rare-character tail.
Training is single-threaded, which makes every artifact byte-reproducible (the
multi-threaded reference defaults are not). SentencePiece's text-oriented
splitting (whitespace, Unicode script, dummy prefix) is replaced wholesale by
Smirk's chemistry-aware Layer-A/B pre-tokenizer, the point of fixing the base.
The two defaults that most directly govern piece granularity, the Unigram
maximum piece length ($16 \to 128$) and the BPE merge-frequency floor
($2 \to 0$), are treated as ablation axes (\S\ref{ssec:r-extras}). The
trainers live in a public, pinned fork of Smirk \citep{smirk-vtc-fork}.

The two trainers budget the six non-\texttt{[UNK]} special tokens
differently: BPE counts them inside the target $V$, while the
Unigram-LM trainer adds them on top. At a matched $V$ the Unigram-LM arm therefore
carries six more content pieces, an offset of $\le 2.4\%$ of $V$ (six pieces at
the smallest $V{=}256$). This lets the Unigram-LM arm compress slightly more than it
otherwise would, so it can only shrink the fertility gap reported in the body,
never inflate it.

For determinism, every tokenizer is trained twice and its two artifacts are
asserted byte-identical; a failed assertion is flagged rather than silently
passed.

\subsection{Sensitivity analysis and structural probes}\label{app:sensitivity}

The sensitivity battery sweeps each training hyperparameter
one-factor-at-a-time across its ladder in Table~\ref{tab:hyperparams}:
geometric ($\times 2$) ladders for the unbounded knobs (the Unigram maximum
piece length, the seed pool, and the BPE merge frequency) and within-range
grids for the bounded knobs, where doubling is undefined (the EM
sub-iterations and shrinking factor). Four pairwise interactions probe
covariation between knobs: maximum piece length $\times$ corpus typology, BPE
merge frequency $\times$ Unigram maximum piece length, sub-iterations $\times$
shrinking factor, and maximum piece length $\times V$. To stay tractable the
battery runs on a fixed representative subsample rather than each full corpus,
justified by the within-PubChem size sweep. Alongside the knob sweeps we
retain four structural probes: subsample redraws, a within-PubChem size sweep,
a size-matched typology probe, and a REAL-Space merge-exhaustion continuity
run.

\begin{figure}[htbp]
  \centering
  \includegraphics[width=\linewidth]{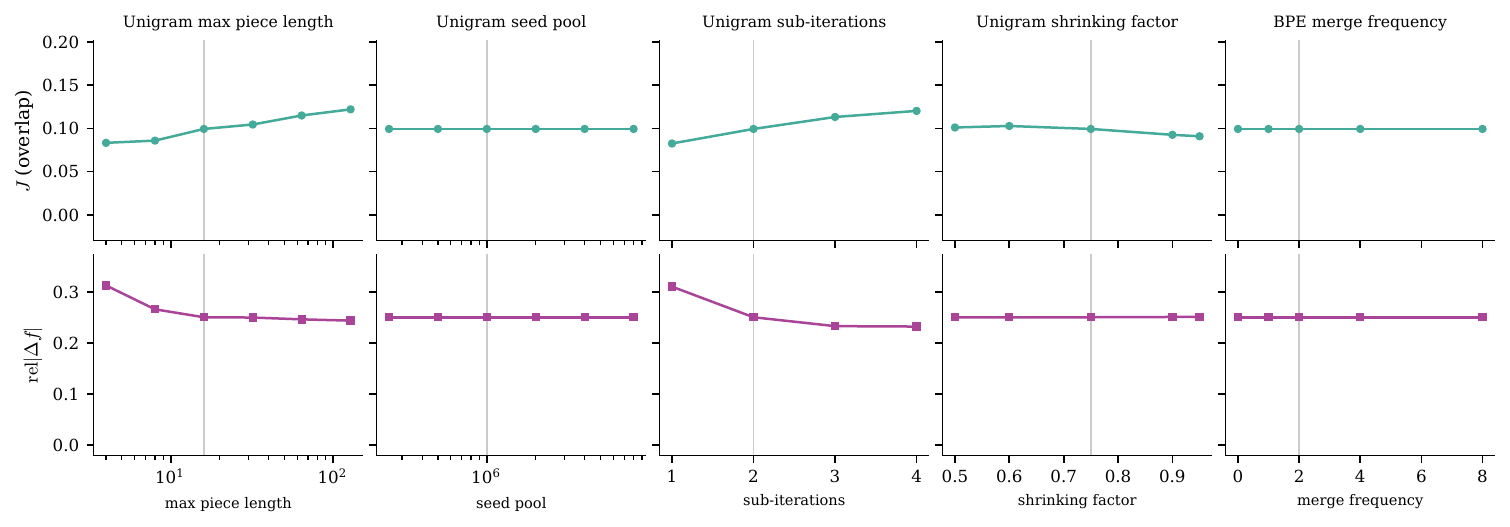}
  \caption{Per-knob sensitivity response curves on the fixed PubChem
  $V{=}512$ NMB subsample: vocabulary overlap ($J$, top) and the relative
  fertility gap ($\mathrm{rel}|\Delta f|$, bottom) as each training hyperparameter is
  swept across its ladder (the Unigram maximum piece length, seed pool,
  sub-iterations, and shrinking factor, plus the BPE merge frequency). The gray
  vertical lines mark the reference default. Both contrasts hold their
  separation on every rung; the seed pool and BPE merge frequency are inert
  (flat).}
  \label{fig:sensitivity-curves}
\end{figure}

\begin{figure}[htbp]
  \centering
  \includegraphics[width=\linewidth]{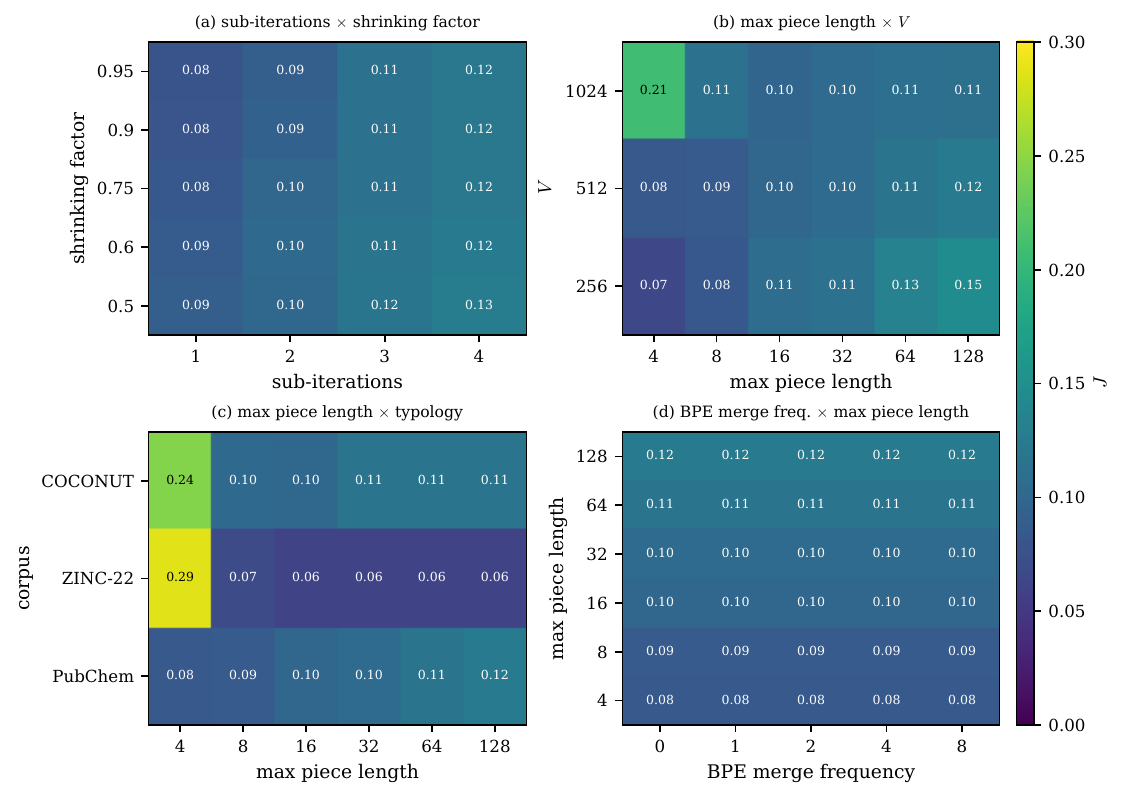}
  \caption{The four pairwise hyperparameter interactions, each a heatmap of
  the cross-arm vocabulary overlap $J$ (per
  cell annotated): (a)~sub-iterations $\times$ shrinking factor, (b)~maximum
  piece length $\times V$, (c)~maximum piece length $\times$ corpus typology,
  and (d)~BPE merge frequency $\times$ Unigram maximum piece length (both arms
  run permissive at once). Every cell stays more than $70\%$ disjoint
  ($J \le 0.29$); overlap rises above the headline $\le 0.161$ range only at the
  shortest-piece (max-piece-length~$4$) setting, peaking on the
  narrow-alphabet~corpora.}
  \label{fig:interactions}
\end{figure}

\begin{figure}[htbp]
  \centering
  \includegraphics[width=0.8\linewidth]{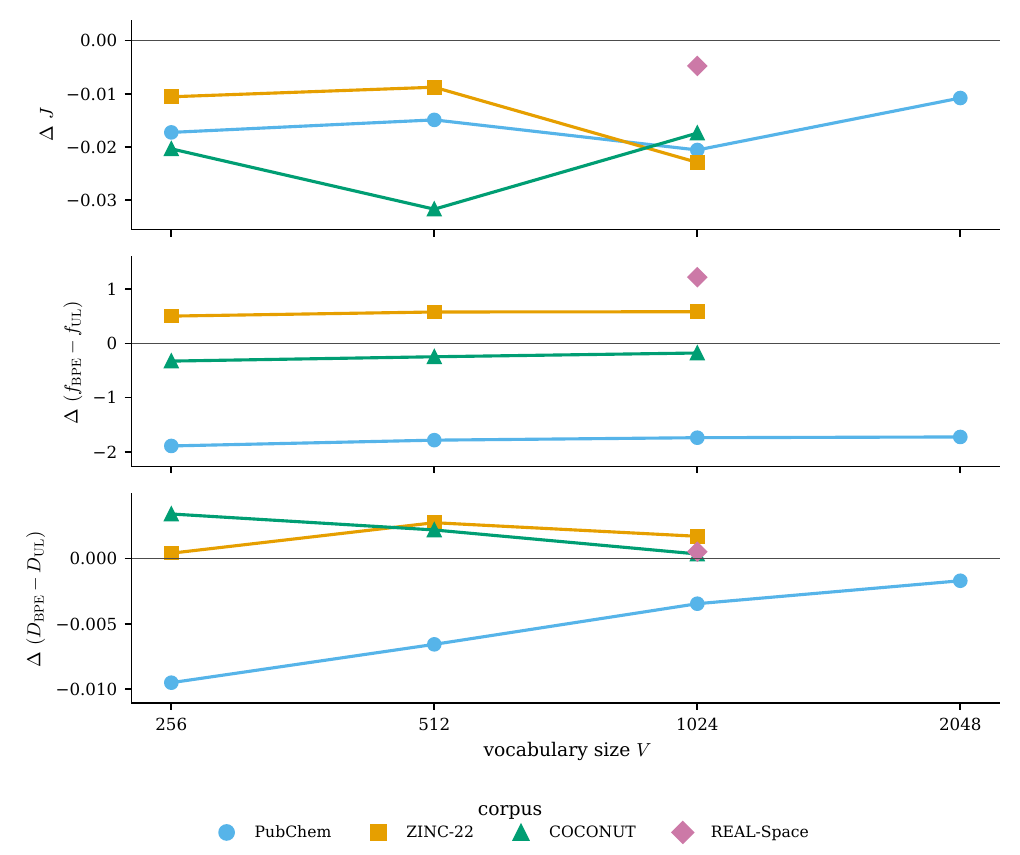}
  \caption{Algorithm$\times$boundary interaction: the signed NMB$-$MB
  difference per contrast, plotted against $V$ with corpus given by color and
  marker. The panels are, top to bottom, membership overlap $J$, the signed
  fertility gap $f_{\mathrm{BPE}}-f_{\mathrm{UL}}$ (in tokens), and the signed
  imbalance gap $D_{\mathrm{BPE}}-D_{\mathrm{UL}}$; each carries a zero baseline
  off which the sign reads. No contrast's outcome flips between the
  opaque-bracket (NMB) and permeable-bracket (MB) policies; the interaction
  terms are modest for overlap and imbalance and larger (in token units) but
  outcome-preserving for fertility. Only PubChem reaches $V{=}2048$; REAL-Space
  is a single $V{=}1024$ point.}
  \label{fig:interaction}
\end{figure}

\subsection{Learnability confirmation}\label{app:learnability}

The learnable-regime check computes the clearance $c_n$, the fraction of
\emph{learned} vocabulary pieces (those above the shared $165$-token base, which is
identical across both arms and chosen by neither algorithm) firing
at least $n$ times in the full training corpus, for
$n \in \{50, 100, 200\}$. The bar $F_{p,n}$ requires $c_n \ge p$; the headline
$F_{95\%,100}$ ($c_{100} \ge 0.95$) is reported per arm in
Table~\ref{tab:results-delta-f}, and the full per-arm $c_n$ sweep is in
Appendix~\ref{app:mechanism}, Table~\ref{tab:results-nsweep}. Reading each $c_n$
against $p \in \{0.90, 0.95, 0.99\}$ recovers all nine $(p,n)$ bar outcomes from that
sweep, so the bar's robustness to both thresholds can be checked directly.

\paragraph{The learnable ceiling and language-model scale.} Modern language
models tokenize with vocabularies one to two orders of magnitude larger than
ours ($32$K for Llama~2 and Mistral, $128$K for Llama~3, $256$K for Gemma),
which raises the question whether the divergence (\S\ref{sec:discussion}) is an
artifact of staying small. It is not. The $F_{95\%,100}$ audit shows the
Unigram-LM arm crossing into the undertrained tail by $V{=}2048$ even on PubChem,
our largest headline corpus (clearance $0.57$, below the $0.95$ bar), while BPE, whose
coarser pieces fire more often, holds at $0.997$; by the $V{=}8192$ anchor both
arms are unsafe (BPE $0.68$, Unigram-LM $0.13$), a third of BPE's pieces and
almost nine-tenths of Unigram-LM's firing too rarely to learn a reliable
embedding. The cause is the base: a $158$-glyph alphabet under hard valence and
ring constraints admits few high-frequency substructures, which is why Smirk's
merge trainer exhausts naturally at $V\approx4{,}331$ (\S\ref{ssec:r-extras}) and
why deployed chemistry tokenizers stay small (SPE and APE learn
$\sim$3{,}000--5{,}300 pieces; \S\ref{ssec:landscape}). Language-model
vocabularies are large because they are sized to the lexical diversity of
multilingual text and code, not a constrained formal language.

\section{Vocabulary inventory}\label{app:vocab}

This appendix inventories the vocabulary in two parts: the fixed base both
arms share (Appendix~\ref{app:base-glyphs}), and the multi-glyph pieces each arm
selects above it (Appendix~\ref{app:multiglyph}), read three ways: the complete
PubChem $V{=}256$ snapshot, the shared core's growth with $V$, and the
mirror-image arm-exclusive sets.

\subsection{The base glyph alphabet}\label{app:base-glyphs}

\input{tables/base_glyphs}

Both arms train over the same fixed Smirk base, installed as length-$1$ pieces,
so the only cross-arm difference is the multi-glyph pieces each \emph{selects}
above it (\S\ref{ssec:training}). Table~\ref{tab:base-glyphs} lists all $165$ base
tokens by OpenSMILES role. Because any OpenSMILES-conformant string decomposes
into them with no \texttt{[UNK]}, the bare base also serves as the conformance
oracle of Appendix~\ref{app:preproc}.

\subsection{Learned multi-glyph pieces}\label{app:multiglyph}

Above the shared base (Appendix~\ref{app:base-glyphs}), each arm selects only a
small set of multi-glyph pieces, and these learned sets are what the overlap $J$
(\S\ref{ssec:r-jaccard}) compares.

\paragraph{The $V{=}256$ snapshot.}
\input{tables/multiglyph_v256}
To make the near-disjointness concrete, we
partition the \emph{complete} learned set for the PubChem $V{=}256$ matched pair
under both boundary policies three ways by cross-algorithm membership
(Table~\ref{tab:multiglyph-v256}). Of the $186$ distinct pieces the two arms select
under NMB, only $10$ are shared ($J{=}0.054$; $13$ of $183$ under MB,
$J{=}0.071$). The shared set is dominated by alkyl chains
(\texttt{CCCC}\,\dots\,\texttt{CCCCCCCCCCCCCCCC}) with a few short ether and amine
linkers, the motifs both algorithms agree to absorb; BPE's exclusive pieces are
short functional fragments (\texttt{=O}, \texttt{C=C}, \texttt{C\#N}) while
Unigram-LM's run longer, toward heteroatom and homopolymeric chains
(\texttt{COCCOCCOCCN}, \texttt{=C=C=C=C=C=C=C=C}). Permitting merges to cross the
bracket boundary (MB) adds a few bracket-internal pieces (\texttt{[C@},
\texttt{[Si}) and the aromatic ring \texttt{ccccc} to the shared set; the picture
is otherwise unchanged. The boundary policy is the minor axis: $124$ of
the $186$ NMB pieces recur under MB (grayed in both tables; the policy-specific
ones left plain), so the selection is largely boundary-robust. Of those $124$, only two even change cross-algorithm membership
with the policy: the aromatic ring \texttt{ccccc}, BPE-only
under NMB but shared under MB, and the heptyl chain \texttt{CCCCCCC}, shared under
NMB but Unigram-LM-only under MB. Neither is a bracket piece, so the policy moves
their membership only indirectly, by perturbing the merge budget.

\paragraph{Growth of the shared core across $V$.}
\input{tables/shared_core_growth}
The three-way split above is a snapshot at $V{=}256$; tracking
the shared set up the vocabulary-size ladder shows it to be a genuine, growing
\emph{core}. Across $V \in \{256, 512, 1024, 2048\}$ the PubChem
BPE\,$\cap$\,Unigram-LM set is strictly nested under both boundary policies (no
piece the two arms once agreed on is ever dropped), growing from $10$ to $311$
pieces under NMB and $13$ to $345$ under MB, with the overlap rising from
$J{=}0.054$ to $0.090$ (NMB) and from $0.071$ to $0.100$ (MB, where it plateaus)
(Table~\ref{tab:shared-core-growth}). The growth is
one-directional: the newly shared pieces are overwhelmingly substructures
Unigram-LM had already isolated at a smaller $V$, which BPE's later merges only
then absorb ($175$ of the $183$ pieces added at $V{=}2048$ under NMB; the reverse
essentially never occurs), the mechanism the body invokes for the
rising-overlap trend (\S\ref{ssec:r-stability}). Even at $V{=}2048$ the core is a
small fraction of either arm's $\sim$1{,}890 multi-glyph pieces, and the agreement
is concentrated in alkyl chains and common heteroatom and ring motifs: the two
algorithms converge only slowly, and on the least distinctive substructures. The
shared core is itself largely boundary-robust: $261$ of its pieces belong to both
the NMB and MB cores (grayed in both tables), leaving only $50$ NMB-specific and
$84$ MB-specific shared pieces.

\paragraph{The arm-exclusive sets.}
\input{tables/arm_exclusive}
The mirror image of the shared core is what each arm keeps to itself.
Table~\ref{tab:arm-exclusive} summarizes the BPE-only and Unigram-LM-only
multi-glyph sets across $V$, and three points stand out. First, the two arms
select a near-equal \emph{number} of exclusive pieces at every $V$ ($1{,}581$ vs
$1{,}578$ at $V{=}2048$, NMB), and these exclusive sets dwarf the shared core by
roughly $5\times$ even at the largest $V$, so the near-disjointness is not a
small-$V$ artifact but persists at every scale. Second, the fate of the two sets
is asymmetric, the per-piece face of the one-directional catch-up above:
every BPE-only piece at $V{=}256$ is still BPE-only at
$V{=}2048$ (Unigram-LM never adopts it), whereas most $V{=}256$ Unigram-LM-only
pieces ($72$ of $87$, NMB) are absorbed into the shared core as BPE's later merges
catch up. Third, the exclusive pieces differ sharply in length (the table's mean-
and max-length columns): Unigram-LM's run longer on average, BPE's in an uncapped
tail, a contrast taken up in full in Appendix~\ref{app:length}.

\begin{figure}[htbp]
  \centering
  \includegraphics[width=\linewidth]{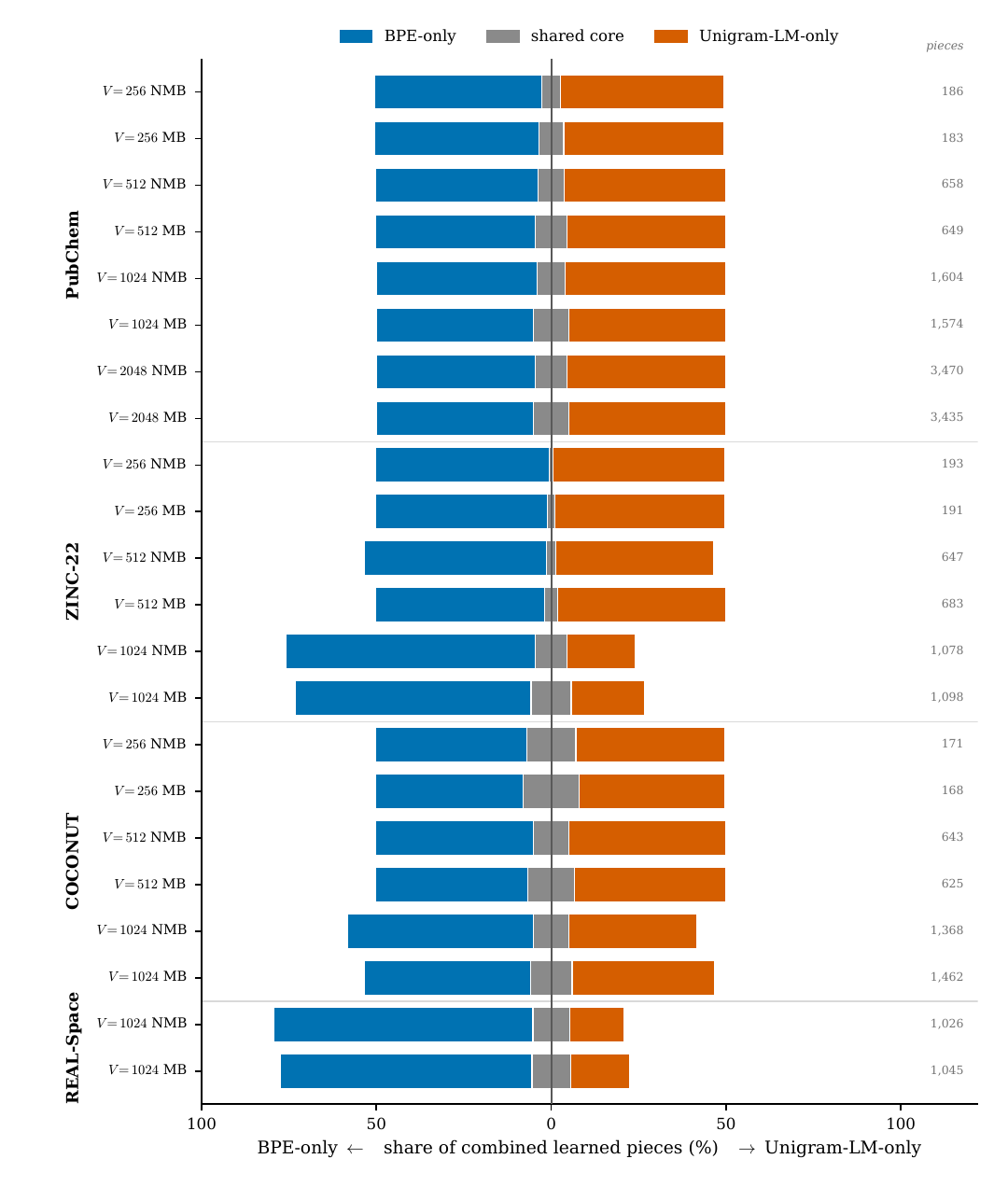}
  \caption{The learned multi-glyph vocabularies are near-disjoint by composition,
  in every matched condition. Each bar splits the two arms' combined learned-piece
  set (their union, normalized to $100\%$) outward from the center: BPE-only
  pieces to the left (blue), Unigram-LM-only to the right (orange), with the
  shared core the thin neutral spine at the center, its width the unweighted
  overlap $J$. The spine stays a sliver throughout, and the two wings are
  near-symmetric except on the narrow-alphabet ZINC-22 $V{=}1024$ and REAL-Space
  cells (and, more mildly, COCONUT $V{=}1024$), where Unigram-LM's pruning
  saturates and its wing shrinks. The arms agree
  least where it matters most: the shared core is dominated by rare pieces, so its
  frequency-weighted share is smaller still ($J_{\mathrm{w}}\le0.05$;
  Fig.~\ref{fig:overlap}). Rows group by corpus (typology order) then $V$ and
  boundary; the right column gives the combined set's absolute size, with per-arm
  realized counts in Table~\ref{tab:results-realized-vocab}.}
  \label{fig:membership}
\end{figure}

\section{Structural character of the learned pieces}\label{app:structural}

The five analyses in this section, which the body reads as one signature of
the greedy-versus-probabilistic mechanism (\S\ref{ssec:r-mechanism}),
characterize the internal structure and chemistry of each arm's pieces: what
kind of substructure each arm makes atomic
(Appendix~\ref{app:composition}), how long its pieces run
(Appendix~\ref{app:length}), which glyphs co-occur within them
(Appendix~\ref{app:cooccurrence}), whether the pieces decompose into
in-vocabulary parts (Appendix~\ref{app:closure}), and whether they keep
functional bonds inside one token (Appendix~\ref{app:fg-alignment}). Together
they resolve a single contrast: BPE builds compositionally closed pieces
that bind whole rings and functional groups, while Unigram-LM keeps
non-closed heteroatom and unsaturated chains. How the cross-arm divergence varies
across corpus typologies is taken up in Appendix~\ref{app:contrasts}; how it
survives non-canonical input (Appendix~\ref{app:robustness}) and
out-of-distribution chemistry (\S\ref{ssec:r-generalization}).

\subsection{Substructure classes of the learned pieces}\label{app:composition}

\input{tables/composition}

The piece lists (Appendix~\ref{app:multiglyph}) suggest the two arms favor different
\emph{kinds} of substructure; Table~\ref{tab:composition} classes every
multi-glyph piece and confirms it. The aromatic-ring split the body reads off it
(\S\ref{ssec:r-jaccard}) is the clearest signal: BPE merges ring fragments
(\texttt{cccc}, \texttt{ccccc}) into whole tokens where Unigram-LM stays
near-atomic. On Unigram-LM's side the table gives the mirror image: its exclusive
pieces skew to aliphatic heteroatom chains, and on the natural-products corpus to
long unsaturated carbon runs ($49\%$ of COCONUT's Unigram-LM-only set, the
cumulated \texttt{=C=C=}$\dots$ (cumulene) motifs). The shared core, by contrast,
is heteroatom-led like Unigram-LM's, but it also holds the saturated alkyl and
(on COCONUT) aromatic motifs both algorithms absorb early: the common
substructures, not either arm's specialty.

\subsection{Piece-length distributions}\label{app:length}

\begin{figure}[htbp]
  \centering
  \includegraphics[width=\linewidth]{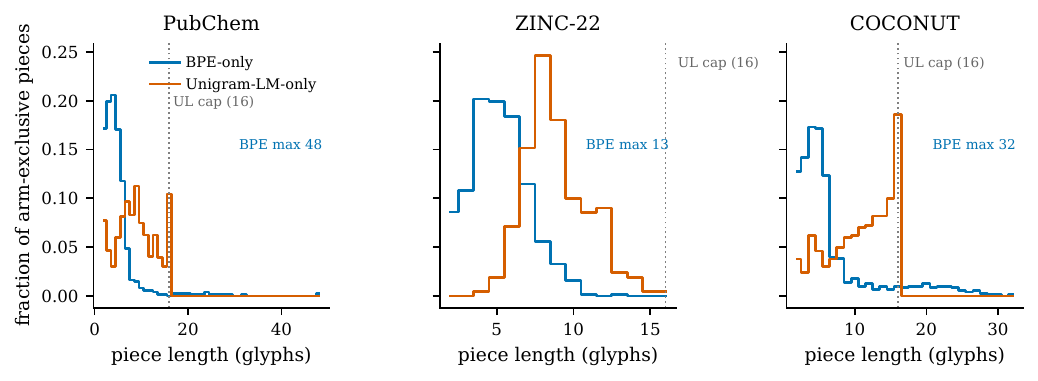}
  \caption{Glyph-length distribution of the BPE-only and Unigram-LM-only
  multi-glyph pieces (PubChem, ZINC-22, COCONUT; $V{=}1024$, NMB). Unigram-LM is
  capped at $16$ glyphs (\texttt{max\_piece\_length}, dotted line); BPE is
  uncapped, with a corpus-dependent tail (annotated maximum per panel). Each
  panel's $x$-axis spans its own length range (the $y$-axis is shared), so the
  annotated maxima, not the on-page tail widths, carry the cross-corpus
  comparison. On ZINC-22 the narrow alphabet keeps even BPE's longest piece
  below the cap.}
  \label{fig:piece-length}
\end{figure}

\begin{figure}[htbp]
  \centering
  \includegraphics[width=0.55\linewidth]{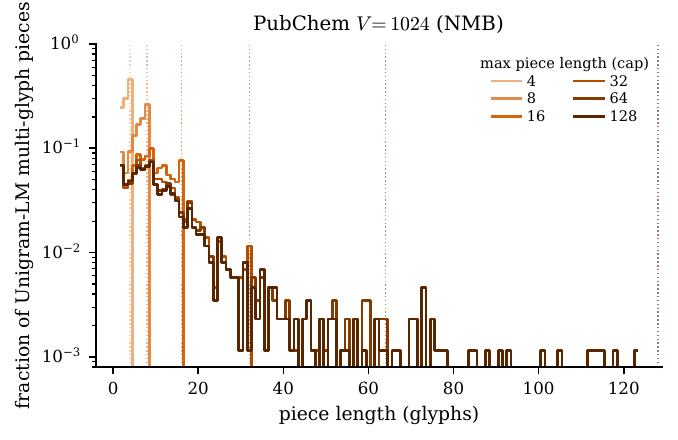}
  \caption{Unigram-LM piece-length distribution on PubChem ($V{=}1024$, NMB) as
  its \texttt{max\_piece\_length} cap is swept over $\{4,8,16,32,64,128\}$ (dotted
  lines mark each cap). The right edge tracks the cap exactly, so the $16$-glyph
  wall in Figure~\ref{fig:piece-length} is the hyperparameter, not an intrinsic
  limit; the mode stays short throughout.}
  \label{fig:piece-length-cap}
\end{figure}

Figure~\ref{fig:piece-length} plots the glyph-length distribution of each arm's
exclusive pieces, and at first it cuts against the fertility gap. \emph{On
average} Unigram-LM's pieces are the longer ones, in all $22$
matched conditions (mean glyph length $9.0$ vs.\ $5.0$ on PubChem $V{=}1024$,
$8.9$ vs.\ $5.1$ on ZINC-22 $V{=}1024$). What BPE owns is the \emph{tail}, and
only where the corpus supports long repeats: uncapped, it accretes a few very
long merges on diverse PubChem (to $48$ glyphs, $96$ at $V{=}2048$) and
scaffold-rich COCONUT (to $32$), while Unigram-LM stops at a hard wall at its
\texttt{max\_piece\_length} default of $16$ (Table~\ref{tab:hyperparams}). On
narrow-alphabet ZINC-22 and REAL-Space, BPE cannot form such repeats, and even
its longest exclusive piece ($10$--$13$ glyphs) is \emph{shorter} than
Unigram-LM's cap. The mean and the tail thus disagree: Unigram-LM wins the mean
everywhere, BPE the tail only where long substructure exists to merge. That
asymmetry is moreover partly an artifact of Unigram-LM's cap: retraining it
across the
\texttt{max\_piece\_length} ladder (Figure~\ref{fig:piece-length-cap}) tracks the
right edge from $4$ to $128$ glyphs and lifts the mean ($3.2$ to $15.0$), yet
even at the most permissive cap the bulk stays short, the cap shifting the tail
and not the mode. None of this coarsens Unigram-LM's segmentation: it samples
likely segmentations rather than greedily applying its longest pieces, so its
longer vocabulary pieces coexist with its higher fertility (more, shorter tokens
in use).

\subsection{Glyph co-occurrence within learned pieces}\label{app:cooccurrence}

\begin{figure}[htbp]
  \centering
  \includegraphics[width=\textwidth]{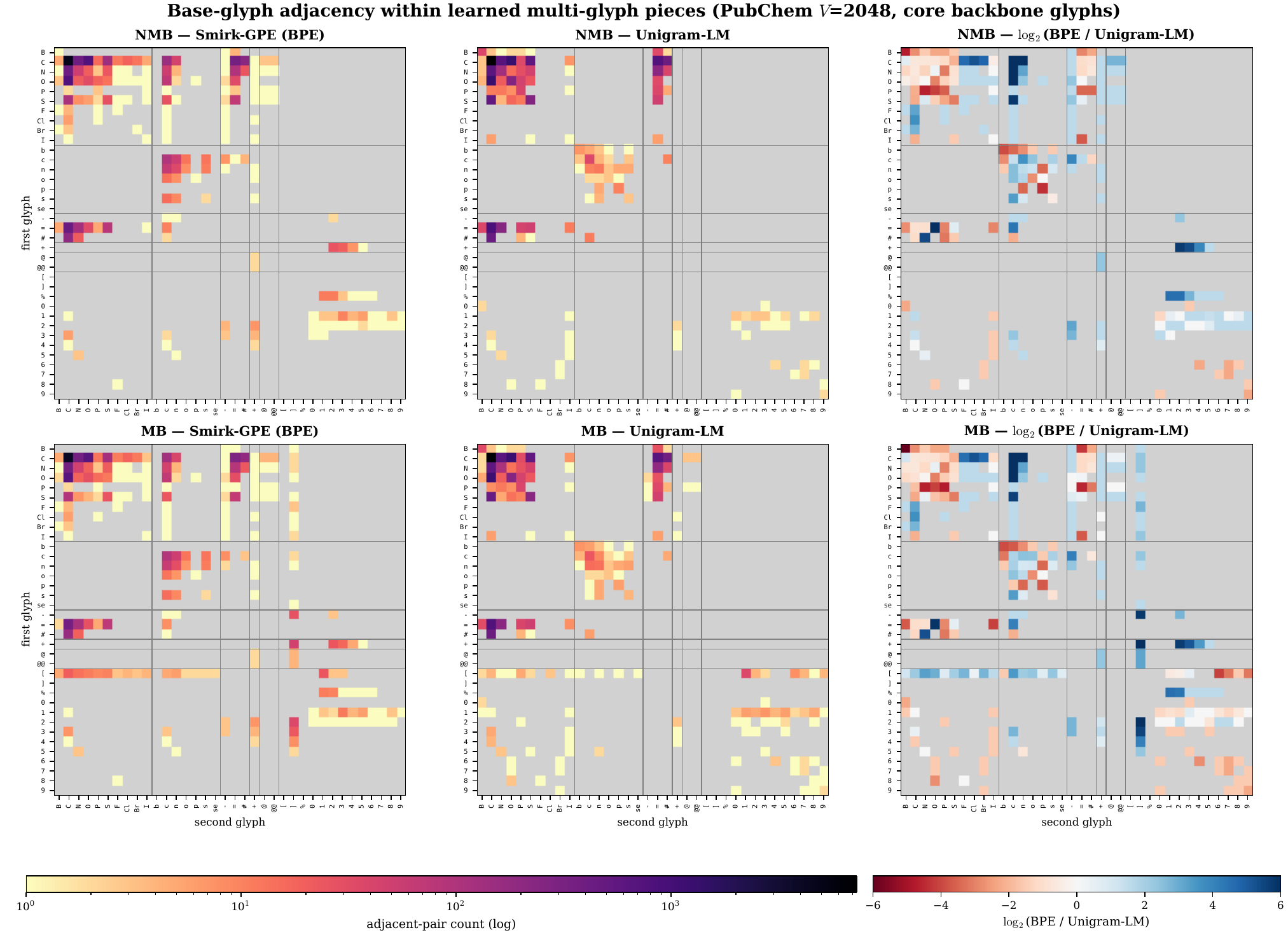}
  \caption{Adjacent base-glyph pairs within learned multi-glyph pieces (PubChem
  $V{=}2048$), over the core backbone glyphs (organic, aromatic, bond, charge,
  chirality, and structural glyphs, grouped by OpenSMILES role with dividers;
  rare bracket-atom element symbols omitted). Rows are the boundary policy
  (no-merge-brackets NMB, merge-brackets MB); the first two columns are Smirk-GPE
  (BPE) and Unigram-LM on a shared log count scale, the third their log-ratio
  (blue: BPE-heavier, red: Unigram-LM-heavier; cells empty in both are grey).
  Under NMB the bracket delimiters \texttt{[}/\texttt{]} never co-occur (merges
  cannot cross them); MB switches the \texttt{[} row on. Both arms are dominated by
  the carbon backbone, weighted differently: on average Unigram-LM builds
  longer, more homo-atomic pieces.}
  \label{fig:glyph-cooccurrence}
\end{figure}

The piece inventories can also be read at the level of constituent glyphs: for
each learned multi-glyph piece we count its adjacent base-glyph pairs
(\texttt{CCO} contributes \texttt{C}--\texttt{C} and \texttt{C}--\texttt{O}).
Figure~\ref{fig:glyph-cooccurrence} maps these for the PubChem $V{=}2048$ pair
over the core backbone glyphs, grouped by OpenSMILES role (the $\sim$$100$ rare
bracket-atom symbols, which merge sparsely and only under MB, are omitted;
Appendix~\ref{app:multiglyph}). The glyph level adds two findings the piece-level
tables cannot show. First, the boundary policy is a structural switch: under
no-merge-brackets (NMB) \emph{no} adjacency touches a bracket delimiter
(\texttt{[}, \texttt{]}; merges cannot cross the boundary), whereas merge-brackets
(MB) turns on the \texttt{[} row of bracket-internal pairs (\texttt{[}--\texttt{C},
\texttt{[}--\texttt{N}, \texttt{[}--ring-digit; $3.4\%$ of BPE adjacencies and
$1.5\%$ of Unigram-LM's then involve \texttt{[}). Second, at near-equal piece
counts Unigram-LM packs $\sim$$1.65\times$ as many adjacencies, the glyph-level
face of its longer pieces (Appendix~\ref{app:length}). Both arms anchor on the
\texttt{C}--\texttt{C} backbone but weight it differently, echoing the composition
split (right column): Unigram-LM leans to sulfur and oxygen chains
(\texttt{C}--\texttt{S}, \texttt{O}--\texttt{O}), BPE into the aromatic--aliphatic
junction (\texttt{C}--\texttt{c}).

\subsection{Compositional closure of the learned vocabularies}\label{app:closure}

\input{tables/closure_detail}

The arm-exclusive sets (Appendix~\ref{app:multiglyph}) and length distributions
(Appendix~\ref{app:length}) describe \emph{which} pieces each arm keeps; a complementary
question is how those pieces relate to one another \emph{within} a single
vocabulary. BPE answers it by construction: built bottom-up, every learned piece
is the concatenation of two pieces already in the vocabulary, so the BPE
vocabulary is \emph{compositionally closed}. Unigram-LM, pruning a seed pool
top-down by likelihood, is under no such constraint, so whether its pieces
decompose into in-vocab parts is an empirical property. We read it off the
realized vocabularies alone (no corpus); it gives a construction-independent
view of the same greedy-versus-probabilistic mechanism the body traces
(\S\ref{ssec:r-mechanism}).

We measure closure identically for both arms from the realized set, so BPE's
value is a measured anchor rather than a restatement of its rule. Let $\mathcal{V}$ be the
full realized vocabulary (the $165$-token base plus the learned multi-glyph
pieces) and $M=\{p\in\mathcal{V}:|p|\ge 2\}$ the multi-glyph set the overlap compares.
Because the base is complete, every single glyph is trivially in $\mathcal{V}$, so the
metrics range over splits and sub-pieces of length $\ge 2$, where the question
is non-trivial. We report three (Table~\ref{tab:results-closure}): the
\emph{binary-split closure} $c_{\mathrm{bin}}$, the fraction of $M$ admitting a
split $p{=}a{\cdot}b$ with both parts in $\mathcal{V}$ (BPE's merge-closure invariant, so
$c_{\mathrm{bin}}^{\mathrm{BPE}}{=}1$ exactly); the \emph{orphan rate}
$c_{\mathrm{orph}}$, the fraction of length-$\ge 3$ pieces with \emph{no} proper
$\ge 2$-glyph sub-piece in $\mathcal{V}$ (pieces that share no building block with the
rest of the vocabulary, zero for BPE by the same invariant); and the stronger
\emph{full-substring closure} $c_{\mathrm{full}}$, the fraction whose every
$\ge 2$-glyph substring is in $\mathcal{V}$, which is non-trivial for both arms.

Three readings. First, Unigram-LM is far less self-referential than BPE in every
one of the $22$ matched conditions: $c_{\mathrm{bin}}^{\mathrm{UL}}$ runs
$0.36$--$0.64$ on the diverse and natural-products corpora and falls to $0.12$ on
the narrowest alphabet (ZINC-22 $V{=}256$), so roughly half its pieces, and on
ZINC-22 nearly nine in ten, do not decompose into in-vocab parts. Second, the gap
survives the stronger metric and an exact BPE anchor: even full-substring closure
separates the arms by a wide margin ($c_{\mathrm{full}}^{\mathrm{BPE}}$
$0.55$--$0.77$ against $c_{\mathrm{full}}^{\mathrm{UL}}\le 0.05$), and that BPE's
own $c_{\mathrm{full}}$ sits well below $1$ confirms the measurement is reading
genuine structure, not echoing the merge rule (merge-closure guarantees one split,
not all substrings). Third, the orphan population is largely the arm-exclusive
tail the other appendices isolate: Unigram-LM's non-closed pieces are
overwhelmingly structural rather than bracket-internal and are the long
heteroatom chains (\texttt{COCOCOCOCOC}, \texttt{OCCOOCCOOCCO},
\texttt{OCCONOCCO}) it keeps without keeping their parts (Appendix~\ref{app:composition},
Appendix~\ref{app:length}), because likelihood scores each piece independently where
merging must first own the sub-pieces. The orphan rate is non-increasing in
$V$ (PubChem NMB $0.45\to 0.18$ over $V{=}256\to 1024$), the same
convergence-from-below the overlap and imbalance trends show, but it never
vanishes, and on the narrow alphabets its orphans are the saturated arm's
pruned-back survivors (Appendix~\ref{app:narrow}).

\subsection{Chemical functional-bond locality}\label{app:fg-alignment}

The closure and composition analyses (Appendix~\ref{app:closure}, Appendix~\ref{app:composition})
describe each vocabulary as a set of pieces; a complementary question is whether
the \emph{realized segmentation} respects chemistry: whether the learned
tokens align with chemically meaningful units, the cheminformatic analogue of
asking whether a subword tokenizer's boundaries respect morphemes. We make this
precise with the chemically salient bonds hardest to keep whole: the
\emph{multiply-bonded heteroatoms}. A non-carbon atom joined by a double or
triple bond is the core of a canonical functional group: the $=$O of a carbonyl,
the $\#$N of a nitrile, the $=$N of an imine, the $=$O on sulfur, phosphorus, or
nitrogen (sulfonyl, phosphoryl, nitro), the $=$S of a thiocarbonyl. It is read
straight off the molecular graph, so the measure needs no curated list
of surface forms. On each held-out molecule, for every such bond, we ask whether
the arm keeps it \emph{token-local}: whether the heteroatom shares a token with
its $=$/$\#$ bond glyph. We measure locality per heteroatom-and-bond rather than
per whole functional group because SMILES branch syntax scatters even a two-atom
group across a parenthesis (a carbonyl is written \texttt{C(=O)}, its carbon a
shared backbone atom), so the chemically meaningful unit is the $=$O, not the
full \texttt{C(=O)}.

\input{tables/fg_alignment_detail}

Table~\ref{tab:results-fg-alignment} reports per-arm locality $\ell$ and the gap
$\Delta\ell=\ell^{\mathrm{BPE}}-\ell^{\mathrm{UL}}$ over the $22$ matched conditions.
The separation is near-total and uniform: BPE keeps $0.95$--$1.00$ of functional
bonds inside a single token, Unigram-LM at most $0.03$, a gap of $0.93$--$0.99$ on
every corpus, vocabulary size, and boundary policy. The carbonyl makes it
sharpest: BPE binds the $=$O in essentially every occurrence
($\ell^{\mathrm{BPE}}_{\mathrm{C=O}}\ge 0.995$), while Unigram-LM does so in
\emph{none}, with $\ell^{\mathrm{UL}}_{\mathrm{C=O}}=0.000$ to three places in all
$22$ conditions, over held-out splits of up to $10^6$ molecules. The one class
Unigram-LM keeps with any regularity is the nitrile
($\ell^{\mathrm{UL}}_{\mathrm{C\#N}}$ up to $0.26$), the terminal \texttt{C\#N}
that is also one of its own short pieces.

This is the in-context face of the vocabulary findings. The orphan rate
(Appendix~\ref{app:closure}) and the arm-exclusive chains (Appendix~\ref{app:multiglyph})
say Unigram-LM spends its multi-glyph budget on long homo-atomic and unsaturated
runs; locality confirms the consequence on real molecules. Where BPE's
frequency-greedy merging crystallizes the small functional motif (\texttt{=O},
\texttt{C\#N}) as a token, Unigram-LM's likelihood pruning leaves the heteroatom
split from its bond, because the same $=$ glyph it declines to bind to oxygen it
spends on the \texttt{=C=C=C} cumulene runs it keeps exclusively
(\S\ref{ssec:r-mechanism}). The reading is on the held-out split with $95\%$
molecule-resampled bootstrap CIs; at these denominators the intervals are tight
to the third decimal.

\section{Corpus-specific contrasts}\label{app:contrasts}

Appendix~\ref{app:multiglyph} inventories the learned pieces on PubChem's diverse
alphabet; this section follows the membership contrast across corpus typologies:
the narrow
drug-like alphabet of ZINC-22 (Appendix~\ref{app:narrow}), the natural-products
corpus COCONUT (Appendix~\ref{app:coconut}), and whether the cross-algorithm
agreement is itself corpus-specific (Appendix~\ref{app:cross-corpus}).

\subsection{Narrow-alphabet contrast: ZINC-22}\label{app:narrow}

\input{tables/narrow_contrast}

ZINC-22, the drug-like corpus, has a far narrower alphabet than PubChem, where
both arms reach the target vocabulary; Table~\ref{tab:narrow-contrast} shows the
narrow alphabet changes the picture in three ways.
First, the divergence is \emph{sharper} at small $V$: the overlap is
$J{=}0.010$ at $V{=}256$ (NMB), roughly five times lower than PubChem's $0.054$,
with only two shared multi-glyph pieces. Second, the Unigram-LM arm
\emph{saturates}: its narrow alphabet offers too few high-likelihood pieces for
the pruning to reach target, so it settles at $\sim$$310$ pieces and adds
essentially nothing from $V{=}512$ to $V{=}1024$, while BPE fills to target
($867$). The two vocabularies are thus sharply unequal in size ($867$ vs $310$ at
$V{=}1024$) where on PubChem they are near-equal ($867$ vs $865$); at $V{=}2048$
the Unigram-LM arm, unsafe by a wide margin, is left untrained, leaving no matched
pair. Third, the Unigram-LM-only set \emph{shrinks} with $V$ ($292 \to 211$, NMB),
the opposite of PubChem: Unigram-LM stops adding pieces while BPE's later merges
absorb its small fixed set into the shared core. The overlap does rise with $V$
(to $0.092$/$0.115$ at $V{=}1024$), but partly mechanically: the size gap caps the
attainable overlap at
$|\mathcal{V}^{\mathrm{multi}}_{\mathrm{UL}}|/|\mathcal{V}^{\mathrm{multi}}_{\mathrm{BPE}}| \approx
0.36$, and the observed value is about a quarter of that ceiling, so genuine
disagreement still carries the result (\S\ref{ssec:r-jaccard}). The combinatorial
REAL-Space corpus, narrow-alphabet for the same reason, behaves identically: its
size gap caps the attainable overlap at $\approx 0.31$, and the observed $0.105$
sits at about a third of that ceiling. On a narrow alphabet, then, the two
algorithms diverge for a second reason beyond selecting different pieces:
Unigram-LM simply runs out of pieces to select.

\subsection{Natural-products contrast: COCONUT}\label{app:coconut}

\input{tables/coconut_contrast}

COCONUT, the natural-products corpus, completes the typology and inverts the
cross-$V$ trend. Table~\ref{tab:coconut-contrast} shows the overlap is
\emph{highest} at small $V$ ($J{=}0.161$ at $V{=}256$, MB; about twice the
highest small-$V$ overlap on any other corpus) and \emph{falls} as $V$ grows (to
$0.119$ at $V{=}1024$), the opposite of PubChem and ZINC-22, where it rises. The
reason is visible in the split: both arms agree on far more pieces at small $V$
than elsewhere ($24$--$27$ shared at $V{=}256$, against PubChem's $10$ and
ZINC-22's $2$), because the dominant natural-product motifs (long alkyl chains
and aromatic rings) are frequent enough that both algorithms absorb them
immediately. The shared core still grows in absolute terms ($24 \to 65 \to 139$,
NMB), but the near-disjoint remainder grows faster, so the shared \emph{fraction}
dilutes. As the smallest corpus, COCONUT also begins to starve the Unigram-LM arm
at $V{=}1024$ (it realizes $640$ pieces against BPE's $867$ under NMB), a milder
version of the ZINC-22 saturation. This early-agreement-then-dilution is not a
step toward convergence: the large-$V$ anchor (\S\ref{ssec:r-stability}) confirms
the remainder stays near-disjoint well beyond the grid.

\subsection{Universality of the agreement across corpora}\label{app:cross-corpus}

A final question: when the two algorithms \emph{do} agree, do they agree on the
\emph{same} pieces regardless of corpus? Largely not. The three
shared cores at $V{=}1024$ (NMB; sizes $128$, $99$, and $139$ for PubChem,
ZINC-22, and COCONUT) overlap only weakly: $23$ pieces between PubChem and
ZINC-22, $31$ between PubChem and COCONUT, and $15$ between ZINC-22 and COCONUT.
Just $9$ pieces belong to all three cores. That universal core is chemically
unremarkable: long alkyl
chains (\texttt{CCCCCCCCC}, \texttt{CCCCCCCCCC}, \texttt{CCCCCCCCCCC}) plus a
handful of generic motifs (\texttt{N\#C}, \texttt{C\#CC}, \texttt{SCCS},
\texttt{C\#CCN}, \texttt{SCCCS}, \texttt{C=CC=CC}). Even cross-algorithm agreement
is therefore largely corpus-specific: each shared core is dominated by pieces
frequent in \emph{that} corpus, not a universal chemical vocabulary the two
algorithms converge on.

\section{Robustness to non-canonical SMILES}\label{app:robustness}\label{app:noncanon}

\begin{figure}[!b]
  \centering
  \includegraphics[width=\linewidth]{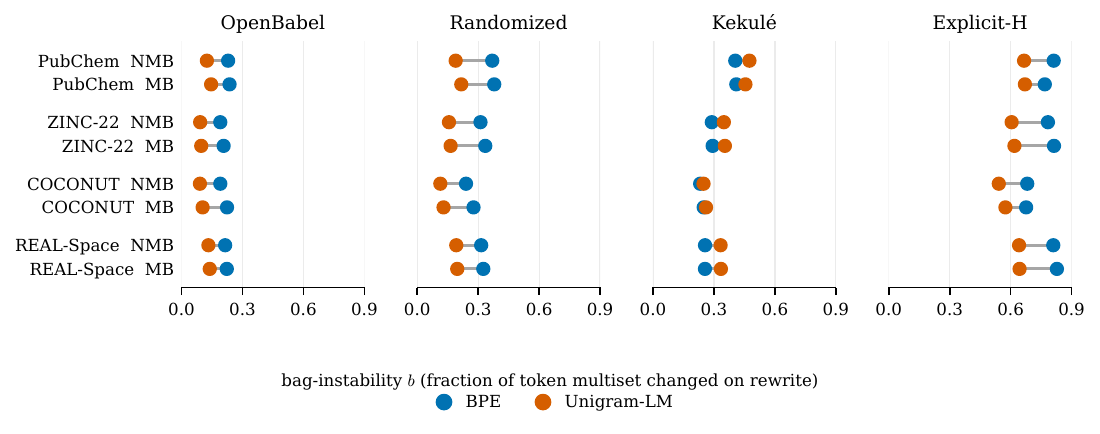}
  \caption{Write-stability under non-canonical SMILES at $V{=}1024$, one dumbbell
  per corpus$\times$boundary joining BPE (blue) to Unigram-LM (orange) by
  bag-instability (fraction of the token multiset that changes on rewrite), with
  the rewrite axes on a shared scale from mildest (OpenBabel) to catastrophic
  (Explicit-H). The arm gap flips sign by axis: Unigram-LM is the more write-stable
  arm (its dot left of BPE's) under randomization and the cross-toolkit OpenBabel
  swap, but the \emph{less} stable under Kekul\'e (dot to the right), which
  destroys the aromatic-lowercase pieces it leans on. Explicit hydrogens are
  catastrophic for both arms. All $22$ conditions, with CIs, are in
  Table~\ref{tab:results-noncanon}.}
  \label{fig:noncanon}
\end{figure}

\input{tables/noncanon_detail}

Every measurement so far reads the tokenizers on the corpus's \emph{canonical}
held-out strings, yet a molecule has many equivalent SMILES, and a deployed model
sees non-canonical ones routinely (a different toolkit's canonical form, or the
randomized strings used for augmentation). We ask, model-free on the trained
families, what happens to each arm's segmentation across that rewrite orbit, and
whether the two algorithms differ. For a seeded subsample of each held-out split
($10{,}000$ molecules) we build five identity-preserving rewrites per molecule.
Four are RDKit-internal, ordered mild to catastrophic to match the difficulty
spectrum a chemical \emph{model} shows: a ring-closure-digit relabel; a Kekul\'e
form; $K$ randomized SMILES (RDKit's default \emph{restricted} randomization, the
augmentation-realistic distribution; the unrestricted variant is a strict
superset and would only widen the gaps); and an all-explicit-hydrogen form
(\texttt{C} to \texttt{[CH4]}). The fifth swaps in OpenBabel's canonical SMILES in
place of RDKit's, gated to identity-preservation by a round-trip back through
RDKit; unlike the four rewrites of RDKit's own output, it is a second
canonicalizer's legitimate form, the most realistic non-canonical input a
multi-source pipeline meets, and a molecule on which the two toolkits agree
contributes a genuine zero. For each arm and axis we measure two senses of
movement against the canonical string: the relative fertility dispersion (does
the token \emph{count} move) and the \emph{bag-instability}, the fraction of the
token multiset that changes (does the set of \emph{pieces} move), each with a
$95\%$ molecule-resampled bootstrap CI. Table~\ref{tab:results-noncanon} shows the
bag-instability per axis and the cross-arm relative fertility gap on canonical
versus randomized strings ($g_{\mathrm{c}}$, $g_{\mathrm{r}}$); the per-arm
dispersions are deposited but not tabulated.

Five readings. First, the granularity gap is not a canonical-notation artifact:
read on the randomized orbit, Unigram-LM stays $23$--$35\%$ finer than BPE in
every one of the $22$ matched conditions, attenuating only slightly from the
canonical gap. Both arms' absolute fertility rises off-canonical (canonical
strings are structurally simpler, with fewer branches and brackets), so the
reported canonical fertilities lower-bound what a deployed model emits; the
relative gap itself is marginally \emph{smaller} on the randomized orbit
($g_{\mathrm{r}} < g_{\mathrm{c}}$ in $21$ of $22$ cells), so the canonical
gap is, if anything, a slight over-estimate of the deployed one. Second,
Unigram-LM is the more write-stable arm: its randomization bag-instability runs
$0.11$--$0.22$ against BPE's $0.24$--$0.38$, lower in all $22$ cells. The
mechanism is the same greedy-versus-probabilistic split the body traces: BPE
applies a fixed merge \emph{sequence}, so reordering the glyph stream changes
which pairs merge first and the realized pieces shift; Unigram-LM scores each
piece independently and Viterbi-decodes, a near-context-free segmentation that
moves less. Third, the arm ranking \emph{flips} by axis (Figure~\ref{fig:noncanon}): under
Kekul\'e, Unigram-LM is the \emph{less} stable arm (bag-instability above BPE's in
$21$ of $22$ cells),
because Kekul\'e destroys every aromatic-lowercase piece and Unigram-LM leans
hardest on aromatic runs (Appendix~\ref{app:composition}). So which algorithm is more
robust depends on the rewrite, a richer answer than a single ranking.

Fourth, the tokenizer's difficulty ordering diverges from a model's. The
ring-digit relabel leaves the token count \emph{exactly} invariant (a clean
floor), and Kekul\'e barely moves it, yet Kekul\'e is the most piece-destructive of
the mild axes: it preserves how \emph{many} tokens but rewrites \emph{which}
ones, a transform a model finds easy. Explicit hydrogens are catastrophic for
both arms (bag-instability $0.54$--$0.83$, with the token count inflating
several-fold), the worst axis at the tokenizer level as it is for models, because
every bare atom becomes a bracketed atom the learned pieces rarely fired on in an
implicit-hydrogen training corpus, the same embedding-tail starvation the
dead-zone confirmation measures (\S\ref{ssec:vregime}); the merge-brackets
policy softens it for BPE, which can then absorb bracket-internal glyphs, tying
this axis back to the boundary axis of the grid.

Fifth, the write-stability gap survives the most realistic non-canonical input of
all: a second toolkit's canonical form. Swapping each molecule's RDKit canonical
SMILES for OpenBabel's, Unigram-LM's bag-instability runs $0.09$--$0.15$ against BPE's
$0.19$--$0.24$, lower in all $22$ cells, so an independent canonicalizer moves
BPE's pieces about twice as far as Unigram-LM's. This is the strongest form of the
stability finding: the gap is not an artifact of RDKit's own randomizer or rewrite
rules, but holds against a genuinely separate implementation that a multi-source
corpus encounters whenever its records were canonicalized by different software.
The swap is milder than full randomization (a single canonical form rather than
the average of $K$ random rewrites) yet sits well above the ring-relabel floor,
and like randomization shifts \emph{which} pieces are chosen, not how many.

\section{Per-condition measurement tables}\label{app:tables}

The body reports the headline values; the exhaustive
per-condition measurements behind them are collected here, the summary
tables first and then four families of measurement-specific detail. Tables tied
to a specific analysis stay with it instead: compositional closure and
functional-bond locality in Appendix~\ref{app:structural}, non-canonical rewrites
in Appendix~\ref{app:robustness}.

\subsection{Overlap, vocabulary, and summary scalars}\label{app:summary}

Table~\ref{tab:results-jaccards} gives the four vocabulary-overlap
weightings, Table~\ref{tab:results-realized-vocab} the realized per-arm
multi-glyph vocabulary, Table~\ref{tab:results-delta-f} the dead-zone surplus
with per-arm $F_{95\%,100}$ clearance, and Table~\ref{tab:results-extras} the
robustness probes, the one table here that steps outside the per-condition grid.
These expand the direct-contrast columns of the body's
Table~\ref{tab:results-seven}. All are at the SentencePiece default $L{=}16$.

The two structural variants drop \emph{bracket-internal} pieces (those whose
every training occurrence falls inside a bracketed atom) and recompute the
overlap over the survivors: $J_{\mathrm{struct}}$ the unweighted Jaccard
(Eq.~\ref{eq:jaccard}) and $J_{\mathrm{w,struct}}$ the frequency-weighted one
(Eq.~\ref{eq:jw}), each arm's weights renormalized over the structural
held-out mass and carrying its own bootstrap CI.

\input{tables/three_jaccards}

\begin{figure}[htbp]
  \centering
  \includegraphics[width=\linewidth]{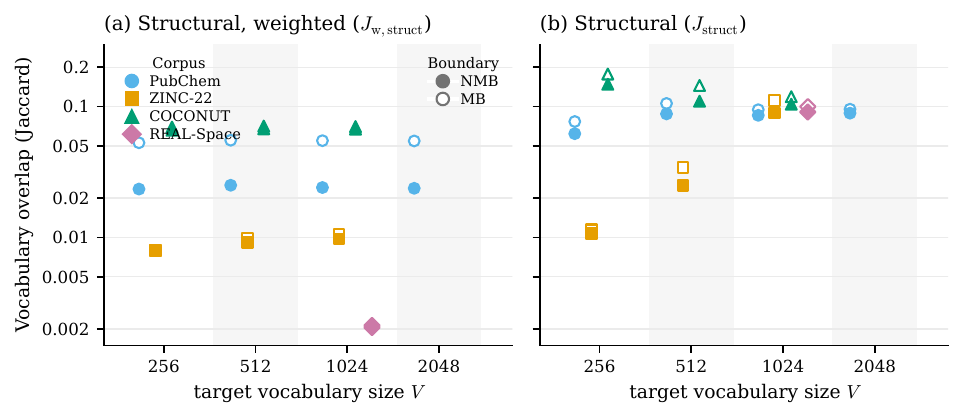}
  \caption{Structural-subword overlap for all $22$ matched conditions, the
  appendix companion to Fig.~\ref{fig:overlap}: frequency-weighted
  ($J_{\mathrm{w,struct}}$, left) and unweighted
  ($J_{\mathrm{struct}}$, right), recomputed after dropping bracket-internal
  pieces. Same encoding as Fig.~\ref{fig:overlap}. Restricting to structural
  pieces barely moves the overlap (every condition stays near-disjoint), so the
  near-disjointness is genuine cross-pretoken
  disagreement, not an artifact of how bracketed atoms are split.}
  \label{fig:overlap-struct}
\end{figure}

\input{tables/realized_vocab}

\begin{figure}[htbp]
  \centering
  \includegraphics[width=\linewidth]{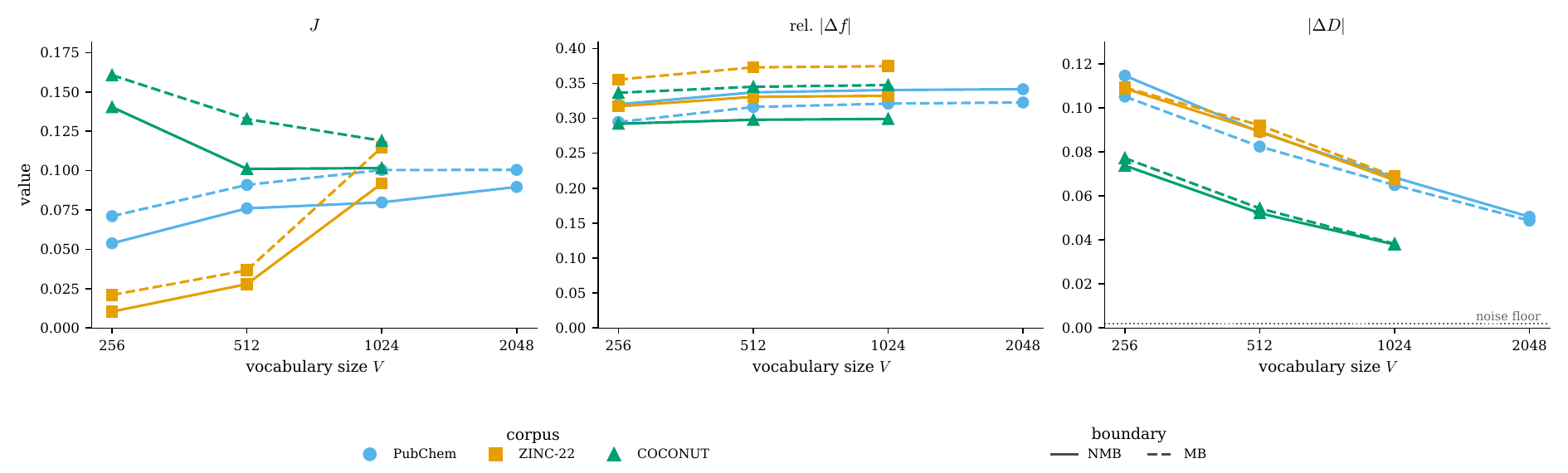}
  \caption{Cross-$V$ trends of the three direct contrasts across the three
  multi-$V$ corpora (the single-$V$ REAL-Space anchor is omitted), the visual
  companion to Table~\ref{tab:results-seven}; color and
  marker give corpus, solid vs.\ dashed the boundary policy (NMB, MB). Overlap
  rises with $V$ on PubChem and ZINC-22 and falls on COCONUT, staying
  near-disjoint throughout; the relative fertility gap rises slightly with
  $V$ and stays large; the imbalance gap $|\Delta D|$ shrinks monotonically with
  $V$ but stays above the noise floor at every $V$.}
  \label{fig:cross-v}
\end{figure}

\input{tables/delta_f}
\input{tables/robustness_extras}

\subsection{Fertility and compression detail}\label{app:fertility}

\input{tables/fertility_detail}

The body (\S\ref{ssec:r-jaccard}) and the headline table
(Table~\ref{tab:results-seven}) report the
granularity contrast only as the \emph{relative} gap rel$|\Delta f|$. This
appendix gives the absolute numbers behind it, the granularity counterpart to
the membership inventory of Appendix~\ref{app:vocab}.
Table~\ref{tab:results-fertility} lists, for every matched condition, each arm's
mean held-out tokens per molecule (with its bootstrap CI), the glyphs-per-token
compression ratio defined in \S\ref{ssec:measurements}, and the absolute and
relative cross-arm gaps; the per-arm fertilities are shown as a dumbbell chart in
the body (Figure~\ref{fig:fertility-curves}).

Two things the relative gap alone hides. First, the absolute gap is large:
Unigram-LM emits $\sim$$12$--$19$ more tokens per molecule than BPE,
on molecules only $\sim$$30$--$55$ tokens long under BPE. Second, the
per-arm CIs are tight: the interval spans at most $\sim$$0.2$ token on the
diverse, drug-like, and combinatorial corpora and $\sim$$1$ token even on the
smaller natural-products corpus, so the $12$--$19$-token gap clears measurement
noise by more than an order of magnitude in every condition. This held-out CI is
the one the cross-arm scalar table (Table~\ref{tab:results-seven}) omits.

\subsection{Boundary nestedness detail}\label{app:nestedness}

\input{tables/nestedness_detail}

The granularity contrast has a positional reading (\S\ref{ssec:measurements}):
because both arms cut the same glyph stream, their boundaries are comparable
position by position. Table~\ref{tab:results-nestedness} gives, for every matched
condition, the boundary Jaccard $J_{\partial}$, the nest and conflict rates, the
fraction of molecules whose BPE parse strictly coarsens Unigram-LM's, and the
per-class conflict localization. Two readings stand out. First, the arms agree on
most cuts and nest rather than cross: conflict is below $0.7\%$ of positions in
every condition, and the nested-molecule fraction exceeds $0.95$ at every
$V{\ge}1024$ cell while staying above $0.80$ even at $V{=}256$, so the fertility
gap is depth, not disagreement. Second, the residual conflict is chemically
structured rather than uniform across substructure classes (heteroatom- and
unsaturation-led on the diverse and natural-products corpora, backbone-led on
alkane-rich ZINC-22), which is why we localize it by class rather than report a
single conflict rate.

\subsection{Token-distribution detail}\label{app:distribution}

\begin{figure}[htbp]
  \centering
  \includegraphics[width=\linewidth]{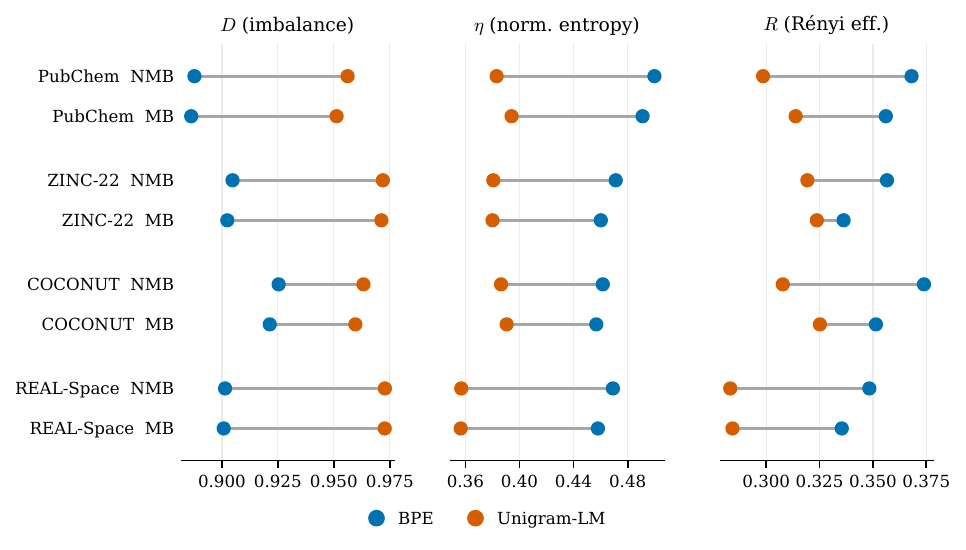}
  \caption{Token-distribution intrinsics at $V{=}1024$, one dumbbell per
  corpus$\times$boundary joining BPE (blue) to Unigram-LM (orange) on each
  metric's own axis: imbalance $D$ (divergence from uniform; left), normalized
  Shannon entropy $\eta$ (center), and R\'enyi efficiency at $\alpha{=}2.5$
  (right). BPE is the more uniform arm in every condition and on all three
  metrics (lower $D$, higher $\eta$, higher $R$), a coherent within-family
  signature, while both arms sit far from uniform ($D \gtrsim 0.85$). Per-condition
  values with CIs in Table~\ref{tab:results-distribution}.}
  \label{fig:distribution-intrinsics}
\end{figure}

\input{tables/distribution_detail}

The distribution sense is reported in the body only through the cross-arm gap
$|\Delta D|$ (Table~\ref{tab:results-seven}); the
three within-family quantities defined in \S\ref{ssec:measurements}
(token-frequency imbalance $D$, normalized Shannon entropy $\eta$, and R\'enyi
efficiency at $\alpha{=}2.5$) are collected per arm here.
Table~\ref{tab:results-distribution} gives all three for every matched condition,
each with its bootstrap CI; Figure~\ref{fig:distribution-intrinsics} plots
them at $V{=}1024$.

Two readings stand out. First, the three metrics form a \emph{coherent} signature: in every
one of the $22$ conditions BPE is the more uniform arm (lower $D$, higher $\eta$,
higher $R$), so the distribution difference, though the weakest of the three
senses, is consistent in sign across all three measures and not an artifact of the
particular $D$ statistic we report. Second, both arms sit \emph{far}
from uniform: absolute $D$ runs $0.76$--$0.97$, roughly $1.5$--$2\times$ the
$\sim$$0.5$ at which \citet{wadell2026smirk} report the chemistry-tokenizer field
clustering. That offset is partly the small-vocabulary regime (a few pieces carry
most of the mass) and partly definitional, and we flag it rather than read across
it: what we report is the cross-arm gap, not the absolute level. The per-arm
CIs are tight ($\le 0.001$ on the diverse, drug-like, and combinatorial corpora),
so the arm separation on each metric far exceeds its CI.

\subsection{Mechanism-diagnostic detail}\label{app:mechanism}

Two of the four mechanism diagnostics are reported in the body only through a
cross-arm gap; their per-arm detail is collected here: whole-pretoken absorption
below, and the dead-zone clearance, whose headline $c_{100}$
(Table~\ref{tab:results-delta-f}) we extend to the full $n$-sweep. The other two,
scaffold fraction and segmentation entropy, are one-sided by construction, so the
gap reported in Table~\ref{tab:results-seven} \emph{is} the absolute.

\input{tables/absorption_detail}

\paragraph{Whole-pretoken absorption.}
Table~\ref{tab:results-absorption} gives each arm's held-out fraction of pretokens
emitted as a single token, with CIs. The contrast is stark and the body's
$\Delta$abs hides it: BPE absorbs almost every pretoken whole
($0.89$--$1.00$ across conditions) while Unigram-LM keeps only
$\sim$$0.72$ ($0.71$--$0.74$) intact, splitting the rest. This is the
per-pretoken face of the fertility gap (\S\ref{ssec:r-mechanism}); the permeable
(MB) boundary lifts BPE's absorption further, toward $1.0$, since merges may then
cross the bracket.

\input{tables/deadzone_nsweep}

\paragraph{Rare-token clearance sweep.}
Table~\ref{tab:results-nsweep} gives the per-arm clearance $c_n$ (fraction of
vocabulary firing $\ge n$ times) at $n \in \{50, 100, 200\}$, the sweep behind the
learnability bar of \S\ref{ssec:vregime}. The headline $F_{95\%,100}$ is the
$c_{100} \ge 0.95$ column; reading any $c_n$ against $p \in \{0.90, 0.95, 0.99\}$
recovers the full $(p,n)$ grid, so the bar's robustness to both thresholds is
visible here. BPE clears the bar at $n{\le}100$ in every condition except its smallest corpus
(COCONUT) at $V{=}1024$; the stricter $n{=}200$ pulls a few narrow-alphabet
ZINC-22 cells just under as well ($0.935$--$0.948$). Unigram-LM's near-atomic
pieces thin it much more on the smaller corpora, the regime where the dead-zone
sign reverses (\S\ref{ssec:r-mechanism}).

\end{document}

%% file: tables/corpora.tex
\begin{table}[htbp]
  \centering
  \caption{The four corpora: three headline typologies plus the REAL-Space
  anchor. Counts are post-preprocessing totals, after any subsampling and
  before the 5\% held-out split.}
  \label{tab:corpora}
  \small
  \begin{tabular}{@{}l l l r l@{}}
    \toprule
    Corpus & Role & Typology & Molecules & Distinguishing feature \\
    \midrule
    PubChem \citep{pubchem2023} & headline & diverse & $\sim$50M & wide alphabet, exotic-atom tail \\
    ZINC-22 \citep{zinc22} & headline & drug-like & $\sim$10M & $\sim$30 atomic types, homogeneous \\
    COCONUT \citep{sorokina2021coconut} & headline & natural-products & $\sim$740K & scaffold-rich, elaborate ring systems \\
    \midrule
    REAL-Space \citep{grygorenko2020real, wadell2026smirk} & anchor & drug-like-narrow & $\sim$136M & combinatorial; super-saturated \\
    \bottomrule
  \end{tabular}
\end{table}

%% file: tables/seven_measurements.tex
\begin{table}[htbp]
  \centering
  \caption{Cross-algorithm measurements per condition (point estimates). \emph{Direct contrasts:} $J$ vocabulary overlap (Jaccard), rel$|\Delta f|$ relative fertility gap, $|\Delta D|$ token-imbalance gap. \emph{Mechanism diagnostics:} $\Delta c_{100}$ dead-zone surplus (cross-arm clearance gap at the $F_{95\%,100}$ bar), $\Delta$abs whole-pretoken-absorption gap, $\Delta$scaf scaffold-fraction gap, $\Delta H_g$ segmentation-entropy gap per glyph. Two-sided $\Delta c_{100}$ and $\Delta$abs are signed BPE $-$ Unigram; one-sided $\Delta$scaf (BPE-only) and $\Delta H_g$ (Unigram-only) print the non-zero arm's magnitude. Bootstrap CIs for the held-out scalars are in the Appendix~\ref{app:tables} detail tables; the exact-set $J$ and $\Delta$scaf carry none. Largest rel$|\Delta f|$ in \textbf{bold}.}
  \label{tab:results-seven}
  \footnotesize
  \begin{tabular}{@{}l r l r r r r r r r@{}}
    \toprule
    & & & \multicolumn{3}{c}{Direct contrasts} & \multicolumn{4}{c}{Mechanism diagnostics} \\
    \cmidrule(lr){4-6}\cmidrule(lr){7-10}
    Corpus & $V$ & Bnd & $J$ & rel$|\Delta f|$ & $|\Delta D|$ & $\Delta c_{100}$ & $\Delta$abs & $\Delta$scaf & $\Delta H_g$ \\
    \midrule
    PubChem & 256 & NMB & 0.054 & 32.0\% & 0.115 & $+0.000$ & $+0.228$ & $+0.004$ & $+0.009$ \\
     & 256 & MB & 0.071 & 29.5\% & 0.105 & $-0.021$ & $+0.237$ & $+0.004$ & $+0.010$ \\
     & 512 & NMB & 0.076 & 33.7\% & 0.089 & $+0.003$ & $+0.243$ & $+0.004$ & $+0.009$ \\
     & 512 & MB & 0.091 & 31.6\% & 0.082 & $-0.006$ & $+0.255$ & $+0.006$ & $+0.010$ \\
     & 1024 & NMB & 0.080 & 34.0\% & 0.068 & $+0.043$ & $+0.246$ & $+0.002$ & $+0.009$ \\
     & 1024 & MB & 0.100 & 32.1\% & 0.065 & $+0.012$ & $+0.259$ & $+0.004$ & $+0.010$ \\
     & 2048 & NMB & 0.090 & 34.2\% & 0.050 & $+0.429$ & $+0.247$ & $+0.004$ & $+0.009$ \\
     & 2048 & MB & 0.100 & 32.3\% & 0.049 & $+0.339$ & $+0.260$ & $+0.001$ & $+0.010$ \\
    \addlinespace
    ZINC-22 & 256 & NMB & 0.010 & 31.7\% & 0.109 & $+0.000$ & $+0.216$ & $+0.000$ & $+0.001$ \\
     & 256 & MB & 0.021 & 35.5\% & 0.109 & $-0.041$ & $+0.266$ & $+0.000$ & $+0.001$ \\
     & 512 & NMB & 0.028 & 33.1\% & 0.089 & $+0.432$ & $+0.229$ & $+0.000$ & $+0.001$ \\
     & 512 & MB & 0.037 & 37.3\% & 0.092 & $+0.385$ & $+0.282$ & $+0.000$ & $+0.001$ \\
     & 1024 & NMB & 0.092 & 33.2\% & 0.067 & $+0.431$ & $+0.231$ & $+0.000$ & $+0.001$ \\
     & 1024 & MB & 0.115 & 37.5\% & 0.069 & $+0.398$ & $+0.284$ & $+0.000$ & $+0.001$ \\
    \addlinespace
    COCONUT & 256 & NMB & 0.140 & 29.2\% & 0.074 & $+0.103$ & $+0.166$ & $+0.004$ & $+0.008$ \\
     & 256 & MB & 0.161 & 33.7\% & 0.077 & $-0.021$ & $+0.266$ & $+0.004$ & $+0.009$ \\
     & 512 & NMB & 0.101 & 29.8\% & 0.052 & $+0.722$ & $+0.171$ & $+0.000$ & $+0.009$ \\
     & 512 & MB & 0.133 & 34.5\% & 0.054 & $+0.603$ & $+0.274$ & $+0.000$ & $+0.009$ \\
     & 1024 & NMB & 0.102 & 29.9\% & 0.038 & $+0.415$ & $+0.173$ & $+0.004$ & $+0.009$ \\
     & 1024 & MB & 0.119 & 34.8\% & 0.038 & $+0.408$ & $+0.275$ & $+0.002$ & $+0.009$ \\
    \addlinespace
    REAL-Space & 1024 & NMB & 0.105 & 36.7\% & 0.071 & $+0.296$ & $+0.262$ & $+0.000$ & $+0.001$ \\
     & 1024 & MB & 0.110 & \textbf{41.0\%} & 0.072 & $+0.280$ & $+0.291$ & $+0.000$ & $+0.001$ \\
    \bottomrule
  \end{tabular}
\end{table}

%% file: tables/transfer_matrix.tex
\begin{table}[htbp]
  \centering
  \caption{Cross-corpus transfer: off-domain fertility ($V{=}1024$, NMB), each cell BPE\,/\,Unigram-LM. Each value is the fertility of the \emph{train}-corpus tokenizer on the \emph{eval} corpus's held-out split, divided by the eval corpus's native tokenizer of the same arm; self-transfer on the diagonal is omitted (trivially $1.00$). BPE transfers at near-native fertility everywhere; Unigram-LM is modestly domain-sensitive, largest where the combinatorial REAL-Space specialist meets natural products (REAL-Space$\to$COCONUT, in bold). Atom-level OOV is below $0.01\%$ in every cell (shared $165$-token base).}
  \label{tab:transfer}
  \small
  \begin{tabular}{@{}l cccc@{}}
    \toprule
    train $\downarrow$ / eval $\rightarrow$ & PubChem & ZINC-22 & COCONUT & REAL-Space \\
    \midrule
    PubChem & --- & 1.00\,/\,0.99 & 1.00\,/\,1.02 & 1.00\,/\,0.96 \\
    ZINC-22 & 1.00\,/\,1.00 & --- & 1.01\,/\,1.03 & 1.00\,/\,0.99 \\
    COCONUT & 1.00\,/\,0.94 & 1.01\,/\,0.96 & --- & 1.00\,/\,0.94 \\
    REAL-Space & 1.00\,/\,1.05 & 1.00\,/\,1.01 & 1.01\,/\,\textbf{1.08} & --- \\
    \bottomrule
  \end{tabular}
\end{table}

%% file: tables/ood_eval.tex
\begin{table}[htbp]
  \centering
  \caption{Out-of-distribution generalization: the PubChem diverse-corpus generalist ($V{=}1024$, NMB) read on adversarial in-spec chemistry it was not fit to. Fertility is mean tokens per molecule; bootstrap CIs negligible at these $n$. rel$|\Delta f|$ is the cross-arm relative fertility gap (Eq.~\ref{eq:fertility}). The BPE-coarser / Unigram-finer divergence persists off-distribution, and atom-level OOV stays negligible (shared $165$-token base), so the contrast is genuine granularity, not coverage.}
  \label{tab:ood-eval}
  \small
  \begin{tabular}{@{}l r rr r rr@{}}
    \toprule
    & & \multicolumn{2}{c}{fertility} & & \multicolumn{2}{c}{atom OOV}\\
    \cmidrule(lr){3-4}\cmidrule(lr){6-7}
    corpus & $n$ & BPE & Unigram-LM & rel$|\Delta f|$ & BPE & Unigram-LM \\
    \midrule
    tmQM (metals) & 97,379 & 65.5 & 82.9 & 23.5\% & 0.002\% & 0.002\% \\
    CycPeptMPDB (macrocycles) & 7,988 & 111.3 & 149.7 & 29.5\% & 0.000\% & 0.000\% \\
    \bottomrule
  \end{tabular}
\end{table}

%% file: tables/hyperparameters.tex
\begin{table}[htbp]
  \centering
  \caption{Tokenizer training hyperparameters, by arm. Each numeric knob is
  held at its reference-implementation default for the headline grid and swept
  across a ladder bracketing that default for the sensitivity analysis
  (\S\ref{ssec:grid}); \textbf{bold} marks the reference default within each
  ladder. Design-forced settings depart from the SentencePiece reference by
  construction, not tuning (\S\ref{ssec:training}). Reference defaults are the
  published library defaults: SentencePiece's \texttt{TrainerSpec}
  (\texttt{sentencepiece\_model.proto}) and Sennrich's subword-nmt
  (\texttt{learn\_bpe.py}, \texttt{-{}-min-frequency}).}
  \label{tab:hyperparams}
  \small
  \begin{tabular}{@{}l l l l l@{}}
    \toprule
    Parameter & Role & Ref.\ default & Headline & Sensitivity ladder \\
    \midrule
    \multicolumn{5}{@{}l}{\emph{Unigram-LM} (SentencePiece reference \citep{kudo2018sentencepiece})} \\
    \texttt{max\_piece\_length} & max glyphs per piece & $16$ & $16$ & $4,8,\mathbf{16},32,64,128$ \\
    \texttt{seed\_size} & seed-pool cap & $10^6$ & $10^6$ & $(2.5,5,\mathbf{10},20,40,80)\!\times\!10^5$ \\
    \texttt{n\_sub\_iterations} & EM iters per prune & $2$ & $2$ & $1,\mathbf{2},3,4$ \\
    \texttt{shrinking\_factor} & kept fraction per prune & $0.75$ & $0.75$ & $0.5,0.6,\mathbf{0.75},0.9,0.95$ \\
    \midrule
    \multicolumn{5}{@{}l}{\emph{BPE} (subword-nmt reference \citep{sennrich2016bpe-nmt})} \\
    \texttt{min\_frequency} & min pair count to merge & $2$ & $2$ & $0,1,\mathbf{2},4,8$ \\
    \midrule
    \multicolumn{5}{@{}l}{\emph{Shared design axes}} \\
    \texttt{vocab\_size} $V$ & target size & --- & $256/512/1024$ & $+2048$; anchor $1024$ \\
    boundary & bracket permeability & --- & NMB\,/\,MB & --- \\
    \midrule
    \multicolumn{5}{@{}l}{\emph{Design-forced} (deviate from SentencePiece by construction)} \\
    \texttt{character\_coverage} & OOV coverage & $0.9995$ & $1.0$ & --- \\
    \texttt{num\_threads} & parallelism & $16$ & $1$ & --- \\
    pre-tokenization & string splitting & whitespace\,/\,script & Smirk Layer-A/B & --- \\
    \bottomrule
  \end{tabular}
\end{table}

%% file: tables/base_glyphs.tex
\begin{table}[htbp]
  \centering
  \caption{The complete Smirk base: the 165-token OpenSMILES glyph alphabet (158 chemistry-grammatical glyphs and 7 special tokens), grouped by OpenSMILES role. Any OpenSMILES-conformant string decomposes into these glyphs with no \texttt{[UNK]}. Inside a bracket atom \texttt{+} and \texttt{-} also denote charge, and \texttt{-} doubles as the single bond.}
  \label{tab:base-glyphs}
  \small
  \begin{tabular}{@{}p{0.27\linewidth} p{0.66\linewidth}@{}}
    \toprule
    Group & Glyphs \\
    \midrule
    Aliphatic organic-subset atoms (10) & \texttt{B}, \texttt{C}, \texttt{N}, \texttt{O}, \texttt{P}, \texttt{S}, \texttt{F}, \texttt{Cl}, \texttt{Br}, \texttt{I} \\
    \addlinespace[3pt]
    Aromatic atoms (8) & \texttt{b}, \texttt{c}, \texttt{n}, \texttt{o}, \texttt{p}, \texttt{s}, \texttt{se}, \texttt{as} \\
    \addlinespace[3pt]
    Other element symbols (bracket atoms) (108) & \texttt{Ac}, \texttt{Ag}, \texttt{Al}, \texttt{Am}, \texttt{Ar}, \texttt{As}, \texttt{At}, \texttt{Au}, \texttt{Ba}, \texttt{Be}, \texttt{Bh}, \texttt{Bi}, \texttt{Bk}, \texttt{Ca}, \texttt{Cd}, \texttt{Ce}, \texttt{Cf}, \texttt{Cm}, \texttt{Cn}, \texttt{Co}, \texttt{Cr}, \texttt{Cs}, \texttt{Cu}, \texttt{Db}, \texttt{Ds}, \texttt{Dy}, \texttt{Er}, \texttt{Es}, \texttt{Eu}, \texttt{Fe}, \texttt{Fl}, \texttt{Fm}, \texttt{Fr}, \texttt{Ga}, \texttt{Gd}, \texttt{Ge}, \texttt{H}, \texttt{He}, \texttt{Hf}, \texttt{Hg}, \texttt{Ho}, \texttt{Hs}, \texttt{In}, \texttt{Ir}, \texttt{K}, \texttt{Kr}, \texttt{La}, \texttt{Li}, \texttt{Lr}, \texttt{Lu}, \texttt{Lv}, \texttt{Mc}, \texttt{Md}, \texttt{Mg}, \texttt{Mn}, \texttt{Mo}, \texttt{Mt}, \texttt{Na}, \texttt{Nb}, \texttt{Nd}, \texttt{Ne}, \texttt{Nh}, \texttt{Ni}, \texttt{No}, \texttt{Np}, \texttt{Og}, \texttt{Os}, \texttt{Pa}, \texttt{Pb}, \texttt{Pd}, \texttt{Pm}, \texttt{Po}, \texttt{Pr}, \texttt{Pt}, \texttt{Pu}, \texttt{Ra}, \texttt{Rb}, \texttt{Re}, \texttt{Rf}, \texttt{Rg}, \texttt{Rh}, \texttt{Rn}, \texttt{Ru}, \texttt{Sb}, \texttt{Sc}, \texttt{Se}, \texttt{Sg}, \texttt{Si}, \texttt{Sm}, \texttt{Sn}, \texttt{Sr}, \texttt{Ta}, \texttt{Tb}, \texttt{Tc}, \texttt{Te}, \texttt{Th}, \texttt{Ti}, \texttt{Tl}, \texttt{Tm}, \texttt{Ts}, \texttt{U}, \texttt{V}, \texttt{W}, \texttt{Xe}, \texttt{Y}, \texttt{Yb}, \texttt{Zn}, \texttt{Zr} \\
    \addlinespace[3pt]
    Bonds (7) & \texttt{-}, \texttt{=}, \texttt{\#}, \texttt{\$}, \texttt{:}, \texttt{/}, \texttt{\textbackslash{}} \\
    \addlinespace[3pt]
    Branches, rings, ring-closures (16) & \texttt{(}, \texttt{)}, \texttt{[}, \texttt{]}, \texttt{.}, \texttt{\%}, \texttt{0}, \texttt{1}, \texttt{2}, \texttt{3}, \texttt{4}, \texttt{5}, \texttt{6}, \texttt{7}, \texttt{8}, \texttt{9} \\
    \addlinespace[3pt]
    Charge and wildcard (2) & \texttt{+}, \texttt{*} \\
    \addlinespace[3pt]
    Chirality (7) & \texttt{@}, \texttt{@@}, \texttt{@TH}, \texttt{@AL}, \texttt{@OH}, \texttt{@SP}, \texttt{@TB} \\
    \addlinespace[3pt]
    Special tokens (7) & \texttt{[UNK]}, \texttt{[BOS]}, \texttt{[EOS]}, \texttt{[SEP]}, \texttt{[PAD]}, \texttt{[CLS]}, \texttt{[MASK]} \\
    \bottomrule
  \end{tabular}
\end{table}

%% file: tables/multiglyph_v256.tex
\providecommand{\bndtok}[1]{\textcolor{black!50}{#1}}
\begin{table}[htbp]
  \centering
  \caption{Learned multi-glyph pieces for the PubChem $V{=}256$ matched pair, split three ways by cross-algorithm membership (shared by both arms, selected only by BPE, and selected only by Unigram-LM) under each boundary policy, with the per-policy overlap $J$ in the block header. Each piece is its concatenated glyph surface, ordered by glyph count then alphabetically; pieces also selected under the other boundary policy (\emph{boundary-robust}: selected under both) are \bndtok{grayed}, leaving the policy-specific pieces plain. Classes larger than 18 pieces are abridged to their first 14 and last 4 (the parenthetical gives the full class size); complete lists are in the data deposit.}
  \label{tab:multiglyph-v256}
  \footnotesize
  \begin{tabular}{@{}l p{0.78\linewidth}@{}}
    \toprule
    Membership & Pieces \\
    \midrule
    \multicolumn{2}{@{}l}{\emph{no-merge-brackets (NMB)}, overlap $J{=}0.054$} \\
    \addlinespace[2pt]
    Shared (10) & \bndtok{\texttt{C\#}}, \bndtok{\texttt{CCCC}}, \bndtok{\texttt{CCCN}}, \bndtok{\texttt{CCNC}}, \bndtok{\texttt{CCOC}}, \bndtok{\texttt{CCCCC}}, \bndtok{\texttt{CCOCC}}, \bndtok{\texttt{CCCCCCC}}, \bndtok{\texttt{CCCCCCCC}}, \bndtok{\texttt{CCCCCCCCCCCCCCCC}} \\
    \addlinespace[3pt]
    BPE-only (89) & \bndtok{\texttt{-c}}, \bndtok{\texttt{-n}}, \texttt{2H}, \bndtok{\texttt{=C}}, \bndtok{\texttt{=N}}, \bndtok{\texttt{=O}}, \bndtok{\texttt{=S}}, \bndtok{\texttt{C=}}, \texttt{C@}, \texttt{C@@}, \bndtok{\texttt{CC}}, \texttt{CH}, \bndtok{\texttt{CN}}, \bndtok{\texttt{CO}}, \textellipsis~(71 more), \bndtok{\texttt{cccnc}}, \bndtok{\texttt{ccncc}}, \bndtok{\texttt{CCCCCC}}, \texttt{OCCOCC} \\
    \addlinespace[3pt]
    Unigram-LM-only (87) & \bndtok{\texttt{10}}, \texttt{Ca+}, \texttt{Cu+}, \texttt{Fe+}, \texttt{Ir+}, \texttt{Mg+}, \texttt{Ni+}, \bndtok{\texttt{OP}}, \texttt{Pd+}, \texttt{Pt+}, \texttt{Ru+}, \texttt{Ti+}, \texttt{Zn+}, \texttt{Zr+}, \textellipsis~(69 more), \bndtok{\texttt{=S=S=S=S=S=S=S=S}}, \texttt{CCC=CCC=CCC=CCC=}, \texttt{CCCCCCC=CCCCCCCC}, \bndtok{\texttt{OCCOCCOCCOCCOCCO}} \\
    \midrule
    \multicolumn{2}{@{}l}{\emph{merge-brackets (MB)}, overlap $J{=}0.071$} \\
    \addlinespace[2pt]
    Shared (13) & \bndtok{\texttt{C\#}}, \texttt{[Si}, \texttt{[C@}, \texttt{[C@@}, \bndtok{\texttt{CCCC}}, \bndtok{\texttt{CCCN}}, \bndtok{\texttt{CCNC}}, \bndtok{\texttt{CCOC}}, \bndtok{\texttt{CCCCC}}, \bndtok{\texttt{CCOCC}}, \bndtok{\texttt{ccccc}}, \bndtok{\texttt{CCCCCCCC}}, \bndtok{\texttt{CCCCCCCCCCCCCCCC}} \\
    \addlinespace[3pt]
    BPE-only (86) & \texttt{+]}, \texttt{-]}, \bndtok{\texttt{-c}}, \bndtok{\texttt{-n}}, \bndtok{\texttt{=C}}, \bndtok{\texttt{=N}}, \bndtok{\texttt{=O}}, \bndtok{\texttt{=S}}, \bndtok{\texttt{C=}}, \bndtok{\texttt{CC}}, \bndtok{\texttt{CN}}, \bndtok{\texttt{CO}}, \bndtok{\texttt{CS}}, \bndtok{\texttt{Cc}}, \textellipsis~(68 more), \bndtok{\texttt{ccccn}}, \bndtok{\texttt{cccnc}}, \bndtok{\texttt{ccncc}}, \bndtok{\texttt{CCCCCC}} \\
    \addlinespace[3pt]
    Unigram-LM-only (84) & \bndtok{\texttt{10}}, \bndtok{\texttt{OP}}, \texttt{[Al}, \texttt{[Fe}, \texttt{[Ge}, \texttt{[Ir}, \texttt{[K}, \texttt{[Li}, \texttt{[Na}, \texttt{[Pt}, \texttt{[Se}, \texttt{[Sn}, \texttt{[Ti}, \texttt{[V}, \textellipsis~(66 more), \bndtok{\texttt{=C=C=C=C=C=C=C=C}}, \bndtok{\texttt{=S=S=S=S=S=S=S=S}}, \texttt{CCOCCOCCOCCOCCOC}, \bndtok{\texttt{OCCOCCOCCOCCOCCO}} \\
    \bottomrule
  \end{tabular}
\end{table}

%% file: tables/shared_core_growth.tex
\providecommand{\bndtok}[1]{\textcolor{black!50}{#1}}
\begin{table}[htbp]
  \centering
  \caption{Growth of the BPE\,$\cap$\,Unigram-LM shared core for the PubChem matched pair under each boundary policy, as the target vocabulary grows $V \in \{256, 512, 1024, 2048\}$. The core is strictly nested (no piece ever leaves), so each row lists only the pieces \emph{newly} shared at that $V$, and the cumulative core at a given $V$ is that row plus all above it. Cumulative core size grows $10 \to 50 \to 128 \to 311$ at overlap $J = 0.054,\, 0.076,\, 0.080,\, 0.090$ (NMB) and $13 \to 59 \to 158 \to 345$ at overlap $J = 0.071,\, 0.091,\, 0.100,\, 0.100$ (MB). Pieces also in the other boundary's shared core (\emph{boundary-robust}: shared under both) are \bndtok{grayed}, leaving the boundary-specific shared pieces plain; pieces are ordered by glyph count then alphabetically. Layers larger than 18 pieces are abridged to their first 14 and last 4 (the parenthetical gives the full layer size); complete lists are in the data deposit.}
  \label{tab:shared-core-growth}
  \footnotesize
  \begin{tabular}{@{}l p{0.79\linewidth}@{}}
    \toprule
    Entered at & Newly shared pieces \\
    \midrule
    \multicolumn{2}{@{}l}{\emph{no-merge-brackets (NMB)}} \\
    \addlinespace[2pt]
    Core, $V\le256$ (10) & \bndtok{\texttt{C\#}}, \bndtok{\texttt{CCCC}}, \bndtok{\texttt{CCCN}}, \bndtok{\texttt{CCNC}}, \bndtok{\texttt{CCOC}}, \bndtok{\texttt{CCCCC}}, \bndtok{\texttt{CCOCC}}, \bndtok{\texttt{CCCCCCC}}, \bndtok{\texttt{CCCCCCCC}}, \bndtok{\texttt{CCCCCCCCCCCCCCCC}} \\
    \addlinespace[3pt]
    New at $V{=}512$ (40) & \bndtok{\texttt{OP}}, \bndtok{\texttt{COP}}, \bndtok{\texttt{N\#C}}, \bndtok{\texttt{O=P}}, \bndtok{\texttt{O=S}}, \texttt{SiH2}, \bndtok{\texttt{nnn}}, \bndtok{\texttt{noc}}, \bndtok{\texttt{C\#CC}}, \bndtok{\texttt{CCCS}}, \bndtok{\texttt{CCNS}}, \bndtok{\texttt{CCSC}}, \bndtok{\texttt{CN=C}}, \bndtok{\texttt{NCCN}}, \textellipsis~(22 more), \bndtok{\texttt{CCCCCCCCCCCC}}, \bndtok{\texttt{CCCCCCCCCCCCC}}, \bndtok{\texttt{CCCCCCCCCCCCCC}}, \bndtok{\texttt{CCCCCCCCCCCCCCC}} \\
    \addlinespace[3pt]
    New at $V{=}1024$ (78) & \texttt{Au+}, \texttt{Cu+}, \bndtok{\texttt{II}}, \bndtok{\texttt{PP}}, \texttt{18F}, \bndtok{\texttt{=NP}}, \texttt{AlH3}, \bndtok{\texttt{N=P}}, \texttt{SiH3}, \bndtok{\texttt{CCSS}}, \bndtok{\texttt{CSSC}}, \bndtok{\texttt{N=NN}}, \bndtok{\texttt{NN=C}}, \bndtok{\texttt{SCCN}}, \textellipsis~(60 more), \bndtok{\texttt{CCCCCCCCCCOC}}, \bndtok{\texttt{CCCCCCCCCCCCOC}}, \bndtok{\texttt{=C=C=C=C=C=C=C=C}}, \bndtok{\texttt{=S=S=S=S=S=S=S=S}} \\
    \addlinespace[3pt]
    New at $V{=}2048$ (183) & \texttt{Al+}, \texttt{AlH}, \texttt{AsH}, \texttt{Hg+}, \bndtok{\texttt{PI}}, \texttt{Pd+}, \texttt{Ru+}, \texttt{Sn+}, \texttt{Zn+}, \bndtok{\texttt{op}}, \texttt{pH}, \texttt{AlH2}, \texttt{AsH2}, \bndtok{\texttt{B=B}}, \textellipsis~(165 more), \bndtok{\texttt{CCOCCOCCOCCOCCNC}}, \bndtok{\texttt{NCCCOCCOCCOCCCNC}}, \bndtok{\texttt{NCCOCCOCCOCCOCCC}}, \bndtok{\texttt{OCCOCCOCCOCCOCCO}} \\
    \midrule
    \multicolumn{2}{@{}l}{\emph{merge-brackets (MB)}} \\
    \addlinespace[2pt]
    Core, $V\le256$ (13) & \bndtok{\texttt{C\#}}, \texttt{[Si}, \texttt{[C@}, \texttt{[C@@}, \bndtok{\texttt{CCCC}}, \bndtok{\texttt{CCCN}}, \bndtok{\texttt{CCNC}}, \bndtok{\texttt{CCOC}}, \bndtok{\texttt{CCCCC}}, \bndtok{\texttt{CCOCC}}, \texttt{ccccc}, \bndtok{\texttt{CCCCCCCC}}, \bndtok{\texttt{CCCCCCCCCCCCCCCC}} \\
    \addlinespace[3pt]
    New at $V{=}512$ (46) & \bndtok{\texttt{OP}}, \texttt{[Ir}, \texttt{[K}, \texttt{[Li}, \texttt{[Na}, \texttt{[Pt}, \texttt{[Sn}, \texttt{[Y}, \bndtok{\texttt{COP}}, \bndtok{\texttt{N\#C}}, \bndtok{\texttt{O=P}}, \bndtok{\texttt{O=S}}, \texttt{[SiH}, \bndtok{\texttt{nnn}}, \textellipsis~(28 more), \bndtok{\texttt{CCCCCCCCCCCC}}, \bndtok{\texttt{CCCCCCCCCCCCC}}, \bndtok{\texttt{CCCCCCCCCCCCCC}}, \bndtok{\texttt{CCCCCCCCCCCCCCC}} \\
    \addlinespace[3pt]
    New at $V{=}1024$ (99) & \texttt{[Ag}, \texttt{[Al}, \texttt{[As}, \texttt{[Au}, \texttt{[Ba}, \texttt{[Bi}, \texttt{[Ca}, \texttt{[Co}, \texttt{[Cr}, \texttt{[Cs}, \texttt{[Cu}, \texttt{[Fe}, \texttt{[Ge}, \texttt{[Hf}, \textellipsis~(81 more), \bndtok{\texttt{OCCOCCOCCOC}}, \bndtok{\texttt{CCCCCCCCCCCCOC}}, \bndtok{\texttt{=C=C=C=C=C=C=C=C}}, \bndtok{\texttt{=S=S=S=S=S=S=S=S}} \\
    \addlinespace[3pt]
    New at $V{=}2048$ (187) & \bndtok{\texttt{II}}, \bndtok{\texttt{PI}}, \bndtok{\texttt{PP}}, \texttt{[Be}, \texttt{[Cd}, \texttt{[Ce}, \texttt{[Eu}, \texttt{[Ga}, \texttt{[Gd}, \texttt{[Hg}, \texttt{[In}, \texttt{[La}, \texttt{[Nb}, \texttt{[Nd}, \textellipsis~(169 more), \bndtok{\texttt{CCOCCOCCOCCOCCNC}}, \bndtok{\texttt{NCCCOCCOCCOCCCNC}}, \bndtok{\texttt{NCCOCCOCCOCCOCCC}}, \bndtok{\texttt{OCCOCCOCCOCCOCCO}} \\
    \bottomrule
  \end{tabular}
\end{table}

%% file: tables/arm_exclusive.tex
\begin{table}[htbp]
  \centering
  \caption{The arm-exclusive multi-glyph sets for the PubChem matched pair across $V$: counts and glyph-length statistics of the pieces each arm selects but the other does not. The two arms select near-equal \emph{numbers} of exclusive pieces at every $V$, and these dwarf the shared core (Table~\ref{tab:shared-core-growth}), so the near-disjointness persists at every scale. The Unigram-LM (UL) exclusive pieces never exceed $16$ glyphs (its \texttt{max\_piece\_length}, Table~\ref{tab:hyperparams}), whereas BPE imposes no length cap, so its exclusive pieces run far longer (degenerate periodic chains such as a $96$-glyph \texttt{OCCOCC}$\dots$ at $V{=}2048$).}
  \label{tab:arm-exclusive}
  \small
  \begin{tabular}{@{}r r r r r r r@{}}
    \toprule
    & \multicolumn{2}{c}{Exclusive pieces} & \multicolumn{2}{c}{Mean glyph len.} & \multicolumn{2}{c}{Max glyph len.} \\
    \cmidrule(lr){2-3}\cmidrule(lr){4-5}\cmidrule(lr){6-7}
    $V$ & BPE & UL & BPE & UL & BPE & UL \\
    \midrule
    \multicolumn{7}{@{}l}{\emph{No-merge-brackets (NMB)}} \\
    256 & 89 & 87 & 2.9 & 6.3 & 6 & 16 \\
    512 & 305 & 303 & 3.8 & 7.6 & 19 & 16 \\
    1024 & 739 & 737 & 5.0 & 9.0 & 48 & 16 \\
    2048 & 1,581 & 1,578 & 6.3 & 10.5 & 96 & 16 \\
    \addlinespace[2pt]
    \multicolumn{7}{@{}l}{\emph{Merge-brackets (MB)}} \\
    256 & 86 & 84 & 3.0 & 5.1 & 6 & 16 \\
    512 & 296 & 294 & 3.7 & 6.2 & 18 & 16 \\
    1024 & 709 & 707 & 4.7 & 8.1 & 48 & 16 \\
    2048 & 1,546 & 1,544 & 5.9 & 10.0 & 96 & 16 \\
    \bottomrule
  \end{tabular}
\end{table}

%% file: tables/composition.tex
\begin{table}[htbp]
  \centering
  \caption{Substructure composition of the shared, BPE-only, and Unigram-LM-only multi-glyph sets, by corpus (matched pairs at $V{=}1024$, NMB). Each piece is classed from its glyphs by priority aromatic atom $>$ aliphatic heteroatom $>$ unsaturated carbon $>$ saturated carbon, and each row gives the percentage of that set's pieces in each class. Aromatic-ring pieces are a BPE specialty the Unigram-LM arm almost never forms; bracket-internal pieces arise only under MB (\S\ref{app:multiglyph}) and do not occur here.}
  \label{tab:composition}
  \small
  \begin{tabular}{@{}l r r r r r@{}}
    \toprule
    Set & $n$ & Sat.\,C & Unsat.\,C & Aromatic & Heteroatom \\
    \midrule
    \multicolumn{6}{@{}l}{\emph{PubChem}} \\
    Shared & 128 & 9\% & 9\% & 11\% & 71\% \\
    BPE-only & 739 & 6\% & 6\% & 34\% & 54\% \\
    Unigram-LM-only & 737 & 0\% & 13\% & 1\% & 86\% \\
    \addlinespace[2pt]
    \multicolumn{6}{@{}l}{\emph{ZINC-22}} \\
    Shared & 99 & 3\% & 9\% & 2\% & 86\% \\
    BPE-only & 768 & 1\% & 6\% & 37\% & 56\% \\
    Unigram-LM-only & 211 & 1\% & 7\% & 0\% & 92\% \\
    \addlinespace[2pt]
    \multicolumn{6}{@{}l}{\emph{COCONUT}} \\
    Shared & 139 & 8\% & 22\% & 30\% & 40\% \\
    BPE-only & 728 & 6\% & 18\% & 27\% & 49\% \\
    Unigram-LM-only & 501 & 0\% & 49\% & 7\% & 44\% \\
    \bottomrule
  \end{tabular}
\end{table}

%% file: tables/closure_detail.tex
\begin{table}[htbp]
  \centering
  \caption{Per-condition within-arm compositional closure. Per matched pair (\S\ref{ssec:measurements}), read from the realized vocabulary alone. $c_{\mathrm{bin}}$ is the binary-split closure: the fraction of multi-glyph pieces with some in-vocab split $p{=}a{\cdot}b$ (both parts in the base-plus-multi vocabulary). It is BPE's merge-closure invariant read off the realized set, so $c_{\mathrm{bin}}^{\mathrm{BPE}}{=}1$ exactly (the correctness anchor) and the orphan rate $c_{\mathrm{orph}}^{\mathrm{BPE}}{=}0$, the latter omitted for BPE. $c_{\mathrm{orph}}^{\mathrm{UL}}$ is the Unigram-LM (UL) orphan rate: the fraction of its length-$\ge 3$ pieces with \emph{no} proper $\ge 2$-glyph sub-piece in vocabulary (pieces that share no building block with the rest of the vocabulary). $c_{\mathrm{full}}$ is the stronger full-substring closure (every $\ge 2$-glyph substring in vocabulary), non-trivial for both arms. Closure is an exact-set quantity and carries no CI. Unigram-LM is far less self-referential than BPE in every condition: roughly half its pieces (as few as $0.12$ on the narrowest alphabet) do not decompose into in-vocab parts.}
  \label{tab:results-closure}
  \footnotesize
  \begin{tabular}{@{}l r l r r r r r@{}}
    \toprule
    & & & \multicolumn{2}{c}{$c_{\mathrm{bin}}$} & & \multicolumn{2}{c}{$c_{\mathrm{full}}$} \\
    \cmidrule(lr){4-5}\cmidrule(lr){7-8}
    Corpus & $V$ & Bnd & BPE & UL & $c_{\mathrm{orph}}^{\mathrm{UL}}$ & BPE & UL \\
    \midrule
    PubChem & 256 & NMB & 1.000 & 0.392 & 0.451 & 0.767 & 0.000 \\
     & 256 & MB & 1.000 & 0.361 & 0.628 & 0.613 & 0.000 \\
     & 512 & NMB & 1.000 & 0.448 & 0.285 & 0.763 & 0.003 \\
     & 512 & MB & 1.000 & 0.510 & 0.341 & 0.657 & 0.003 \\
     & 1024 & NMB & 1.000 & 0.482 & 0.181 & 0.713 & 0.002 \\
     & 1024 & MB & 1.000 & 0.534 & 0.207 & 0.601 & 0.001 \\
     & 2048 & NMB & 1.000 & 0.445 & 0.121 & 0.678 & 0.004 \\
     & 2048 & MB & 1.000 & 0.470 & 0.139 & 0.610 & 0.004 \\
    \addlinespace
    ZINC-22 & 256 & NMB & 1.000 & 0.124 & 0.608 & 0.714 & 0.000 \\
     & 256 & MB & 1.000 & 0.216 & 0.567 & 0.645 & 0.000 \\
     & 512 & NMB & 1.000 & 0.426 & 0.329 & 0.706 & 0.000 \\
     & 512 & MB & 1.000 & 0.482 & 0.269 & 0.690 & 0.000 \\
     & 1024 & NMB & 1.000 & 0.426 & 0.329 & 0.627 & 0.000 \\
     & 1024 & MB & 1.000 & 0.499 & 0.266 & 0.613 & 0.000 \\
    \addlinespace
    COCONUT & 256 & NMB & 1.000 & 0.536 & 0.360 & 0.701 & 0.047 \\
     & 256 & MB & 1.000 & 0.536 & 0.464 & 0.621 & 0.024 \\
     & 512 & NMB & 1.000 & 0.431 & 0.205 & 0.686 & 0.021 \\
     & 512 & MB & 1.000 & 0.482 & 0.240 & 0.624 & 0.019 \\
     & 1024 & NMB & 1.000 & 0.589 & 0.138 & 0.604 & 0.013 \\
     & 1024 & MB & 1.000 & 0.636 & 0.119 & 0.552 & 0.011 \\
    \addlinespace
    REAL-Space & 1024 & NMB & 1.000 & 0.464 & 0.296 & 0.596 & 0.000 \\
     & 1024 & MB & 1.000 & 0.444 & 0.314 & 0.594 & 0.000 \\
    \bottomrule
  \end{tabular}
\end{table}

%% file: tables/fg_alignment_detail.tex
\begin{table}[htbp]
  \centering
  \caption{Per-condition within-arm chemical functional-bond locality. Per matched pair (\S\ref{ssec:measurements}), read on the held-out split. A \emph{functional bond} is a multiply-bonded heteroatom, a non-carbon atom joined by a double or triple bond (the $=$O of a carbonyl, the $\#$N of a nitrile, the $=$N of an imine, the $=$O on sulfur, phosphorus, or nitrogen, the $=$S of a thiocarbonyl), the cores of the canonical functional groups, read straight off the molecular graph. Locality $\ell$ is the fraction of those bonds the arm keeps inside a single token (the heteroatom sharing a token with its bond glyph); $\Delta\ell{=}\ell^{\mathrm{BPE}}{-}\ell^{\mathrm{UL}}$. $\ell_{\mathrm{C{=}O}}$ isolates the carbonyl class. BPE keeps nearly every functional bond local (the carbonyl essentially always); Unigram-LM (UL) keeps almost none (the carbonyl column $\ell^{\mathrm{UL}}_{\mathrm{C{=}O}}$ is $0.000$ to three places in every condition), spending its unsaturation budget on long homo-atomic carbon runs, not on binding the heteroatom to its bond. 95\% molecule-resampled bootstrap CIs are in the deposited per-condition records.}
  \label{tab:results-fg-alignment}
  \footnotesize
  \begin{tabular}{@{}l r l r r r r r@{}}
    \toprule
    & & & \multicolumn{2}{c}{$\ell$} & & \multicolumn{2}{c}{$\ell_{\mathrm{C{=}O}}$} \\
    \cmidrule(lr){4-5}\cmidrule(lr){7-8}
    Corpus & $V$ & Bnd & BPE & UL & $\Delta\ell$ & BPE & UL \\
    \midrule
    PubChem & 256 & NMB & 0.961 & 0.023 & 0.938 & 0.999 & 0.000 \\
     & 256 & MB & 0.948 & 0.022 & 0.927 & 0.995 & 0.000 \\
     & 512 & NMB & 0.988 & 0.026 & 0.962 & 1.000 & 0.000 \\
     & 512 & MB & 0.987 & 0.026 & 0.961 & 0.999 & 0.000 \\
     & 1024 & NMB & 0.991 & 0.027 & 0.964 & 1.000 & 0.000 \\
     & 1024 & MB & 0.990 & 0.027 & 0.963 & 1.000 & 0.000 \\
     & 2048 & NMB & 0.991 & 0.027 & 0.965 & 1.000 & 0.000 \\
     & 2048 & MB & 0.991 & 0.027 & 0.964 & 1.000 & 0.000 \\
    \addlinespace
    ZINC-22 & 256 & NMB & 0.992 & 0.007 & 0.985 & 1.000 & 0.000 \\
     & 256 & MB & 0.985 & 0.007 & 0.977 & 1.000 & 0.000 \\
     & 512 & NMB & 0.998 & 0.007 & 0.991 & 1.000 & 0.000 \\
     & 512 & MB & 0.998 & 0.007 & 0.990 & 1.000 & 0.000 \\
     & 1024 & NMB & 0.998 & 0.007 & 0.991 & 1.000 & 0.000 \\
     & 1024 & MB & 0.998 & 0.007 & 0.991 & 1.000 & 0.000 \\
    \addlinespace
    COCONUT & 256 & NMB & 0.987 & 0.004 & 0.983 & 0.999 & 0.000 \\
     & 256 & MB & 0.982 & 0.004 & 0.978 & 0.999 & 0.000 \\
     & 512 & NMB & 0.997 & 0.005 & 0.992 & 1.000 & 0.000 \\
     & 512 & MB & 0.996 & 0.006 & 0.990 & 1.000 & 0.000 \\
     & 1024 & NMB & 0.998 & 0.005 & 0.993 & 1.000 & 0.000 \\
     & 1024 & MB & 0.998 & 0.006 & 0.991 & 1.000 & 0.000 \\
    \addlinespace
    REAL-Space & 1024 & NMB & 0.995 & 0.001 & 0.994 & 1.000 & 0.000 \\
     & 1024 & MB & 0.995 & 0.001 & 0.994 & 1.000 & 0.000 \\
    \bottomrule
  \end{tabular}
\end{table}

%% file: tables/narrow_contrast.tex
\begin{table}[htbp]
  \centering
  \caption{The ZINC-22 matched pair across $V$: each arm's multi-glyph vocabulary size, the three-way cross-algorithm split (shared / BPE-only / Unigram-LM-only), and the overlap $J$, under both boundary policies (UL denotes Unigram-LM).}
  \label{tab:narrow-contrast}
  \small
  \begin{tabular}{@{}r r r r r r r@{}}
    \toprule
    & \multicolumn{2}{c}{Multi-glyph vocab.} & \multicolumn{3}{c}{Cross-arm split} & \\
    \cmidrule(lr){2-3}\cmidrule(lr){4-6}
    $V$ & BPE & UL & Shared & BPE-only & UL-only & $J$ \\
    \midrule
    \multicolumn{7}{@{}l}{\emph{No-merge-brackets (NMB)}} \\
    256 & 98 & 97 & 2 & 96 & 95 & 0.010 \\
    512 & 355 & 310 & 18 & 337 & 292 & 0.028 \\
    1024 & 867 & 310 & 99 & 768 & 211 & 0.092 \\
    \addlinespace[2pt]
    \multicolumn{7}{@{}l}{\emph{Merge-brackets (MB)}} \\
    256 & 98 & 97 & 4 & 94 & 93 & 0.021 \\
    512 & 355 & 353 & 25 & 330 & 328 & 0.037 \\
    1024 & 867 & 357 & 126 & 741 & 231 & 0.115 \\
    \bottomrule
  \end{tabular}
\end{table}

%% file: tables/coconut_contrast.tex
\begin{table}[htbp]
  \centering
  \caption{The COCONUT matched pair across $V$: each arm's multi-glyph vocabulary size, the three-way cross-algorithm split (shared / BPE-only / Unigram-LM-only), and the overlap $J$, under both boundary policies (UL denotes Unigram-LM).}
  \label{tab:coconut-contrast}
  \small
  \begin{tabular}{@{}r r r r r r r@{}}
    \toprule
    & \multicolumn{2}{c}{Multi-glyph vocab.} & \multicolumn{3}{c}{Cross-arm split} & \\
    \cmidrule(lr){2-3}\cmidrule(lr){4-6}
    $V$ & BPE & UL & Shared & BPE-only & UL-only & $J$ \\
    \midrule
    \multicolumn{7}{@{}l}{\emph{No-merge-brackets (NMB)}} \\
    256 & 98 & 97 & 24 & 74 & 73 & 0.140 \\
    512 & 355 & 353 & 65 & 290 & 288 & 0.101 \\
    1024 & 867 & 640 & 139 & 728 & 501 & 0.102 \\
    \addlinespace[2pt]
    \multicolumn{7}{@{}l}{\emph{Merge-brackets (MB)}} \\
    256 & 98 & 97 & 27 & 71 & 70 & 0.161 \\
    512 & 355 & 353 & 83 & 272 & 270 & 0.133 \\
    1024 & 867 & 769 & 174 & 693 & 595 & 0.119 \\
    \bottomrule
  \end{tabular}
\end{table}

%% file: tables/noncanon_detail.tex
\begin{table}[htbp]
  \centering
  \caption{Per-condition within-arm robustness to non-canonical SMILES. Per matched pair (\S\ref{ssec:measurements}), on a seeded held-out subsample; the paired columns $b^{\mathrm{BPE}}$ and $b^{\mathrm{UL}}$ are the BPE and Unigram-LM (UL) arms. $b$ is the \emph{bag-instability}: the fraction of an arm's token multiset that changes when the molecule is rewritten (mean $1$ minus the multiset Jaccard versus the canonical string), under four identity-preserving rewrite axes: Randomized (RDKit restricted randomization, the augmentation-realistic distribution of \citet{aruspous2019randomized-smiles}), Kekul\'e, Explicit-H (all-explicit-hydrogen, AMORE's catastrophic axis \citep{ganeeva2025chemllm-robustness}), and OpenBabel (the cross-toolkit swap to OpenBabel's canonical SMILES, gated to identity-preservation by a round-trip through RDKit). $g_{\mathrm{c}}$ and $g_{\mathrm{r}}$ are the relative fertility gap $\mathrm{rel}|\Delta f|$ (Eq.~\ref{eq:fertility}) on the canonical strings and on the randomized orbit; their closeness shows the granularity gap survives the orbit. A ring-digit-relabel floor (omitted) leaves the token count exactly invariant. Figure~\ref{fig:noncanon} plots the per-axis arm-stability pattern at $V{=}1024$; 95\% molecule-resampled bootstrap CIs are in the deposited per-condition records.}
  \label{tab:results-noncanon}
  \footnotesize
  \resizebox{\ifdim\width>\linewidth\linewidth\else\width\fi}{!}{%
  \begin{tabular}{@{}l r l r r r r r r r r r r@{}}
    \toprule
    & & & \multicolumn{2}{c}{Randomized} & \multicolumn{2}{c}{Kekul\'e} & \multicolumn{2}{c}{Explicit-H} & \multicolumn{2}{c}{OpenBabel} & \multicolumn{2}{c}{Fertility gap} \\
    \cmidrule(lr){4-5}\cmidrule(lr){6-7}\cmidrule(lr){8-9}\cmidrule(lr){10-11}\cmidrule(lr){12-13}
    Corpus & $V$ & Bnd & $b^{\mathrm{BPE}}$ & $b^{\mathrm{UL}}$ & $b^{\mathrm{BPE}}$ & $b^{\mathrm{UL}}$ & $b^{\mathrm{BPE}}$ & $b^{\mathrm{UL}}$ & $b^{\mathrm{BPE}}$ & $b^{\mathrm{UL}}$ & $g_{\mathrm{c}}$ & $g_{\mathrm{r}}$ \\
    \midrule
    PubChem & 256 & NMB & 0.351 & 0.193 & 0.413 & 0.477 & 0.813 & 0.665 & 0.219 & 0.126 & 28.4\% & 24.8\% \\
     & 256 & MB & 0.358 & 0.219 & 0.420 & 0.475 & 0.824 & 0.669 & 0.224 & 0.145 & 25.6\% & 23.4\% \\
     & 512 & NMB & 0.375 & 0.189 & 0.408 & 0.475 & 0.811 & 0.666 & 0.230 & 0.125 & 32.3\% & 27.4\% \\
     & 512 & MB & 0.384 & 0.218 & 0.413 & 0.455 & 0.770 & 0.670 & 0.237 & 0.146 & 30.3\% & 26.9\% \\
     & 1024 & NMB & 0.369 & 0.189 & 0.405 & 0.475 & 0.812 & 0.666 & 0.229 & 0.125 & 33.4\% & 28.6\% \\
     & 1024 & MB & 0.379 & 0.217 & 0.410 & 0.455 & 0.768 & 0.670 & 0.236 & 0.146 & 31.5\% & 28.2\% \\
     & 2048 & NMB & 0.367 & 0.189 & 0.397 & 0.475 & 0.812 & 0.666 & 0.229 & 0.125 & 33.7\% & 29.0\% \\
     & 2048 & MB & 0.377 & 0.217 & 0.402 & 0.455 & 0.761 & 0.670 & 0.236 & 0.146 & 32.0\% & 28.5\% \\
    \addlinespace
    ZINC-22 & 256 & NMB & 0.311 & 0.156 & 0.298 & 0.348 & 0.781 & 0.604 & 0.189 & 0.092 & 28.5\% & 27.3\% \\
     & 256 & MB & 0.333 & 0.167 & 0.303 & 0.359 & 0.814 & 0.618 & 0.204 & 0.099 & 31.9\% & 30.6\% \\
     & 512 & NMB & 0.315 & 0.156 & 0.293 & 0.349 & 0.783 & 0.605 & 0.192 & 0.092 & 31.9\% & 29.7\% \\
     & 512 & MB & 0.338 & 0.164 & 0.298 & 0.360 & 0.813 & 0.619 & 0.208 & 0.097 & 35.7\% & 33.0\% \\
     & 1024 & NMB & 0.311 & 0.156 & 0.290 & 0.349 & 0.784 & 0.605 & 0.191 & 0.092 & 32.5\% & 30.5\% \\
     & 1024 & MB & 0.335 & 0.164 & 0.294 & 0.354 & 0.813 & 0.619 & 0.207 & 0.097 & 36.3\% & 33.9\% \\
    \addlinespace
    COCONUT & 256 & NMB & 0.239 & 0.113 & 0.252 & 0.249 & 0.704 & 0.541 & 0.190 & 0.090 & 29.2\% & 26.9\% \\
     & 256 & MB & 0.274 & 0.128 & 0.272 & 0.272 & 0.775 & 0.574 & 0.222 & 0.103 & 34.4\% & 31.4\% \\
     & 512 & NMB & 0.240 & 0.114 & 0.240 & 0.249 & 0.703 & 0.542 & 0.191 & 0.091 & 29.6\% & 27.3\% \\
     & 512 & MB & 0.277 & 0.129 & 0.258 & 0.261 & 0.760 & 0.574 & 0.224 & 0.104 & 35.0\% & 32.0\% \\
     & 1024 & NMB & 0.240 & 0.114 & 0.232 & 0.249 & 0.682 & 0.542 & 0.191 & 0.091 & 29.7\% & 27.4\% \\
     & 1024 & MB & 0.277 & 0.129 & 0.249 & 0.261 & 0.676 & 0.574 & 0.224 & 0.104 & 35.1\% & 32.1\% \\
    \addlinespace
    REAL-Space & 1024 & NMB & 0.315 & 0.192 & 0.256 & 0.333 & 0.810 & 0.641 & 0.215 & 0.132 & 31.9\% & 32.0\% \\
     & 1024 & MB & 0.325 & 0.197 & 0.256 & 0.334 & 0.828 & 0.644 & 0.223 & 0.139 & 35.2\% & 34.9\% \\
    \bottomrule
  \end{tabular}%
  }
\end{table}

%% file: tables/three_jaccards.tex
\begin{table}[htbp]
  \centering
  \caption{Cross-algorithm vocabulary overlap (Jaccard). $J$ unweighted, $J_{\mathrm{struct}}$ structural-subword, $J_{\mathrm{w}}$ frequency-weighted, $J_{\mathrm{w,struct}}$ frequency-weighted over structural subwords only; all four are vocabulary Jaccards on matched arm pairs. $J$ and $J_{\mathrm{struct}}$ are exact set quantities; the molecule-resampled bootstrap CIs on $J_{\mathrm{w}}$ and $J_{\mathrm{w,struct}}$ span $\le 0.002$ in every condition and are deposited in full. The smallest and largest $J$ are in \textbf{bold}.}
  \label{tab:results-jaccards}
  \footnotesize
  \begin{tabular}{@{}l r l r r r r@{}}
    \toprule
    & & & \multicolumn{2}{c}{Unweighted} & \multicolumn{2}{c}{Freq-weighted} \\
    \cmidrule(lr){4-5}\cmidrule(lr){6-7}
    Corpus & $V$ & Bnd & $J$ & $J_{\mathrm{struct}}$ & $J_{\mathrm{w}}$ & $J_{\mathrm{w,struct}}$ \\
    \midrule
    PubChem & 256 & NMB & 0.054 & 0.062 & 0.021 & 0.023 \\
     & 256 & MB & 0.071 & 0.077 & 0.048 & 0.053 \\
     & 512 & NMB & 0.076 & 0.088 & 0.023 & 0.025 \\
     & 512 & MB & 0.091 & 0.106 & 0.050 & 0.055 \\
     & 1024 & NMB & 0.080 & 0.086 & 0.022 & 0.024 \\
     & 1024 & MB & 0.100 & 0.095 & 0.049 & 0.055 \\
     & 2048 & NMB & 0.090 & 0.089 & 0.022 & 0.024 \\
     & 2048 & MB & 0.100 & 0.095 & 0.049 & 0.055 \\
    \addlinespace
    ZINC-22 & 256 & NMB & \textbf{0.010} & 0.011 & 0.006 & 0.008 \\
     & 256 & MB & 0.021 & 0.011 & 0.006 & 0.008 \\
     & 512 & NMB & 0.028 & 0.025 & 0.008 & 0.009 \\
     & 512 & MB & 0.037 & 0.034 & 0.008 & 0.010 \\
     & 1024 & NMB & 0.092 & 0.089 & 0.008 & 0.010 \\
     & 1024 & MB & 0.115 & 0.111 & 0.009 & 0.011 \\
    \addlinespace
    COCONUT & 256 & NMB & 0.140 & 0.148 & 0.043 & 0.066 \\
     & 256 & MB & \textbf{0.161} & 0.177 & 0.045 & 0.070 \\
     & 512 & NMB & 0.101 & 0.110 & 0.043 & 0.068 \\
     & 512 & MB & 0.133 & 0.145 & 0.046 & 0.071 \\
     & 1024 & NMB & 0.102 & 0.104 & 0.043 & 0.067 \\
     & 1024 & MB & 0.119 & 0.119 & 0.045 & 0.070 \\
    \addlinespace
    REAL-Space & 1024 & NMB & 0.105 & 0.091 & 0.002 & 0.002 \\
     & 1024 & MB & 0.110 & 0.100 & 0.002 & 0.002 \\
    \bottomrule
  \end{tabular}
\end{table}

%% file: tables/realized_vocab.tex
\begin{table}[htbp]
  \centering
  \caption{Realized per-arm multi-glyph vocabulary per condition. Per arm, the sets the Jaccards compare, excluding the shared $165$-token base. BPE keeps the full atomic base and fills the rest with merges to target; Unigram-LM (UL) prunes to at or below target and may shed rarely-used base glyphs, reallocating to multi-glyph pieces, so on diverse corpora it packs comparably many (UL/BPE near $1$). On narrow alphabets at larger $V$ its pruning runs short of high-likelihood pieces and bottoms out well below target (UL/BPE $\ll 1$). The Jaccard is a set ratio robust to this gap; its effect on the overlap ceiling is discussed in \S\ref{ssec:r-jaccard}.}
  \label{tab:results-realized-vocab}
  \footnotesize
  \begin{tabular}{@{}l r l r r r@{}}
    \toprule
    & & & \multicolumn{2}{c}{$|\mathcal{V}^{\mathrm{multi}}|$} & \\
    \cmidrule(lr){4-5}
    Corpus & $V$ & Bnd & BPE & UL & UL/BPE \\
    \midrule
    PubChem & 256 & NMB & 99 & 97 & 0.98 \\
     & 256 & MB & 99 & 97 & 0.98 \\
     & 512 & NMB & 355 & 353 & 0.99 \\
     & 512 & MB & 355 & 353 & 0.99 \\
     & 1024 & NMB & 867 & 865 & 1.00 \\
     & 1024 & MB & 867 & 865 & 1.00 \\
     & 2048 & NMB & 1,892 & 1,889 & 1.00 \\
     & 2048 & MB & 1,891 & 1,889 & 1.00 \\
    \addlinespace
    ZINC-22 & 256 & NMB & 98 & 97 & 0.99 \\
     & 256 & MB & 98 & 97 & 0.99 \\
     & 512 & NMB & 355 & 310 & 0.87 \\
     & 512 & MB & 355 & 353 & 0.99 \\
     & 1024 & NMB & 867 & 310 & 0.36 \\
     & 1024 & MB & 867 & 357 & 0.41 \\
    \addlinespace
    COCONUT & 256 & NMB & 98 & 97 & 0.99 \\
     & 256 & MB & 98 & 97 & 0.99 \\
     & 512 & NMB & 355 & 353 & 0.99 \\
     & 512 & MB & 355 & 353 & 0.99 \\
     & 1024 & NMB & 867 & 640 & 0.74 \\
     & 1024 & MB & 867 & 769 & 0.89 \\
    \addlinespace
    REAL-Space & 1024 & NMB & 867 & 267 & 0.31 \\
     & 1024 & MB & 867 & 293 & 0.34 \\
    \bottomrule
  \end{tabular}
\end{table}

%% file: tables/delta_f.tex
\begin{table}[htbp]
  \centering
  \caption{Dead-zone surplus: per-arm $F_{95\%,100}$ clearance $c_{100}$ and its cross-arm difference $\Delta c_{100}$. $c_{100}^{\mathrm{BPE}}$, $c_{100}^{\mathrm{UL}}$ (columns BPE, UL; UL $=$ Unigram-LM) are the fraction of each arm's vocabulary clearing the $F_{95\%,100}$ rare-token-tail bar; $\Delta c_{100}$ is their difference (BPE $-$ UL). $\dagger$ corpus too small to certify the tail at this $V$.}
  \label{tab:results-delta-f}
  \footnotesize
  \begin{tabular}{@{}l r l r r r@{}}
    \toprule
    & & & \multicolumn{2}{c}{$c_{100}$} & \\
    \cmidrule(lr){4-5}
    Corpus & $V$ & Bnd & BPE & UL & $\Delta c_{100}$ \\
    \midrule
    PubChem & 256 & NMB & 1.000 & 1.000 & $+0.000$ \\
     & 256 & MB & 0.979 & 1.000 & $-0.021$ \\
     & 512 & NMB & 1.000 & 0.997 & $+0.003$ \\
     & 512 & MB & 0.994 & 1.000 & $-0.006$ \\
     & 1024 & NMB & 1.000 & 0.957 & $+0.043$ \\
     & 1024 & MB & 0.991 & 0.979 & $+0.012$ \\
     & 2048$^{\dagger}$ & NMB & 0.997 & 0.567 & $+0.429$ \\
     & 2048$^{\dagger}$ & MB & 0.980 & 0.641 & $+0.339$ \\
    \addlinespace
    ZINC-22 & 256 & NMB & 1.000 & 1.000 & $+0.000$ \\
     & 256 & MB & 0.959 & 1.000 & $-0.041$ \\
     & 512$^{\dagger}$ & NMB & 1.000 & 0.568 & $+0.432$ \\
     & 512$^{\dagger}$ & MB & 0.977 & 0.592 & $+0.385$ \\
     & 1024$^{\dagger}$ & NMB & 0.999 & 0.568 & $+0.431$ \\
     & 1024$^{\dagger}$ & MB & 0.984 & 0.585 & $+0.398$ \\
    \addlinespace
    COCONUT & 256$^{\dagger}$ & NMB & 1.000 & 0.897 & $+0.103$ \\
     & 256 & MB & 0.979 & 1.000 & $-0.021$ \\
     & 512$^{\dagger}$ & NMB & 0.994 & 0.272 & $+0.722$ \\
     & 512$^{\dagger}$ & MB & 0.969 & 0.365 & $+0.603$ \\
     & 1024$^{\dagger}$ & NMB & 0.566 & 0.152 & $+0.415$ \\
     & 1024$^{\dagger}$ & MB & 0.576 & 0.168 & $+0.408$ \\
    \addlinespace
    REAL-Space & 1024$^{\dagger}$ & NMB & 1.000 & 0.704 & $+0.296$ \\
     & 1024$^{\dagger}$ & MB & 0.993 & 0.713 & $+0.280$ \\
    \bottomrule
  \end{tabular}
\end{table}

%% file: tables/robustness_extras.tex
\begin{table}[htbp]
  \centering
  \caption{Robustness extras: subsample, size sweep, seed-cap, prune-schedule, merge-exhaustion. Outside the headline grid. Subsample redraw: Unigram $c_{100}$ across three independent draws of the same corpus. Size sweep: the same cell trained at 5M / 15M / 50M molecules. Seed cap and prune schedule: multi-glyph vocabulary Jaccard of a one-armed hyperparameter probe against its baseline (1 = identical piece set). Merge exhaustion: the realized vocabulary where \texttt{GpeTrainer} terminated naturally below a $50{,}000$ cap.}
  \label{tab:results-extras}
  \footnotesize
  \begin{tabular}{@{}l l l l l@{}}
    \toprule
    Probe & Setting & Metric & Value & Reading \\
    \midrule
    Subsample redraw & PubChem $V$=512 NMB ($\times$3) & Unigram $c_{100}$ & 0.912--0.921 & spread 0.008 \\
     & ZINC-22 $V$=512 NMB ($\times$3) & Unigram $c_{100}$ & 0.497--0.505 & spread 0.009 \\
    Size sweep & PubChem 5/15/50M & Unigram $c_{100}$ & 0.921 / 0.994 / 0.997 & $\uparrow$ with size \\
    Seed cap & PubChem U $V$=1024 MB & multi-glyph $J$ (1e6/8e6) & 1.000 & inert \\
    Prune schedule & PubChem U $V$=256/$V$=512 MB & multi-glyph $J$ (0.75/0.9) & 0.865 / 0.924 & schedule-sensitive \\
    Merge exhaustion & REAL-Space GPE, cap 50,000 & realized $|\mathcal{V}|$ & 4,331 & natural ($<$ cap) \\
    \bottomrule
  \end{tabular}
\end{table}

%% file: tables/fertility_detail.tex
\begin{table}[htbp]
  \centering
  \caption{Per-condition absolute fertility and compression ratio. Per matched pair. $f$ mean held-out tokens per molecule (95\% molecule-resampled bootstrap CI in brackets); $\tfrac{g}{t}$ glyphs per token (the compression ratio of \S\ref{ssec:measurements}, higher $=$ coarser); $|\Delta f|$ the absolute and rel$|\Delta f|$ the relative cross-arm fertility gap (Eq.~\ref{eq:fertility}). Unigram-LM (UL) segments to more tokens at a lower compression ratio than BPE in every condition.}
  \label{tab:results-fertility}
  \footnotesize
  \resizebox{\ifdim\width>\linewidth\linewidth\else\width\fi}{!}{%
  \begin{tabular}{@{}l r l r r r r r r@{}}
    \toprule
    & & & \multicolumn{2}{c}{$f$} & \multicolumn{2}{c}{$\tfrac{g}{t}$} & & \\
    \cmidrule(lr){4-5}\cmidrule(lr){6-7}
    Corpus & $V$ & Bnd & BPE & UL & BPE & UL & $|\Delta f|$ & rel$|\Delta f|$ \\
    \midrule
    PubChem & 256 & NMB & 36.9 [36.9, 37.0] & 51.0 [50.9, 51.1] & 1.45 [1.45, 1.45] & 1.05 [1.05, 1.05] & 14.1 & 32.0\% \\
     & 256 & MB & 35.2 [35.2, 35.3] & 47.4 [47.3, 47.5] & 1.52 [1.52, 1.52] & 1.13 [1.13, 1.13] & 12.2 & 29.5\% \\
     & 512 & NMB & 36.3 [36.2, 36.3] & 51.0 [50.9, 51.1] & 1.47 [1.47, 1.47] & 1.05 [1.05, 1.05] & 14.7 & 33.7\% \\
     & 512 & MB & 34.4 [34.3, 34.5] & 47.3 [47.2, 47.4] & 1.55 [1.55, 1.55] & 1.13 [1.13, 1.13] & 12.9 & 31.6\% \\
     & 1024 & NMB & 36.1 [36.1, 36.2] & 50.9 [50.8, 51.0] & 1.48 [1.48, 1.48] & 1.05 [1.05, 1.05] & 14.8 & 34.0\% \\
     & 1024 & MB & 34.2 [34.1, 34.3] & 47.3 [47.2, 47.4] & 1.56 [1.56, 1.56] & 1.13 [1.13, 1.13] & 13.1 & 32.1\% \\
     & 2048 & NMB & 36.1 [36.0, 36.2] & 50.9 [50.8, 51.0] & 1.48 [1.48, 1.48] & 1.05 [1.05, 1.05] & 14.9 & 34.2\% \\
     & 2048 & MB & 34.1 [34.1, 34.2] & 47.3 [47.2, 47.4] & 1.56 [1.56, 1.57] & 1.13 [1.13, 1.13] & 13.1 & 32.3\% \\
    \addlinespace
    ZINC-22 & 256 & NMB & 32.7 [32.6, 32.7] & 45.0 [45.0, 45.0] & 1.40 [1.40, 1.40] & 1.01 [1.01, 1.01] & 12.3 & 31.7\% \\
     & 256 & MB & 29.6 [29.6, 29.7] & 42.5 [42.4, 42.5] & 1.54 [1.54, 1.54] & 1.07 [1.07, 1.07] & 12.8 & 35.5\% \\
     & 512 & NMB & 32.2 [32.2, 32.2] & 45.0 [44.9, 45.0] & 1.42 [1.42, 1.42] & 1.01 [1.01, 1.01] & 12.8 & 33.1\% \\
     & 512 & MB & 29.1 [29.1, 29.1] & 42.4 [42.4, 42.5] & 1.57 [1.57, 1.57] & 1.07 [1.07, 1.08] & 13.3 & 37.3\% \\
     & 1024 & NMB & 32.2 [32.1, 32.2] & 45.0 [44.9, 45.0] & 1.42 [1.42, 1.42] & 1.01 [1.01, 1.01] & 12.8 & 33.2\% \\
     & 1024 & MB & 29.0 [29.0, 29.1] & 42.4 [42.4, 42.5] & 1.57 [1.57, 1.57] & 1.07 [1.07, 1.08] & 13.4 & 37.5\% \\
    \addlinespace
    COCONUT & 256 & NMB & 55.3 [54.9, 55.7] & 74.2 [73.7, 74.7] & 1.44 [1.44, 1.45] & 1.08 [1.07, 1.08] & 18.9 & 29.2\% \\
     & 256 & MB & 46.0 [45.7, 46.2] & 64.6 [64.1, 65.0] & 1.74 [1.73, 1.74] & 1.24 [1.23, 1.24] & 18.6 & 33.7\% \\
     & 512 & NMB & 54.9 [54.6, 55.3] & 74.2 [73.7, 74.7] & 1.45 [1.45, 1.45] & 1.08 [1.07, 1.08] & 19.2 & 29.8\% \\
     & 512 & MB & 45.5 [45.3, 45.8] & 64.5 [64.1, 64.9] & 1.75 [1.75, 1.76] & 1.24 [1.24, 1.24] & 19.0 & 34.5\% \\
     & 1024 & NMB & 54.9 [54.5, 55.3] & 74.2 [73.7, 74.7] & 1.45 [1.45, 1.46] & 1.08 [1.07, 1.08] & 19.3 & 29.9\% \\
     & 1024 & MB & 45.5 [45.1, 45.8] & 64.6 [64.1, 65.0] & 1.76 [1.75, 1.76] & 1.24 [1.23, 1.24] & 19.1 & 34.8\% \\
    \addlinespace
    REAL-Space & 1024 & NMB & 32.7 [32.7, 32.7] & 47.4 [47.4, 47.4] & 1.46 [1.46, 1.46] & 1.00 [1.00, 1.00] & 14.7 & 36.7\% \\
     & 1024 & MB & 30.9 [30.9, 30.9] & 46.8 [46.8, 46.8] & 1.54 [1.54, 1.54] & 1.02 [1.02, 1.02] & 15.9 & 41.0\% \\
    \bottomrule
  \end{tabular}%
  }
\end{table}

%% file: tables/nestedness_detail.tex
\begin{table}[htbp]
  \centering
  \caption{Per-condition cross-arm boundary nestedness. Per matched pair (\S\ref{ssec:measurements}). Both arms cut the same glyph stream, so their boundaries are subsets of the same inter-glyph positions. $J_{\partial}$ the boundary Jaccard over \emph{cut} positions (agree-cut over agree-cut $+$ nest $+$ conflict); \emph{nest} the share of positions where Unigram-LM cuts and BPE merges (the fertility gap of Table~\ref{tab:results-fertility} read positionally) and \emph{conflict} the share where BPE cuts and Unigram-LM merges (genuine crossing), both over all positions; \emph{nested} the fraction of molecules with zero conflict, i.e.\ whose BPE parse is a strict coarsening of Unigram-LM's. $\mathrm{cut}^{c}$ localizes conflict: the fraction of multi-glyph Unigram-LM pieces of class $c$ (heteroatom, unsaturated carbon, saturated carbon) that BPE cuts through, with no entry for a class the corpus never emits. Conflict is near zero in every condition.}
  \label{tab:results-nestedness}
  \footnotesize
  \begin{tabular}{@{}l r l r r r r r r r@{}}
    \toprule
    & & & & & & & \multicolumn{3}{c}{Conflict $\mathrm{cut}^{c}$} \\
    \cmidrule(lr){8-10}
    Corpus & $V$ & Bnd & $J_{\partial}$ & conflict & nest & nested & het & uns & sat \\
    \midrule
    PubChem & 256 & NMB & 0.709 & 0.0052 & 0.274 & 0.818 & 0.465 & 0.348 & 0.164 \\
     & 256 & MB & 0.726 & 0.0062 & 0.239 & 0.802 & 0.451 & 0.325 & 0.199 \\
     & 512 & NMB & 0.703 & 0.0017 & 0.282 & 0.934 & 0.170 & 0.136 & 0.047 \\
     & 512 & MB & 0.716 & 0.0024 & 0.249 & 0.913 & 0.182 & 0.152 & 0.052 \\
     & 1024 & NMB & 0.702 & 0.0008 & 0.283 & 0.970 & 0.083 & 0.065 & 0.013 \\
     & 1024 & MB & 0.716 & 0.0010 & 0.250 & 0.963 & 0.091 & 0.075 & 0.014 \\
     & 2048 & NMB & 0.702 & 0.0003 & 0.284 & 0.986 & 0.039 & 0.034 & 0.004 \\
     & 2048 & MB & 0.715 & 0.0004 & 0.251 & 0.984 & 0.041 & 0.036 & 0.005 \\
    \addlinespace
    ZINC-22 & 256 & NMB & 0.717 & 0.0015 & 0.277 & 0.951 & 1.000 & 1.000 & 1.000 \\
     & 256 & MB & 0.687 & 0.0019 & 0.289 & 0.939 & 1.000 & 1.000 & 1.000 \\
     & 512 & NMB & 0.709 & 0.0004 & 0.286 & 0.982 & 0.401 & 0.235 & 1.000 \\
     & 512 & MB & 0.677 & 0.0005 & 0.300 & 0.977 & 0.427 & 0.320 & 0.550 \\
     & 1024 & NMB & 0.709 & 0.0001 & 0.287 & 0.998 & 0.054 & 0.032 & 0.183 \\
     & 1024 & MB & 0.677 & 0.0001 & 0.300 & 0.997 & 0.058 & 0.030 & 0.089 \\
    \addlinespace
    COCONUT & 256 & NMB & 0.738 & 0.0021 & 0.242 & 0.889 & 0.649 & 0.680 & 0.059 \\
     & 256 & MB & 0.702 & 0.0027 & 0.239 & 0.858 & 0.751 & 0.701 & 0.053 \\
     & 512 & NMB & 0.736 & 0.0004 & 0.245 & 0.973 & 0.197 & 0.330 & 0.011 \\
     & 512 & MB & 0.700 & 0.0006 & 0.242 & 0.966 & 0.176 & 0.359 & 0.011 \\
     & 1024 & NMB & 0.736 & 0.0001 & 0.245 & 0.993 & 0.053 & 0.072 & 0.004 \\
     & 1024 & MB & 0.699 & 0.0001 & 0.243 & 0.991 & 0.051 & 0.083 & 0.003 \\
    \addlinespace
    REAL-Space & 1024 & NMB & 0.683 & 0.0000 & 0.315 & 0.998 & 0.074 & 0.009 & --- \\
     & 1024 & MB & 0.652 & 0.0001 & 0.342 & 0.997 & 0.086 & 0.010 & --- \\
    \bottomrule
  \end{tabular}
\end{table}

%% file: tables/distribution_detail.tex
\begin{table}[htbp]
  \centering
  \caption{Per-condition token-distribution intrinsics (imbalance, entropy, R\'enyi efficiency). Within-family, per matched pair (\S\ref{ssec:measurements}), each a held-out per-arm value with a 95\% molecule-resampled bootstrap CI. $D$ token-frequency imbalance (divergence from uniform, Eq.~\ref{eq:imbalance}; $0$ uniform, $1$ maximally concentrated), $\eta$ normalized Shannon entropy, $R$ R\'enyi efficiency at $\alpha{=}2.5$. BPE is more uniform than Unigram-LM (UL) in every condition. The cross-arm gap $|\Delta D|$ is in Table~\ref{tab:results-seven}.}
  \label{tab:results-distribution}
  \footnotesize
  \resizebox{\ifdim\width>\linewidth\linewidth\else\width\fi}{!}{%
  \begin{tabular}{@{}l r l r r r r r r@{}}
    \toprule
    & & & \multicolumn{2}{c}{$D$} & \multicolumn{2}{c}{$\eta$} & \multicolumn{2}{c}{$R$} \\
    \cmidrule(lr){4-5}\cmidrule(lr){6-7}\cmidrule(lr){8-9}
    Corpus & $V$ & Bnd & BPE & UL & BPE & UL & BPE & UL \\
    \midrule
    PubChem & 256 & NMB & 0.765 [0.765, 0.765] & 0.880 [0.879, 0.880] & 0.616 [0.616, 0.616] & 0.478 [0.478, 0.479] & 0.466 [0.466, 0.466] & 0.373 [0.373, 0.373] \\
     & 256 & MB & 0.764 [0.764, 0.765] & 0.870 [0.869, 0.870] & 0.606 [0.606, 0.606] & 0.492 [0.492, 0.492] & 0.453 [0.452, 0.453] & 0.392 [0.392, 0.392] \\
     & 512 & NMB & 0.837 [0.837, 0.838] & 0.926 [0.926, 0.926] & 0.554 [0.554, 0.554] & 0.426 [0.425, 0.426] & 0.410 [0.410, 0.410] & 0.332 [0.331, 0.332] \\
     & 512 & MB & 0.836 [0.836, 0.836] & 0.918 [0.918, 0.918] & 0.545 [0.545, 0.545] & 0.438 [0.438, 0.438] & 0.397 [0.397, 0.397] & 0.349 [0.349, 0.349] \\
     & 1024 & NMB & 0.888 [0.888, 0.888] & 0.956 [0.956, 0.956] & 0.500 [0.500, 0.500] & 0.383 [0.383, 0.383] & 0.368 [0.368, 0.368] & 0.298 [0.298, 0.299] \\
     & 1024 & MB & 0.886 [0.886, 0.886] & 0.951 [0.951, 0.951] & 0.491 [0.491, 0.491] & 0.394 [0.394, 0.394] & 0.356 [0.356, 0.356] & 0.314 [0.314, 0.314] \\
     & 2048 & NMB & 0.924 [0.923, 0.924] & 0.974 [0.974, 0.974] & 0.455 [0.455, 0.455] & 0.348 [0.348, 0.348] & 0.334 [0.334, 0.334] & 0.271 [0.271, 0.271] \\
     & 2048 & MB & 0.922 [0.922, 0.922] & 0.971 [0.971, 0.971] & 0.447 [0.446, 0.447] & 0.358 [0.358, 0.358] & 0.323 [0.323, 0.323] & 0.285 [0.285, 0.285] \\
    \addlinespace
    ZINC-22 & 256 & NMB & 0.798 [0.798, 0.798] & 0.907 [0.907, 0.907] & 0.583 [0.583, 0.583] & 0.476 [0.476, 0.476] & 0.450 [0.450, 0.450] & 0.399 [0.399, 0.399] \\
     & 256 & MB & 0.797 [0.796, 0.797] & 0.906 [0.906, 0.906] & 0.569 [0.568, 0.569] & 0.475 [0.475, 0.475] & 0.426 [0.426, 0.426] & 0.405 [0.404, 0.405] \\
     & 512 & NMB & 0.859 [0.859, 0.859] & 0.948 [0.948, 0.948] & 0.523 [0.523, 0.523] & 0.423 [0.423, 0.423] & 0.397 [0.396, 0.397] & 0.355 [0.355, 0.355] \\
     & 512 & MB & 0.855 [0.855, 0.856] & 0.948 [0.947, 0.948] & 0.511 [0.511, 0.511] & 0.422 [0.422, 0.422] & 0.374 [0.374, 0.374] & 0.360 [0.360, 0.360] \\
     & 1024 & NMB & 0.905 [0.905, 0.905] & 0.972 [0.972, 0.972] & 0.471 [0.471, 0.471] & 0.381 [0.381, 0.381] & 0.357 [0.356, 0.357] & 0.319 [0.319, 0.319] \\
     & 1024 & MB & 0.902 [0.902, 0.903] & 0.971 [0.971, 0.971] & 0.460 [0.460, 0.460] & 0.380 [0.380, 0.380] & 0.336 [0.336, 0.336] & 0.324 [0.324, 0.324] \\
    \addlinespace
    COCONUT & 256 & NMB & 0.824 [0.824, 0.825] & 0.898 [0.898, 0.898] & 0.575 [0.574, 0.576] & 0.483 [0.482, 0.483] & 0.469 [0.469, 0.470] & 0.384 [0.384, 0.385] \\
     & 256 & MB & 0.819 [0.818, 0.820] & 0.896 [0.896, 0.897] & 0.569 [0.568, 0.570] & 0.488 [0.488, 0.489] & 0.442 [0.441, 0.443] & 0.407 [0.406, 0.408] \\
     & 512 & NMB & 0.887 [0.887, 0.888] & 0.939 [0.939, 0.940] & 0.513 [0.512, 0.513] & 0.429 [0.429, 0.430] & 0.416 [0.415, 0.416] & 0.342 [0.341, 0.343] \\
     & 512 & MB & 0.882 [0.881, 0.882] & 0.936 [0.936, 0.936] & 0.507 [0.507, 0.508] & 0.434 [0.434, 0.435] & 0.391 [0.390, 0.391] & 0.362 [0.361, 0.363] \\
     & 1024 & NMB & 0.925 [0.925, 0.926] & 0.963 [0.963, 0.963] & 0.462 [0.461, 0.462] & 0.386 [0.386, 0.387] & 0.374 [0.374, 0.374] & 0.308 [0.307, 0.308] \\
     & 1024 & MB & 0.921 [0.921, 0.922] & 0.960 [0.959, 0.960] & 0.457 [0.456, 0.457] & 0.391 [0.390, 0.391] & 0.351 [0.351, 0.352] & 0.325 [0.325, 0.326] \\
    \addlinespace
    REAL-Space & 1024 & NMB & 0.901 [0.901, 0.901] & 0.973 [0.973, 0.973] & 0.469 [0.469, 0.469] & 0.357 [0.357, 0.357] & 0.348 [0.348, 0.348] & 0.283 [0.283, 0.283] \\
     & 1024 & MB & 0.901 [0.901, 0.901] & 0.973 [0.973, 0.973] & 0.458 [0.458, 0.458] & 0.357 [0.357, 0.357] & 0.335 [0.335, 0.335] & 0.284 [0.284, 0.284] \\
    \bottomrule
  \end{tabular}%
  }
\end{table}

%% file: tables/absorption_detail.tex
\begin{table}[htbp]
  \centering
  \caption{Per-condition whole-pretoken absorption. Per matched pair: the held-out fraction of pretokens emitted as a single token \citep{reddy2025diminishing-tokenization}, per arm with a 95\% molecule-resampled bootstrap CI, and the cross-arm gap $\Delta$abs (BPE $-$ UL). BPE absorbs almost every pretoken whole; Unigram-LM (UL) splits far more, the per-pretoken face of the fertility gap (\S\ref{ssec:r-mechanism}).}
  \label{tab:results-absorption}
  \footnotesize
  \begin{tabular}{@{}l r l r r r@{}}
    \toprule
    & & & \multicolumn{2}{c}{abs} & \\
    \cmidrule(lr){4-5}
    Corpus & $V$ & Bnd & BPE & UL & $\Delta$abs \\
    \midrule
    PubChem & 256 & NMB & 0.952 [0.951, 0.952] & 0.723 [0.723, 0.724] & $+0.228$ \\
     & 256 & MB & 0.974 [0.974, 0.974] & 0.738 [0.738, 0.738] & $+0.237$ \\
     & 512 & NMB & 0.967 [0.966, 0.967] & 0.724 [0.724, 0.724] & $+0.243$ \\
     & 512 & MB & 0.993 [0.993, 0.993] & 0.739 [0.738, 0.739] & $+0.255$ \\
     & 1024 & NMB & 0.970 [0.970, 0.970] & 0.724 [0.724, 0.724] & $+0.246$ \\
     & 1024 & MB & 0.997 [0.997, 0.998] & 0.739 [0.739, 0.739] & $+0.259$ \\
     & 2048 & NMB & 0.971 [0.971, 0.971] & 0.724 [0.724, 0.724] & $+0.247$ \\
     & 2048 & MB & 0.999 [0.999, 0.999] & 0.739 [0.739, 0.739] & $+0.260$ \\
    \addlinespace
    ZINC-22 & 256 & NMB & 0.931 [0.931, 0.931] & 0.715 [0.715, 0.715] & $+0.216$ \\
     & 256 & MB & 0.982 [0.982, 0.982] & 0.716 [0.715, 0.716] & $+0.266$ \\
     & 512 & NMB & 0.945 [0.945, 0.945] & 0.715 [0.715, 0.715] & $+0.229$ \\
     & 512 & MB & 0.998 [0.998, 0.998] & 0.716 [0.716, 0.716] & $+0.282$ \\
     & 1024 & NMB & 0.946 [0.946, 0.946] & 0.715 [0.715, 0.715] & $+0.231$ \\
     & 1024 & MB & 1.000 [1.000, 1.000] & 0.716 [0.716, 0.716] & $+0.284$ \\
    \addlinespace
    COCONUT & 256 & NMB & 0.888 [0.887, 0.889] & 0.723 [0.722, 0.723] & $+0.166$ \\
     & 256 & MB & 0.990 [0.990, 0.991] & 0.724 [0.723, 0.725] & $+0.266$ \\
     & 512 & NMB & 0.895 [0.894, 0.895] & 0.723 [0.722, 0.724] & $+0.171$ \\
     & 512 & MB & 0.998 [0.998, 0.998] & 0.724 [0.724, 0.725] & $+0.274$ \\
     & 1024 & NMB & 0.896 [0.895, 0.897] & 0.723 [0.722, 0.724] & $+0.173$ \\
     & 1024 & MB & 1.000 [1.000, 1.000] & 0.724 [0.724, 0.725] & $+0.275$ \\
    \addlinespace
    REAL-Space & 1024 & NMB & 0.970 [0.970, 0.970] & 0.708 [0.708, 0.709] & $+0.262$ \\
     & 1024 & MB & 1.000 [1.000, 1.000] & 0.709 [0.708, 0.709] & $+0.291$ \\
    \bottomrule
  \end{tabular}
\end{table}

%% file: tables/deadzone_nsweep.tex
\begin{table}[htbp]
  \centering
  \caption{Rare-token clearance $c_n$ across the firing-count sweep ($n \in \{50,100,200\}$, $p{=}0.95$). Following \citet{gowda2020optimal-vocab-nmt}, $c_n$ is the fraction of an arm's vocabulary firing at least $n$ times in the training corpus. The learnability bar $F_{p,n}$ is $c_n \ge p$, so reading each $c_n$ against $p \in \{0.90, 0.95, 0.99\}$ gives every $(p,n)$ bar outcome; $c_{100}$ at $p{=}0.95$ is the headline $F_{95\%,100}$ (Table~\ref{tab:results-delta-f}). $\dagger$ corpus too small to certify the tail at this $V$ (as in Table~\ref{tab:results-delta-f}).}
  \label{tab:results-nsweep}
  \footnotesize
  \begin{tabular}{@{}l r l r r r r r r@{}}
    \toprule
    & & & \multicolumn{2}{c}{$c_{50}$} & \multicolumn{2}{c}{$c_{100}$} & \multicolumn{2}{c}{$c_{200}$} \\
    \cmidrule(lr){4-5}\cmidrule(lr){6-7}\cmidrule(lr){8-9}
    Corpus & $V$ & Bnd & BPE & UL & BPE & UL & BPE & UL \\
    \midrule
    PubChem & 256 & NMB & 1.000 & 1.000 & 1.000 & 1.000 & 1.000 & 1.000 \\
     & 256 & MB & 0.979 & 1.000 & 0.979 & 1.000 & 0.979 & 1.000 \\
     & 512 & NMB & 1.000 & 1.000 & 1.000 & 0.997 & 1.000 & 0.992 \\
     & 512 & MB & 0.994 & 1.000 & 0.994 & 1.000 & 0.994 & 0.997 \\
     & 1024 & NMB & 1.000 & 0.988 & 1.000 & 0.957 & 1.000 & 0.881 \\
     & 1024 & MB & 0.991 & 0.993 & 0.991 & 0.979 & 0.990 & 0.939 \\
     & 2048$^{\dagger}$ & NMB & 0.997 & 0.746 & 0.997 & 0.567 & 0.996 & 0.430 \\
     & 2048$^{\dagger}$ & MB & 0.983 & 0.803 & 0.980 & 0.641 & 0.975 & 0.501 \\
    \addlinespace
    ZINC-22 & 256 & NMB & 1.000 & 1.000 & 1.000 & 1.000 & 1.000 & 1.000 \\
     & 256 & MB & 0.959 & 1.000 & 0.959 & 1.000 & 0.948 & 1.000 \\
     & 512$^{\dagger}$ & NMB & 1.000 & 0.668 & 1.000 & 0.568 & 0.994 & 0.439 \\
     & 512$^{\dagger}$ & MB & 0.977 & 0.686 & 0.977 & 0.592 & 0.972 & 0.479 \\
     & 1024$^{\dagger}$ & NMB & 1.000 & 0.668 & 0.999 & 0.568 & 0.935 & 0.439 \\
     & 1024$^{\dagger}$ & MB & 0.985 & 0.678 & 0.984 & 0.585 & 0.938 & 0.473 \\
    \addlinespace
    COCONUT & 256$^{\dagger}$ & NMB & 1.000 & 1.000 & 1.000 & 0.897 & 1.000 & 0.804 \\
     & 256 & MB & 0.979 & 1.000 & 0.979 & 1.000 & 0.979 & 0.938 \\
     & 512$^{\dagger}$ & NMB & 0.994 & 0.382 & 0.994 & 0.272 & 0.983 & 0.212 \\
     & 512$^{\dagger}$ & MB & 0.972 & 0.484 & 0.969 & 0.365 & 0.958 & 0.280 \\
     & 1024$^{\dagger}$ & NMB & 0.765 & 0.206 & 0.566 & 0.152 & 0.395 & 0.119 \\
     & 1024$^{\dagger}$ & MB & 0.790 & 0.221 & 0.576 & 0.168 & 0.401 & 0.131 \\
    \addlinespace
    REAL-Space & 1024$^{\dagger}$ & NMB & 1.000 & 0.712 & 1.000 & 0.704 & 0.998 & 0.637 \\
     & 1024$^{\dagger}$ & MB & 0.993 & 0.720 & 0.993 & 0.713 & 0.990 & 0.652 \\
    \bottomrule
  \end{tabular}
\end{table}